\pdfoutput=1

\documentclass[11pt]{article}

\usepackage[final]{acl}
\usepackage[most]{tcolorbox}
\usepackage{times}
\usepackage{latexsym}
\usepackage{algorithm}
\usepackage{float}
\usepackage{pifont}
\usepackage{algorithmic}
\usepackage{amsthm}
\usepackage[T1]{fontenc}
\usepackage{tcolorbox}
\usepackage{enumitem}
\usepackage{pifont} 
\usepackage[utf8]{inputenc}
\usepackage{microtype}
\usepackage{inconsolata}
\usepackage{makecell}
\usepackage{multirow}
\usepackage{setspace}
\usepackage{threeparttable}
\usepackage{CJKutf8}
\usepackage{enumitem}
\usepackage{enumerate}
\usepackage{amsfonts,amssymb,amsmath} 
\usepackage{color}
\usepackage{graphicx}
\usepackage{tabularx} 
\usepackage{placeins}
\usepackage{float} 
\usepackage[normalem]{ulem}

\usepackage{xcolor}
\usepackage{booktabs}
\usepackage{tcolorbox}
\usepackage{mathabx}
\usepackage{xcolor,colortbl}

\tcbuselibrary{skins, breakable, theorems}

\definecolor{myblue}{RGB}{221, 233, 247}
\newcommand{\classbg}[1]{\setlength{\fboxsep}{1pt}\colorbox{myblue}{\textbf{#1}}}

\lstdefinelanguage{mycase}{
    basicstyle=\scriptsize\ttfamily, 
    moredelim = [s][\color{mygrey}]{\{}{\}},
}
\usepackage{listings}
\tcbuselibrary{listings, most}  

\newtcblisting{showcase}[1][]{
  enhanced,
  arc=0em,
  boxrule=.5pt,
  listing only,
  listing options={
    language=mycase,
    upquote=true,
    basicstyle=\scriptsize \ttfamily \setlength{\baselineskip}{1.1\baselineskip},
    breaklines=true,  
    breakindent=0pt,
    xleftmargin=0pt,
    xrightmargin=0pt,
    aboveskip=-4pt,
    belowskip=-4pt,
    columns=fullflexible,
    escapeinside={|}{|},
  },
  colback=white,
  colframe=gray,
  breakable,  
  colbacktitle=gray!10,        
  coltitle=black,              
  attach boxed title to top center={yshift=-3mm},
  #1
}

\newtcolorbox[list inside=prompt,auto counter,number within=section]{prompt}[1][]{
    colbacktitle=black!60,
    coltitle=white,
    fontupper=\footnotesize,
    boxsep=5pt,
    left=0pt,
    right=0pt,
    top=0pt,
    bottom=0pt,
    boxrule=1pt,
    #1,
}

\newtcolorbox[auto counter,number within=chapter]{prompt2}[1][]{
  enhanced,
  breakable,
  fontupper=\footnotesize,
  fonttitle=\scshape,
  title={Definition \thetcbcounter},
  #1,
}
\usepackage[%
    framemethod=tikz,
    skipbelow=\topskip,
    skipabove=\topskip
]{mdframed}
\mdfsetup{%
    leftmargin=0pt,
    rightmargin=0pt,
    backgroundcolor=bggray,
    middlelinecolor=black,
    roundcorner=3
}

\newmdenv[
  backgroundcolor=black!05,
  linecolor=quoteborder,
  skipabove=1em,
  skipbelow=1em,
  leftline=true,
  topline=false,
  bottomline=false,
  rightline=false,
  linecolor=black!40,
  linewidth=4pt,
  font=\small,
  leftmargin=0cm
]{prompt_env}

\title{Who Writes What: Unveiling the Impact of Author Roles on AI-generated Text Detection}

\author{
 \textbf{Jiatao Li\textsuperscript{1,2}},
 \textbf{Xiaojun Wan\textsuperscript{1}}
\\
 \textsuperscript{1}Wangxuan Institute of Computer Technology, Peking University
\\
 \textsuperscript{2}Information Management Department, Peking University
 \\
 \texttt{{leejames@stu.pku.edu.cn}},
 \texttt{wanxiaojun@pku.edu.cn}
}

\begin{document}
\maketitle

\begin{abstract}
The rise of Large Language Models (LLMs) necessitates accurate AI-generated text detection. However, current approaches largely overlook the influence of author characteristics. We investigate how sociolinguistic attributes—gender, CEFR proficiency, academic field, and language environment—impact state-of-the-art AI text detectors. Using the ICNALE corpus of human-authored texts and parallel AI-generated texts from diverse LLMs, we conduct a rigorous evaluation employing multi-factor ANOVA and weighted least squares (WLS). Our results reveal significant biases: CEFR proficiency and language environment consistently affected detector accuracy, while gender and academic field showed detector-dependent effects. These findings highlight the crucial need for socially aware AI text detection to avoid unfairly penalizing specific demographic groups. We offer novel empirical evidence, a robust statistical framework, and actionable insights for developing more equitable and reliable detection systems in real-world, out-of-domain contexts. This work paves the way for future research on bias mitigation, inclusive evaluation benchmarks, and socially responsible LLM detectors.
\end{abstract}
\section{Introduction}
\begin{figure*}[!t]
    \centering
    \includegraphics[width=1\textwidth]{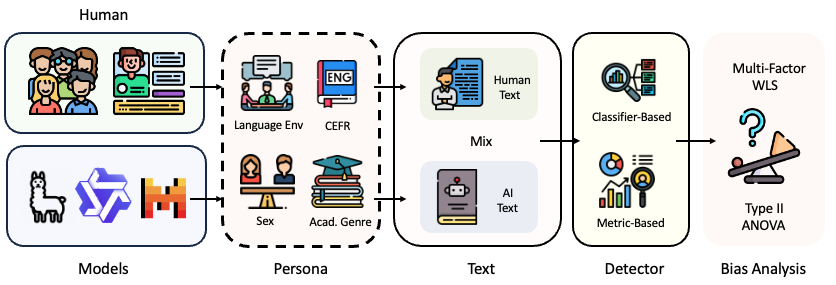}
    \caption{Research workflow: ICNALE data and LLM-generated text with author attributes are used for out-of-domain evaluation of AI text detectors and subsequent bias analysis using multi-factor ANOVA and weighted OLS.}
    \label{fig:workflow}
\end{figure*}

Large Language Models (LLMs) have transformed the way we produce and consume written communication, driving remarkable advancements in text generation across a wide array of domains. From social media content moderation to automated report writing in academia, the breadth and sophistication of these models have brought renewed focus to a pressing challenge: detecting AI-generated text. Although recent efforts have expanded detection benchmarks to include multiple LLMs, multilingual settings, and adversarial perturbations \citep{tao2024reliabledetectionllmgeneratedtexts, dugan2024raidsharedbenchmarkrobust, Li2024mage, wang2024m4gtbenchevaluationbenchmarkblackbox}, the \emph{human dimension} of text production has received far less scrutiny. Traditional detection pipelines predominantly center on model-level and data-centric strategies---e.g., sampling, prompting, and adversarial augmentation---while overlooking social and linguistic factors tied to the authors themselves.

Yet decades of sociolinguistic and applied linguistics research underscore that writing style varies systematically by gender, language proficiency, disciplinary norms, and cultural context \citep{hyland2000disciplinary, lantolf2000sociocultural, long1996role}. Such differences manifest in vocabulary choices, syntactic complexity, and rhetorical conventions, which can inadvertently bias AI text detectors if not carefully addressed. For example, non-native speakers or lower-proficiency writers might be wrongly flagged as AI-generated due to errors or unidiomatic expressions that differ from native-level training data. Conversely, sophisticated second-language learners might produce polished texts that resemble those generated by larger LLMs, thereby complicating detection efforts. In the absence of benchmarks that account for such diversity, existing detectors risk disproportionately penalizing particular demographic or educational groups.

Motivated by this gap, we pose the question: \emph{``Who is writing?''} Specifically, we investigate whether and how author-level attributes---including gender, Common European Framework of Reference (CEFR) proficiency, academic field, and language environment (e.g., native vs.\ EFL/ESL)---systematically influence AI (or AI-generated) text detection outcomes. To this end, we craft a comprehensive end-to-end analysis. First, we source human-authored texts with rich metadata from the ICNALE corpus, a well-established learner corpus that spans multiple Asian regions and diverse proficiency levels \citep{Ishikawa2013icnale}. Next, we generate parallel AI-written texts using a range of modern LLMs (e.g., Qwen, LLaMA, Mistral). We then evaluate cutting-edge off-the-shelf detectors under out-of-domain conditions---a scenario closely mirroring real-world deployments where detectors must generalize beyond their training distributions. Figure~\ref{fig:workflow} offers a concise overview of our end-to-end methodology, culminating in an out-of-domain evaluation that mirrors real-world use cases.

Our findings, drawn from rigorous \emph{t}-tests and ANOVA analyses, show that \textbf{CEFR proficiency level} and \textbf{language environment} exert consistent and sizable effects on detection accuracy across nearly all detectors. By contrast, gender- and academic field-related biases tend to be more detector-dependent, with some models demonstrating marked variability and others remaining largely agnostic. These results suggest a strong need to incorporate demographic and linguistic attributes into AI text detection benchmarks and training protocols. Without doing so, state-of-the-art detectors are liable to unfairly penalize certain writer populations or misclassify legitimate texts.

In summary, our work highlights the \emph{human-centered} challenges that arise when AI text detection collides with real-world diversity. By shifting the focus toward the people behind the text, we aim to catalyze a broader conversation on socially responsible model development, enhanced debiasing strategies, and inclusive evaluation frameworks. We hope that this renewed emphasis will pave the way for fairer, more robust, and ultimately more trustworthy detection systems. 

Our contributions are as follows \footnote{Our code and data are released at \url{https://github.com/leejamesss/AuthorAwareDetection} 
to facilitate related research.}: 
\begin{itemize}
\item We introduce a 67K-text dataset that combines human-authored ICNALE essays with parallel AI outputs from 12 modern LLMs, each labeled with detailed persona metadata. This resource enables rigorous, human-centered analysis in AI text detection. 
\item We propose a multi-factor WLS framework with Type~II ANOVA, controlling for multiple attributes to isolate each factor’s influence on detection errors and pinpoint where bias emerges. This approach can also be extended to future bias analysis tasks that consider additional demographic or linguistic variables.
\item We demonstrate that human-level attributes—particularly language proficiency and environment—can overshadow model-based differences, causing significant biases in AI text detection. 
\end{itemize}
\section{Related Work}
\label{sec:related}

\paragraph{Author Characteristics and Linguistic Theory}
Linguistic research demonstrates how author characteristics influence writing style.  Functional linguistics and discourse analysis show that registers and genres vary in vocabulary, syntax, and pragmatics \citep{biber1998corpus}, creating distinctions between AI- and human-generated text. Sociolinguistics and Second Language Acquisition (SLA) research reveals textual differences (e.g., lexical diversity, grammatical complexity) between native and non-native speakers \citep{hyland2000disciplinary, lantolf2000sociocultural, long1996role}.  \citet{hyland2000disciplinary} specifically highlights how disciplinary norms in academic writing create stylistic variations.  "World Englishes" theory \citep{kachru1985standards, jenkins2003world} further emphasizes the diverse forms of English usage across cultures.  This linguistic diversity complicates AI text detection, potentially introducing bias if detectors are trained on limited data and fail to generalize across varied social and cultural backgrounds.

\paragraph{AI Text Detection Datasets and Benchmarks}

As LLMs continue to advance in various NLP tasks, the need to distinguish AI-generated text from human-authored text has become a hot research topic. Early work explored supervised learning or feature engineering, but as models scale, attention has shifted to multi-scenario, multi-model, and multilingual detection benchmarks.

Examples include CUDRT, which focuses on polished and post-processed LLM text \citep{tao2024reliabledetectionllmgeneratedtexts}; RAID, a comprehensive benchmark with multiple models, domains, languages, and adversarial samples \citep{dugan2024raidsharedbenchmarkrobust}; MAGE, which systematically combines models and domains for in- and out-of-domain evaluations \citep{Li2024mage}; M4GT-Bench, for multilingual, black-box detection \citep{wang2024m4gtbenchevaluationbenchmarkblackbox}; and HC3, which compares ChatGPT and human expert outputs \citep{guo2023closechatgpthumanexperts}.

Despite this progress in creating diverse detection benchmarks, the human dimension—specifically, the demographic and identity attributes of authors—has received limited attention. The influence of factors like gender, native language, academic discipline, and language proficiency on detection accuracy and potential bias remains largely unexplored.  This study addresses this gap by analyzing how these key author attributes affect AI text detection.

\section{Dataset Creation}

\subsection{Overview}
We construct our dataset from 2,569 learners in the International Corpus Network of Asian Learners of English (ICNALE)~\citep{Ishikawa2013icnale}, each providing short essays on two prompts: banning smoking in restaurants and the importance of part-time jobs. This yields 5,138 human-authored texts. ICNALE also includes rich demographic and identity data (e.g., sex, academic genre, proficiency), as shown in Table~\ref{tab:attribute_dist}.

To capture a broader range of writing styles, we pair each human-authored text with parallel responses from 12 modern LLMs. Each model thus produces another 5,138 essays, adding up to 61,656 AI-generated texts. Overall, our corpus contains 66,794 samples. We acquired the ICNALE data under its official license, following all guidelines. The consistent task prompts and detailed metadata enable controlled investigations of how AI models handle varied writer profiles.

Table~\ref{tab:all-datasets} compares our dataset to other publicly available resources of AI-generated text. While many incorporate multiple models, none (to our knowledge) matches our depth of persona annotations. Our corpus thus serves as a valuable benchmark for research on model fairness and sociolinguistic bias. Detailed ICNALE task prompts appear in Appendix~\ref{appendix:task_prompt}.

\begin{table}[ht!]
\centering
\resizebox{\columnwidth}{!}{%
  \small
  \setlength{\tabcolsep}{4pt}
  \renewcommand{\arraystretch}{1.25}
  \begin{tabular}{lccc}
    \toprule
    \textbf{Name} & \textbf{Size} & \textbf{MultiModel} & \textbf{Persona}\\
    \midrule
    TuringBench~\cite{uchendu2021turingbench}        & 200k  & \ding{51} & \ding{55}\\
    RuATD~\cite{Shamardina_2022}                     & 215k  & \ding{51} & \ding{55}\\
    HC3~\cite{guo2023close}                          & 26.9k & \ding{55} & \ding{55}\\
    MGTBench~\cite{he2023mgtbench}                   & 2817  & \ding{51} & \ding{55}\\
    MULTITuDE~\cite{macko-etal-2023-multitude}       & 74.1k & \ding{51} & \ding{55}\\
    AuText2023~\cite{sarvazyan2023overview}          & 160k  & \ding{55} & \ding{55}\\
    M4~\cite{wang2023m4}                             & 122k  & \ding{51} & \ding{55}\\
    CCD~\cite{wang2023evaluating}                    & 467k  & \ding{55} & \ding{55}\\
    IMDGSP~\cite{mosca-etal-2023-distinguishing}     & 29k   & \ding{51} & \ding{55}\\
    HC-Var~\cite{xu2023generalization}               & 145k  & \ding{55} & \ding{55}\\
    HC3 Plus~\cite{su2024hc3}                        & 210k  & \ding{55} & \ding{55}\\
    MAGE~\cite{Li2024mage}                           & 447k  & \ding{51} & \ding{55}\\
    RAID~\cite{dugan2024raidsharedbenchmarkrobust}   & 6.2M  & \ding{51} & \ding{55}\\
    \midrule
    \textbf{Ours}                                    & 67k   & \ding{51} & \ding{51}\\
    \bottomrule
  \end{tabular}
}
\caption{Comparison of publicly available AI-generated text datasets. \emph{MultiModel} indicates whether each benchmark includes texts from multiple LLMs (\ding{51}) or only one (\ding{55}). \emph{Persona} indicates whether the dataset provides demographic or identity metadata.}
\label{tab:all-datasets}
\end{table}

\subsection{Author Attribute Construction}
In selecting author attributes, we consider both data feasibility and theoretical grounding. Practically, we choose attributes with minimal missing values, relatively balanced distributions, and manageable category sizes. Theoretically, we prioritize attributes that are well-studied in sociolinguistics, second-language acquisition (SLA), and academic discourse, so they can reveal deeper insights into potential biases. Accordingly, we focus on four key dimensions:

\paragraph{Gender (Sex).}
As indicated by \citet{cameron2005language,gal2012language}, gender is shaped by societal and cultural constructions, potentially giving rise to “gendered” linguistic features. Table~\ref{tab:attribute_dist} shows that 1,430 learners self-identified as female (\emph{F}) and 1,139 as male (\emph{M}), for a total of 2,569.

\paragraph{Academic Field (Acad.\ Genre).}
Rooted in genre theory \citep{swales2014genre} and Bourdieu’s concept of academic capital \citep{bourdieu1991language}, different disciplines (e.g., Humanities, Social Sciences) may exhibit distinct rhetorical styles \citep{hyland2000disciplinary}. Our dataset includes four such academic fields, with Sciences \& Technology (1,034) being the largest group (Table~\ref{tab:attribute_dist}).

\paragraph{CEFR Proficiency Level (CEFR).}
The Common European Framework of Reference for Languages \citep{council2001common} provides a graded scale for language proficiency (A2, B1, B2, etc.), widely used in SLA research \citep{krashen2006input,long1996role,lantolf2000sociocultural}. As shown in Table~\ref{tab:attribute_dist}, the most common levels in our dataset are B1\_1 (914) and B1\_2 (881), with smaller counts at A2\_0 (470), B2\_0 (231), and XX\_0 (73; native speakers).

\paragraph{Language Environment (NS/EFL/ESL).}
Following Kachru’s notion of “World Englishes” \citep{kachru1985standards} and Jenkins’ work on English as a lingua franca \citep{jenkins2003world}, we distinguish between native speakers (NS, 73 learners), EFL (English as a Foreign Language, 1,886 learners), and ESL (English as a Second Language, 610 learners). Such differences in exposure and context can shape writing styles that AI text detectors may treat unevenly.

\begin{table}[t]
\centering
\small
\begin{tabular}{lll}
\toprule
\textbf{Attribute} & \textbf{Value} & \textbf{Count} \\
\midrule
\multirow{5}{*}{CEFR}   
 & B1\_1 & 914 \\
 & B1\_2 & 881 \\
 & A2\_0 & 470 \\
 & B2\_0 & 231 \\
 & XX\_0 & 73 \\
\midrule
\multirow{4}{*}{Acad.\ Genre}  
 & Sciences \& Tech. & 1,034 \\
 & Social Sciences   & 762 \\
 & Humanities        & 674 \\
 & Life Sciences     & 99 \\
\midrule
\multirow{3}{*}{Lang.\ Env.}  
 & EFL & 1,886 \\
 & ESL & 610 \\
 & NS  & 73 \\
\midrule
\multirow{2}{*}{Sex}
 & F   & 1,430 \\
 & M   & 1,139 \\
\bottomrule
\end{tabular}
\caption{Distribution of author attributes across 2,569 learners.}
\label{tab:attribute_dist}
\end{table}

These four attributes are both available at scale (with minimal missing data) and theoretically grounded in prior linguistic research. By focusing on them, we can systematically examine how AI detectors handle diverse writer profiles and detect potential biases. Table~\ref{tab:attribute_dist} shows the distribution of each category in our dataset.

\paragraph{Definition of Subgroup.}  For our analysis, we define a subgroup based on the values of a categorical author attribute.  Let \(A\) represent a categorical attribute (e.g., CEFR level, Language Environment, Academic Genre, Sex).  This attribute \(A\) can take on a set of  \(k\) possible values, denoted as  \(\{\mathcal{A}_1, \ldots, \mathcal{A}_k\}\).  Each \(\mathcal{A}_i\) (where \(i \in \{1, 2, \ldots, k\}\)) represents a specific value of the attribute \(A\) and defines a distinct subgroup. For example, if \(A\) is CEFR level, then \(\mathcal{A}_1\) might be A2\_0, \(\mathcal{A}_2\) might be B1\_1, and so on.  All texts associated with a particular attribute value \(\mathcal{A}_i\) constitute that subgroup.

\subsection{Generators}

We employ a diverse set of Large Language Models (LLMs) to generate texts, spanning different architectures, training sets, and parameter sizes (0.5B–72B). Our lineup includes the Qwen2.5 family \citep{qwen2,qwen2.5} from Alibaba DAMO Academy, recognized for strong instruction-following performance and extended-context capabilities; the LLaMA3.1 and LLaMA3.2 models \citep{dubey2024llama3herdmodels}, noted for their open-source availability and efficient pre-training; and the compact yet high-performing Mistral \citep{jiang2023mistral7b}.

Table~\ref{tab:LLMs} lists the specific model versions used. Each model was prompted with carefully designed instructions to emulate learners with diverse gender, academic fields, CEFR proficiency levels, and language environments (see Appendix~\ref{sec:prompt-content}). By covering a wide range of parameter scales and training paradigms, our approach seeks to capture heterogeneous AI-generated outputs and enable us to evaluate detection robustness across various author attributes.

\begin{table}[t]
\centering
\small
\begin{tabular}{lll}
\toprule
\textbf{LLM Name} & \textbf{Series} & \textbf{Params} \\
\midrule
Qwen2.5-0.5B-Instruct & Qwen2.5 & 0.5B \\
Qwen2.5-1.5B-Instruct & Qwen2.5 & 1.5B \\
Qwen2.5-3B-Instruct   & Qwen2.5 & 3B \\
Qwen2.5-7B-Instruct   & Qwen2.5 & 7B \\
Qwen2.5-14B-Instruct  & Qwen2.5 & 14B \\
Qwen2.5-32B-Instruct  & Qwen2.5 & 32B \\
Qwen2.5-72B-Instruct  & Qwen2.5 & 72B \\
\midrule
Llama3.1-8b-instruct  & Llama3.1 & 8B \\
Llama3.1-70b-instruct & Llama3.1 & 70B \\
\midrule
Llama3.2-1b-instruct  & Llama3.2 & 1B \\
Llama3.2-3b-instruct  & Llama3.2 & 3B \\
\midrule
Mistral-Small-Instruct-2409 & Mistral & 22B \\
\bottomrule
\end{tabular}
\caption{List of employed LLMs}
\label{tab:LLMs}
\end{table}

\section{Detector}
\label{sec:detector}
\subsection{Detector Selection}
Building upon the methodology of \citet{dugan2024raidsharedbenchmarkrobust} (RAID), we evaluate detectors from two categories: \textbf{classifier-based}, \textbf{metric-based}. Classifier-based detectors typically involve fine-tuning a pre-trained language model such as RoBERTa \cite{liu2019roberta}, while metric-based detectors compute a value based on generative model probabilities. 

Specifically, following RAID’s unified pipeline and detector set, we tested the detectors summarized in Table~\ref{tab:detectors}.

\begin{table}[ht]
\centering
\small
\begin{tabular}{l p{0.6\linewidth}}
\toprule
\textbf{Category} & \textbf{Detectors} \\
\midrule
\textbf{Classifier-Based} &
  RoBERTa-Base (GPT2) \cite{solaiman-etal-2019}\\
  & RoBERTa-Large (GPT2) \cite{liu2019roberta}\\
  & RoBERTa-Base (ChatGPT) \cite{guo2023close}\\
  & RADAR \cite{hu2023radar}\\
\midrule
\textbf{Metric-Based} &
  GLTR \cite{gehrmann-etal-2019-gltr} \\
  & Binoculars \cite{hans2024spotting}\\
  & Fast DetectGPT \cite{bao2023fastdetectgpt}\\
  & DetectGPT \cite{mitchell-etal-2023-detectgpt} \\
  & LLMDet \cite{wu-etal-2023-llmdet}\\
\bottomrule
\end{tabular}
\caption{Overview of Detectors Evaluated}
\label{tab:detectors}
\end{table}

In contrast to \citet{Li2024mage}, we adopt RAID’s principle of analyzing off-the-shelf models without further fine-tuning them on our target dataset \citep{dugan2024raidsharedbenchmarkrobust}. This setup directly assesses the generalization capability of these pre-trained detectors. Meanwhile, for the metric-based detectors, we utilize the default generative models within each repository to replicate realistic usage scenarios.\footnote{For additional detector details, please refer to Appendix~\ref{app:detectors}.}

\subsection{Detector Evaluation}
In keeping with RAID’s evaluation paradigm, each detector produces a scalar score for a given sequence, and we transform this score into a binary AI/human judgment by setting a threshold $\tau$. We then calibrate $\tau$ so that the false positive rate (FPR) against human-authored text remains at 5\%. Accuracy at a fixed 5\% FPR thus measures how effectively each detector identifies machine-generated text without unfairly penalizing human writers. This aligns with recent trends in robustness evaluations \citep{hans2024spotting, krishna2023paraphrasing, soto2024fewshot}.\footnote{\citet{dugan2024raidsharedbenchmarkrobust} discusses detailed rationale for choosing FPR-based calibration over traditional precision/recall/F1 metrics.}

Table~\ref{tab:fpr_results} replicates the RAID benchmark’s illustration of false positive rates at naive thresholds. Evidently, failing to calibrate or disclose thresholds can cause prohibitive false positive rates—a risk we aim to mitigate by following RAID’s FPR-based framework.

\begin{table}[t]
    \small
    \centering
    \begin{tabular}{l|c|c|c|c}
    \toprule
    &$\tau$=0.25&$\tau$=0.5&$\tau$=0.75&$\tau$=0.95\\
    \midrule
    R-B GPT2&8.71\%&6.59\%&5.18\%&3.38\%\\
    R-L GPT2&6.14\%&2.91\%&1.46\%&0.25\%\\
    R-B CGPT&21.6\%&15.8\%&15.1\%&10.4\%\\
    RADAR&7.48\%&3.48\%&2.17\%&1.23\%\\
    \midrule
    GLTR&100\%&99.3\%&21.0\%&0.05\%\\
    F-DetectGPT&47.3\%&23.2\%&13.1\%&1.70\%\\
    LLMDet&97.9\%&96.0\%&92.0\%&75.3\%\\
    Binoculars&0.07\%&0.00\%&0.00\%&0.00\%\\
    \bottomrule
    \end{tabular}
    \caption{False Positive Rates for detectors on RAID at naive choices of threshold ($\tau$). We see that, for open-source detectors, thresholding naively results in unacceptably high false positive rates.}
    \label{tab:fpr_results}
\end{table}

\section{Bias Analysis}
\label{sec:bias}

This section illustrates our approach for identifying and quantifying \emph{bias} in AI text detection with respect to different author attributes (e.g., gender, CEFR level, academic field, and language environment). Our primary goal is to examine whether certain subgroups experience systematically higher or lower detection accuracy, even after controlling for other factors. Below, we introduce the core principles, define bias mathematically, and discuss why our \emph{multi-factor} method is both appropriate and advantageous over simpler alternatives. Detailed algorithmic steps appear in Appendix~\ref{sec:appendix-tests}.

\subsection{Task Definition}
\paragraph{Input-Output Principle.} Our pipeline takes as \emph{input} texts labeled as \emph{human-authored} or \emph{LLM-generated} (aggregating outputs from all LLMs). Each text includes metadata describing the author's (or LLM emulated author's) demographic and linguistic attributes. Each detector provides a binary classification (human or AI). We then group all texts (human and LLM-generated) by these attributes (e.g., CEFR level) to compute and compare detection accuracy across subgroups.

\paragraph{Definition of Bias.} 
We define bias in terms of statistically significant differences in detection accuracy across subgroups. Let \(A\) represent a categorical author attribute (e.g., CEFR level) with values \(\{\mathcal{A}_1, \ldots, \mathcal{A}_k\}\), and let \(\mathrm{Accuracy}(A=a)\) be the detection accuracy for texts where \(A=a\). A detector is considered \emph{biased} with respect to \(A\) if our multi-factor analysis reveals a significant difference in accuracy between any two attribute values \(a_i\) and \(a_j\).

We quantify this bias using a two-stage process:
1) ANOVA \(p\)-values: We first use Type II ANOVA to determine whether attribute \(A\) has a statistically significant overall effect on detection accuracy, while controlling for other attributes. A \(p\)-value below 0.05 indicates that the attribute significantly impacts detector performance.

2) Post-hoc Comparisons using LSMeans: If (and only if) the ANOVA \(p\)-value for attribute \(A\) is significant, we proceed to post-hoc pairwise comparisons.  These comparisons use Least-Squares Means (LSMeans), which are adjusted means for each level of \(A\), holding other factors constant.  We employ pairwise Wald tests with Holm correction on the LSMeans to reveal the magnitude and direction of significant differences between specific subgroups.  This pinpoints which subgroups are disproportionately affected by the detector.

\subsection{Multi-Factor Weighted Least Squares and ANOVA.}
We employ a \textbf{multi-factor regression} framework to isolate each attribute’s effect on detection accuracy, using ordinary least squares (OLS) adapted to unbalanced data through \textbf{weighted least squares (WLS)}. We then perform \textbf{Type~II ANOVA} to determine each factor’s unique contribution once other attributes are held constant \citep{scheffe_analysis_1999}. This approach is well-suited for the uneven subgroup sizes common in sociolinguistic and fairness research \citep{hardt2016equalityopportunitysupervisedlearning}, as WLS estimates how accuracy varies with each factor while Type~II ANOVA tests whether including a particular attribute (e.g., \textit{language environment}) significantly reduces unexplained variance. By including all attributes jointly, we disentangle their individual effects and correct for subgroup size imbalances.

In contrast to single-factor tests—which examine each attribute in isolation—our multi-factor WLS+ANOVA approach accounts for overlapping or correlated attributes. For instance, CEFR level and language environment may be intertwined if certain regions tend to have higher proficiency, and ignoring these dependencies risks spurious conclusions or masking real biases \citep{angwin2016machine}. Although \emph{rule-based} or \emph{single-factor} methods can provide initial insights (e.g., focusing on one attribute at a time), they cannot robustly address the interactions among multiple variables \citep{solon2017fairness}. Our multi-factor framework systematically partitions the variance in detection accuracy explained by each attribute, giving a more reliable measure of bias and enabling fairer AI text detection systems.

\paragraph{High-Level Implementation.}
Algorithm~\ref{alg:wls-anova-main} summarizes our multi-factor WLS and Type II ANOVA procedure.  Broadly, we (1) aggregate accuracy and assign weights for each unique combination of attribute values; (2) fit a WLS model, treating each attribute as a factor; and (3) apply Type II ANOVA by iteratively removing each attribute to assess its unique contribution to the model.

\begin{algorithm}[!h]
\caption{Multi-Factor WLS + Type II ANOVA}
\label{alg:wls-anova-main}
\renewcommand{\algorithmicrequire}{\textbf{Input:}}
\renewcommand{\algorithmicensure}{\textbf{Output:}}
\begin{algorithmic}[1]
    \REQUIRE Dataset of \(n\) texts, each with a binary classification \(y_i\) (0: human, 1: AI) and a set of attribute values (e.g., CEFR=B1, Sex=F, ...).
    \ENSURE ANOVA table (F-statistics, \(p\)-values) and LSMeans.

    \STATE \textbf{Data Preparation:}
        \STATE \quad Group texts by unique combinations of attribute values.
        \STATE \quad Calculate the mean accuracy \(a_j\) and weight \(w_j\) (text count) for each group \(j\).
        \STATE \quad Treat each attribute as a factor.

    \STATE \textbf{Fit WLS Model:} Estimate parameters \(\boldsymbol{\beta}\) using weighted least squares, weighting each group's contribution by \(w_j\).

    \STATE \textbf{Perform Type II ANOVA:} For each attribute:
        \begin{itemize}[leftmargin=1.5em]
        \item Construct a reduced model by removing the attribute.
        \item Refit the WLS model to the reduced model, obtaining \(\text{RSS}_{\text{reduced}}\).
        \item Compute the \(F\)-statistic and \(p\)-value by comparing \(\text{RSS}_{\text{reduced}}\) to the full model's \(\text{RSS}_{\text{full}}\).
        \end{itemize}

    \STATE \textbf{Calculate LSMeans:} Compute LSMeans for each attribute level using the full WLS model.

    \STATE \RETURN ANOVA table and LSMeans.
\end{algorithmic}
\end{algorithm}

\paragraph{Outputs and Usage.}
The WLS and ANOVA procedure yields (i) ANOVA tables with each attribute's \(F\)-statistic, \(p\)-value, and significance, and (ii) LSMeans, providing model-adjusted accuracy for each attribute combination. By accounting for correlated attributes and weighting groups by size, we detect and interpret biases without discarding data or ignoring interactions. Appendix~\ref{appendix:hyp_testing_wls} provides full implementation details, including pseudocode (Algorithm~\ref{alg:wls-anova}) and our Python \texttt{statsmodels} setup.

\paragraph{References to Additional Analyses.}
While our \emph{primary} approach is the \textbf{multi-factor} WLS, we also conduct complementary tests for confirmation or exploratory checks. First, we perform \textbf{single-factor analyses}, such as Welch’s \emph{t}-test or one-way (weighted) ANOVA, when focusing on a single categorical attribute in isolation. Second, we explore \textbf{data subsetting}, in which we match certain attributes or filter the data to reduce confounding (often shrinking the sample size). Finally, we implement \textbf{down-sampling} to balance group sizes, potentially discarding data from larger subgroups (see Appendix~\ref{sec:appendix-tests}). Although these approaches can yield valuable insights, they are \emph{less comprehensive} in controlling for multiple attributes simultaneously, reinforcing the importance of multi-factor methodologies as the central tool in our bias assessment.

\subsection{Hypothesis Testing Results}
\label{sec:hyp_test_results}

To quantify how each demographic or contextual factor influences detector performance, we conduct a two-stage bias analysis. First, we use multi-factor Weighted Least Squares (WLS) regression and Type II ANOVA to determine which author attributes (factors) significantly influence detector accuracy. Second, for those attributes found to be significant in the ANOVA, we conduct post-hoc pairwise comparisons using Least-Squares Means (LSMeans) to identify which specific levels of the attribute exhibit significant differences in accuracy.

\paragraph{ANOVA Results.} Table~\ref{tab:bias_table_neat} presents the ANOVA results, with \(p\)-values for each factor (CEFR, Sex, Academic Genre, Language Environment) and detector. A \(p\)-value below 0.05 signifies a statistically significant effect of the factor on accuracy, controlling for other factors.  \textbf{CEFR proficiency level} (\texttt{cefr}) is highly significant for all detectors, indicating a strong, consistent bias related to language proficiency.  \textbf{Language environment} (\texttt{language\_env}) is significant for most detectors, suggesting the context of English learning (EFL, ESL, or NS) matters. \textbf{Sex} (\texttt{Sex}) shows no significant effect, implying no evidence of gender-based bias. \textbf{Academic genre} (\texttt{academic\_genre}) exhibits detector-dependent effects, with about half the detectors showing significant differences across fields.

\begin{table}[ht]
    \centering
    \small
    \begin{tabular}{>{\raggedright\arraybackslash}m{1.7cm}
    >{\raggedright\arraybackslash}m{0.7cm}
    >{\raggedright\arraybackslash}m{0.7cm}
    >{\raggedright\arraybackslash}m{1.2cm}
    >{\raggedright\arraybackslash}m{1.2cm}}
        \toprule
        \textbf{Detector} & 
        \textbf{CEFR} & 
        \textbf{Sex} & 
        \textbf{Academic Genre} & 
        \textbf{Language Env.} \\
        \midrule
        \rowcolors{2}{gray!10}{white}
        binoculars      & \cellcolor{green!20}\textbf{Yes}  &  No  &  No  & \cellcolor{green!20}\textbf{Yes} \\
        chatgpt-roberta & \cellcolor{green!20}\textbf{Yes}  &  No  &  No  &  No  \\
        detectgpt       & \cellcolor{green!20}\textbf{Yes}  &  No  & \cellcolor{green!20}\textbf{Yes}  & \cellcolor{green!20}\textbf{Yes} \\
        fastdetectgpt   & \cellcolor{green!20}\textbf{Yes}  &  No  &  No  & \cellcolor{green!20}\textbf{Yes} \\
        fastdetectllm   & \cellcolor{green!20}\textbf{Yes}  &  No  & \cellcolor{green!20}\textbf{Yes}  & \cellcolor{green!20}\textbf{Yes}  \\
        gltr            & \cellcolor{green!20}\textbf{Yes}  &  No  &  No  & \cellcolor{green!20}\textbf{Yes} \\
        gpt2-base       & \cellcolor{green!20}\textbf{Yes}  &  No  & \cellcolor{green!20}\textbf{Yes}  & \cellcolor{green!20}\textbf{Yes} \\
        gpt2-large      & \cellcolor{green!20}\textbf{Yes}  &  No  & \cellcolor{green!20}\textbf{Yes}  & \cellcolor{green!20}\textbf{Yes} \\
        llmdet          & \cellcolor{green!20}\textbf{Yes}  &  No  & \cellcolor{green!20}\textbf{Yes}  & \cellcolor{green!20}\textbf{Yes} \\
        radar           & \cellcolor{green!20}\textbf{Yes}  &  No  &  No  & \cellcolor{green!20}\textbf{Yes} \\
        \bottomrule
    \end{tabular}
    \caption{ANOVA $p$-values for each factor across detectors.  Columns show significance (Sig.?). ``Yes'' indicates $p<0.05$ (significant), ``No'' otherwise. Green-shaded cells indicate significant results (p < 0.05).}
    \label{tab:bias_table_neat}
\end{table}

\paragraph{Post-Hoc Comparisons and LSMeans.}

\begin{table*}[h!]
\centering
\resizebox{\textwidth}{!}{
\begin{tabular}{llcccccccccc}
\toprule
\textbf{Factor} & \textbf{Level} & \textbf{binoc.} & \textbf{chatgpt-r} & \textbf{detectgpt} & \textbf{fdgpt} & \textbf{fdllm} & \textbf{gltr} & \textbf{gpt2-base} & \textbf{gpt2-large} & \textbf{llmdet} & \textbf{radar} \\
\midrule
\multirow{5}{*}{CEFR} & A2\_0 & 0.9482 & 0.7480 & 0.7944 & 0.8963 & 0.4873 & 0.8366 & 0.6493 & 0.5402 & 0.5063 & 0.6886 \\
 & B1\_1 & 0.9443 & 0.7408 & 0.8072 & 0.8887 & 0.4842 & 0.8305 & 0.6399 & 0.5676 & 0.4961 & 0.7267 \\
 & B1\_2 & 0.9475 & 0.7377 & 0.8228 & 0.9023 & 0.4786 & 0.8206 & 0.5881 & 0.5410 & 0.4812 & 0.7210 \\
 & B2\_0 & 0.9507 & 0.7045 & 0.8289 & 0.9143 & 0.4828 & 0.7967 & 0.5623 & 0.5462 & 0.4581 & 0.7020 \\
 & XX\_0 & 0.8981 & 0.7410 & 0.7722 & 0.8748 & 0.4975 & 0.7180 & 0.5482 & 0.5162 & 0.5037 & 0.6686 \\
\midrule
\multirow{3}{*}{LangEnv} & EFL & 0.9482 & -- & 0.7944 & 0.8963 & 0.4873 & 0.8366 & 0.6493 & 0.5402 & 0.5063 & 0.6886 \\
 & ESL & 0.9337 & -- & 0.7867 & 0.8941 & 0.4837 & 0.8346 & 0.6504 & 0.5521 & 0.5250 & 0.6671 \\
 & NS & 0.8981 & -- & 0.7722 & 0.8748 & 0.4975 & 0.7180 & 0.5482 & 0.5162 & 0.5037 & 0.6686 \\
\midrule
\multirow{4}{*}{AcadGenre} & Humanities & -- & -- & 0.7944 & -- & 0.4873 & -- & 0.6493 & 0.5402 & 0.5063 & -- \\
 & Social Sciences & -- & -- & 0.7884 & -- & 0.4856 & -- & 0.6409 & 0.5238 & 0.5097 & -- \\
 & Sciences \& Tech & -- & -- & 0.7759 & -- & 0.4813 & -- & 0.6866 & 0.5705 & 0.5109 & -- \\
 & Life Sciences & -- & -- & 0.7547 & -- & 0.4732 & -- & 0.6741 & 0.5484 & 0.5265 & -- \\
\bottomrule
\end{tabular}
}
\caption{LSMeans for CEFR Level, Language Environment (EFL, ESL, NS), and Academic Genre (Humanities, Social Sciences, Sciences \& Technology, Life Sciences), across detectors. A ``--'' indicates that the overall ANOVA for that factor and detector was not statistically significant (see Table~\ref{tab:bias_table_neat}), so LSMeans are not reported.}
\label{tab:lsmeans_combined}
\end{table*}

\begin{table}[ht!]
\centering
\small
\begin{tabular}{llll}
\toprule
\textbf{Factor} & \textbf{Detector} & \textbf{Comparison} & \textbf{Sig.?} \\
\midrule
\multirow{9}{*}{CEFR} & \multirow{4}{*}{binoc.}
 & B2\_0 vs XX\_0 & Yes \\
 &  & A2\_0 vs XX\_0 & Yes \\
 &  & B1\_2 vs XX\_0 & Yes \\
 &  & B1\_1 vs XX\_0 & Yes \\
\cmidrule(lr){2-4}
 & \multirow{2}{*}{chatgpt-r}
 & A2\_0 vs B2\_0 & Yes \\
 &  & B1\_1 vs B2\_0 & Yes \\
\cmidrule(lr){2-4}
 & \multirow{2}{*}{detectgpt}
 & B2\_0 vs A2\_0 & Yes \\
 &  & B2\_0 vs XX\_0 & Yes \\
\midrule
\multirow{4}{*}{Lang. Env.} & \multirow{3}{*}{binoc.}
 & EFL vs ESL & Yes \\
 &  & EFL vs NS  & Yes \\
 &  & ESL vs NS  & Yes \\
\cmidrule(lr){2-4}
 & detectgpt
 & EFL vs NS & Yes \\
\midrule
\multirow{10}{*}{Acad. Genre} & \multirow{3}{*}{detectgpt}
 & Hum. vs S.\&T. & Yes \\
 &  & Hum. vs Life S.  & Yes \\
 &  & Soc. S. vs Life S.  & Yes \\
\cmidrule(lr){2-4}
 & \multirow{2}{*}{gpt2-base}
 & S.\&T. vs Hum. & Yes \\
 &  & S.\&T. vs Soc. S. & Yes \\
\cmidrule(lr){2-4}
 & \multirow{2}{*}{gpt2-large}
 & S.\&T. vs Hum. & Yes \\
 &  & S.\&T. vs Soc. S. & Yes \\
\cmidrule(lr){2-4}
 & \multirow{3}{*}{llmdet}
 & Life S. vs S.\&T. & Yes \\
 &  & Life S. vs Soc. S. & Yes \\
 &  & Life S. vs Hum. & Yes \\
\bottomrule
\end{tabular}
\caption{Post-hoc pairwise comparisons (Wald test, Holm correction) for CEFR Level, Language Environment, and Academic Genre. Only showing significant pairs for detectors where the overall ANOVA for the respective factor was significant (see Table~\ref{tab:bias_table_neat}).}
\label{tab:posthoc_combined}
\end{table}

For factors with significant overall effects in the ANOVA (Table~\ref{tab:bias_table_neat}), we conduct post-hoc pairwise comparisons using the Wald test with Holm correction. These tests, performed on the LSMeans, identify which specific levels of a factor differ significantly in detection accuracy. The LSMeans (Table~\ref{tab:lsmeans_combined}) represent the model-adjusted mean accuracy for each subgroup, controlling for all other factors. In other words, they estimate the average detection accuracy for a particular subgroup (e.g., CEFR level B1), assuming all other attributes (Sex, Academic Genre, Language Environment) are held constant at their average values across the dataset. This allows us to isolate the effect of the factor of interest. A "--" in Table~\ref{tab:lsmeans_combined} indicates a non-significant ANOVA result for that factor and detector, precluding meaningful post-hoc tests. Table~\ref{tab:posthoc_combined} presents the significant comparisons. To interpret the direction of significant differences (i.e., which subgroup has higher accuracy), the LSMeans values in Table~\ref{tab:lsmeans_combined} must be compared. Table~\ref{tab:posthoc_combined} reveals the following key patterns:

\textbf{CEFR Level:} `binoculars` shows the most extensive bias; all non-native CEFR levels differ significantly from native speakers (XX\_0), with higher accuracy for non-native levels (see Table~\ref{tab:lsmeans_combined}). `chatgpt-roberta` distinguishes A2 from B2, and B1.1 from B2. `detectgpt` distinguishes B2 from both A2 and XX\_0.

\textbf{Language Environment:}  For `binoculars`, all pairwise comparisons (EFL vs. ESL, EFL vs. NS, ESL vs. NS) are significant, with EFL and ESL generally showing higher accuracy than NS (see Table~\ref{tab:lsmeans_combined}). `detectgpt` shows a significant difference only between EFL and NS.

\textbf{Academic Genre:} Significant differences are found for `detectgpt`, `gpt2-base`, `gpt2-large`, and `llmdet`, but specific differing pairs vary by detector. For instance, `detectgpt` distinguishes Humanities from both Science \& Technology and Life Sciences, and Social Sciences from Life Sciences.

\section{Conclusion}
\label{sec:conclusion}
In this study, we examined how social and linguistic attributes—gender, CEFR proficiency, academic field, and language environment—affect AI text detection in realistic, out-of-domain settings. Our analyses revealed that all tested detectors are highly sensitive to CEFR level and language environment, while biases tied to gender and academic field manifest more inconsistently across models. These findings highlight the pivotal role of author diversity in shaping detection performance and underscore the need for socially aware benchmarks, debiasing strategies, and more inclusive training data. By recognizing the nuanced ways in which “who is writing” influences text characteristics, future research can foster more equitable and reliable AI detection systems that effectively serve diverse linguistic and cultural communities.

\section*{Limitations}
While our study offers valuable insights into how author-level attributes influence AI text detection, our evaluation is primarily based on open-source detectors and the ICNALE corpus, which predominantly comprises texts from Asian English learners. Future work should extend this analysis to additional systems and more diverse datasets to further validate the generalizability of our findings.

\section*{Ethical Consideration}
We follow ICNALE’s terms of use and do not redistribute its original data. Instead, we offer a public tool for analyzing locally obtained ICNALE files, ensuring researchers can replicate our workflow independently. Synthetic texts were generated under proper licenses; we share only generation methods and derived results to respect intellectual property and privacy.
\section*{Acknowledgments}
This work was supported by Beijing Science and Technology Program (Z231100007423011) and Key Laboratory of Science, Technology and Standard in Press Industry (Key Laboratory of Intelligent Press Media Technology). We appreciate the anonymous reviewers for their helpful comments. Xiaojun Wan is the corresponding author.

\bibliography{custom}
\appendix
\section{Details of multi-factor WLS}
\label{appendix:hyp_testing_wls}

To rigorously assess potential bias in each detector’s predictions across various author attributes, we adopt a \textbf{multi-factor (multivariate) analysis} with a significance threshold of \(\alpha = 0.05\). This reveals whether a focal attribute (e.g., \emph{gender}) retains an effect on detection outcomes once we \emph{control} for other variables that might jointly influence performance (e.g., \emph{CEFR level}, \emph{language environment}).

\paragraph{Weighted Least Square for Detector Accuracy.}
In our implementation, each row in the dataset often represents an aggregated outcome (e.g., mean accuracy) across several underlying samples. Let \(w_i\) be the total sample size for the \(i\)-th aggregated row, and let \(\text{accuracy}_i \in [0,1]\) be the observed detection accuracy for that group. We fit a WLS model
\begin{equation*}
\label{eq:weighted-ols}
  \min_{\boldsymbol{\beta}}
  \sum_{i=1}^{n}
  w_i \;(\text{accuracy}_i \;-\; \beta_0 \;-\; \sum_{k=1}^p \beta_k x_{ik})^{2}.
\end{equation*}
In Python \texttt{statsmodels}, this is accomplished by specifying \texttt{weights=}\(w_i\) in a model such as
\[
  \texttt{accuracy} \;\sim\; C(\texttt{gender}) \;+\; C(\texttt{CEFR}) \;+\; \dots
\]
where each predictor is treated as a categorical variable \(C(\cdot)\). If we let \(\hat{y}_i\) be the model’s prediction, then the weighted sum of squares is
\[
  \text{WSSE} 
  \;=\;
  \sum_{i=1}^n 
    w_i \;\bigl(\text{accuracy}_i - \hat{y}_i\bigr)^2.
\]
This penalizes errors in proportion to the sample size \(w_i\).

\paragraph{Multi-Factor Type~II ANOVA.}
After estimating \(\boldsymbol{\beta}\), we evaluate each attribute’s \emph{unique} contribution via \textbf{Type~II ANOVA}. Concretely, for each predictor (e.g., \emph{gender}), we compare:
\[
\begin{aligned}
  \text{RSS}_{\mathrm{reduced}} &\;(\text{model without gender}) \\
  &\quad \text{vs.} \\
  \text{RSS}_{\mathrm{full}} &\;(\text{model with gender}),
\end{aligned}
\]
where \(\text{RSS}\) is the \emph{weighted} residual sum of squares. The partial \(F\)-test for that attribute is
\begin{equation*}
\label{eq:partialF}
F 
\;=\;
\frac{\bigl(\text{RSS}_{\mathrm{reduced}} - \text{RSS}_{\mathrm{full}}\bigr)/\Delta p}
     {\;\text{RSS}_{\mathrm{full}} / \bigl(n - p_{\mathrm{full}}\bigr)},
\end{equation*}
where \(\Delta p\) is the difference in number of parameters between the two models, and \(p_{\mathrm{full}}\) is the total parameter count. If the resulting \(p\)-value falls below \(\alpha\), we conclude that \emph{gender} explains additional variance not captured by the other attributes.

\paragraph{Partial \(R^2\).}
As an effect-size measure, we may compute a \emph{partial \(R^2\)} for each predictor:
\begin{align*}
R^2_{\mathrm{partial}}
&= \frac{\text{RSS}_{\mathrm{reduced}} - \text{RSS}_{\mathrm{full}}}
         {\text{RSS}_{\mathrm{reduced}}} \\
&= 1 - \frac{\text{RSS}_{\mathrm{full}}}{\text{RSS}_{\mathrm{reduced}}}.
\end{align*}
This indicates what fraction of the previously unexplained variation is accounted for by reintroducing the focal attribute.

\paragraph{Pseudo-Code Sketch for Multi-Factor Analysis}
Algorithm~\ref{alg:wls-anova} provides a pseudocode flow that mirrors our Python \texttt{statsmodels} procedure for fitting a WLS model with multiple categorical predictors and performing \textbf{Type~II ANOVA} to assess each predictor’s significance.

By controlling for multiple attributes at once, our multi-factor framework helps isolate each attribute’s \emph{direct} influence on detection accuracy, thereby mitigating spurious correlations and confounding. This process is implemented in Python using \texttt{statsmodels} (for WLS and Type~II ANOVA) alongside supporting libraries for data preparation and hypothesis testing.

\paragraph{Partitioning the Variance}
\label{sec:variance-partition}

When we treat accuracy as a continuous outcome, we can write the total sum of squares (TSS) as
\[
  \text{TSS} 
  \;=\;
  \sum_{i=1}^n 
    w_i \,\Bigl(\text{accuracy}_i - \overline{\text{accuracy}}\Bigr)^{2},
\]
where \(\overline{\text{accuracy}}\) is the weighted grand mean. Fitting the model in Equation~\ref{eq:weighted-ols} yields predictions \(\hat{y}_i\), letting us define
\[
  \text{WSR} 
  = \sum_{i=1}^n w_i 
  \bigl(\hat{y}_i - \overline{\text{accuracy}}\bigr)^2.
\]
\noindent
\textit{Here, WSR is the “weighted sum of regression”.}
and
\[
  \text{WSSE}
  \;=\;
  \sum_{i=1}^n w_i
   \Bigl(\text{accuracy}_i - \hat{y}_i\Bigr)^2.
\]
Hence, \(\text{TSS} = \text{WSR} + \text{WSSE}\). In a multi-factor model, the partial \(F\)-tests correspond to whether \(\text{WSSE}\) drops sufficiently when an attribute is included.

\paragraph{Implementation.}
We implement all these procedures in Python. We leverage \texttt{statsmodels} for WLS fits and Type~II ANOVA, \texttt{scipy.stats} for complementary \emph{t}-tests and distribution checks, and custom matching/down-sampling routines to handle partial confounds. Part~\ref{sec:variance-partition} details how variance in accuracy is decomposed into regression (WSR) and residual (WSSE) sums of squares, while Tables~\ref{tab:bias_table} present the resulting \(p\)-values and significance decisions at \(\alpha=0.05\). Through this process, we can identify which author factors (e.g., \emph{gender}, \emph{CEFR}, \emph{environment}) exhibit genuine biases within the AI text detection pipeline, and which factors do not remain significant once confounds are accounted for.

\begin{algorithm}[!h]
\caption{Multi-Factor WLS + Type II ANOVA}
\label{alg:wls-anova}
\renewcommand{\algorithmicrequire}{\textbf{Input:}}
\renewcommand{\algorithmicensure}{\textbf{Output:}}
\begin{algorithmic}[1]
    \REQUIRE Dataset of \(n\) texts, each with:
             \begin{itemize}
                 \item A binary classification \(y_i \in \{0, 1\}\) (0: human, 1: AI).
                 \item An attribute vector \(\mathbf{x}_i\) (e.g., CEFR=B1, Sex=F, ...).
             \end{itemize}
             Significance level \(\alpha\) (e.g., 0.05).
    \ENSURE ANOVA table (F-statistics, \(p\)-values) and LSMeans.

    \STATE \textbf{Data Preparation:}
        \begin{itemize}
            \item Group texts by unique attribute combinations (\(\mathbf{x}_i\)).
            \item For each group \(j\), compute the mean accuracy \(a_j\) and weight \(w_j\) (number of texts in group).
            \item Treat each attribute as a factor.
        \end{itemize}
    \STATE \textbf{Fit WLS Model:} Estimate parameters \(\boldsymbol{\beta}\) by minimizing the weighted sum of squared errors:
    \[
    \min_{\boldsymbol{\beta}} \sum_{j=1}^g w_j\,(a_j - \mathbf{x}_j^T \boldsymbol{\beta})^2,
    \]
    where \(g\) is the number of groups, and \(\mathbf{x}_j\) is a representative attribute vector for group \(j\).

    \STATE \textbf{Perform Type II ANOVA:} For each attribute:
        \begin{itemize}[leftmargin=1.5em]
        \item Construct a reduced model by removing the attribute.
        \item Refit the WLS model to the reduced model, obtaining \(\text{RSS}_{\text{reduced}}\).
        \item Compute the \(F\)-statistic:
        \[F = \frac{(\text{RSS}_{\text{reduced}} - \text{RSS}_{\text{full}}) / \Delta p}{\text{RSS}_{\text{full}} / (n - p_{\text{full}})},\]
         where \(\Delta p\) is the difference in the number of parameters between the full and reduced models, and \(p_{\text{full}}\) is the number of parameters in the full model.
        \item Calculate the \(p\)-value from the \(F\)-distribution.
        \end{itemize}

    \STATE \textbf{Calculate LSMeans:} Compute LSMeans for each attribute level using the full WLS model.
    \STATE \RETURN ANOVA table and LSMeans.
\end{algorithmic}
\end{algorithm}

\clearpage
\section{Detector Details}
\label{app:detectors}

\paragraph{RoBERTa (GPT-2)} \cite{solaiman-etal-2019} This detector is a RoBERTa model \cite{liu2019roberta} fine-tuned on outputs from GPT-2.  The training dataset consists of GPT-2 generations from various decoding strategies (greedy, top-k=50, and random sampling) in an open-domain setting.  This detector has been a long-standing baseline in the field. We use both the base and large versions, obtained from OpenAI\footnote{\url{https://openaipublic.azureedge.net/gpt-2/detector-models/v1/detector-large.pt}}.

\paragraph{RoBERTa (ChatGPT)} \cite{guo2023close} This RoBERTa-base \cite{liu2019roberta} detector is fine-tuned on the HC3 dataset, which contains approximately 27,000 question-answer pairs, with answers generated by both humans and ChatGPT.  The questions span diverse domains, including Reddit, medicine, finance, and law.  We access the detector via HuggingFace Datasets: \texttt{Hello-SimpleAI/chatgpt-detector-roberta}.

\paragraph{RADAR} \cite{hu2023radar} This detector is a fine-tuned Vicuna 7B model (itself a fine-tune of LLaMA 7B), trained in a generative adversarial setting.  The training involved a paraphraser model designed to fool the detector, and the detector was trained to distinguish between paraphraser outputs, human-written text from the WebText dataset, and generations from the original language model. We access RADAR via HuggingFace: \texttt{TrustSafeAI/RADAR-Vicuna-7B}.

\paragraph{GLTR} \cite{gehrmann-etal-2019-gltr} Originally designed as an interactive tool to aid human detection of generated text, GLTR has become a standard baseline in detector robustness evaluations.  It analyzes the likelihood of text under a language model, binning tokens based on their predicted probabilities. These bins then serve as features for detection. We use the default GLTR settings\footnote{\url{https://github.com/HendrikStrobelt/detecting-fake-text}}: a cutoff rank of 10 and GPT-2 small as the language model.

\paragraph{DetectGPT} \cite{mitchell-etal-2023-detectgpt} This zero-shot detector leverages the observation that LM-generated text often resides in regions of negative curvature within the model's log probability function. DetectGPT compares the log probability of an input text, computed by the target LM, to the average log probability of slightly perturbed versions of the text (generated using a separate masked language model like T5). A significant drop in log probability for the perturbed text indicates a higher likelihood of machine generation.

\paragraph{FastDetectGPT} \cite{bao2023fastdetectgpt} This detector is an optimized version of DetectGPT \cite{mitchell-etal-2023-detectgpt}, achieving a 340x speedup without sacrificing accuracy.  Following the original implementation, we use GPT-Neo-2.7B as the scoring model and GPT-J-6B as the reference model for generating perturbations.  Neither of these models was used to generate the continuations in our dataset.

\paragraph{FastDetectLLM}  Referenced as \texttt{fastdetectllm} in the RAID benchmark code \citep{dugan2024raidsharedbenchmarkrobust}, this zero-shot detector is conceptually related to FastDetectGPT \cite{mitchell-etal-2023-detectgpt}.  FastDetectLLM directly uses the average log-rank of input tokens, predicted by a scoring language model (default: GPT-Neo-2.7B), as its detection metric. Lower average log-ranks suggest a higher likelihood of machine generation. This approach bypasses FastDetectGPT's perturbation and sampling steps, significantly improving speed.

\paragraph{Binoculars} \cite{hans2024spotting} This detector uses the ratio of perplexity to cross-entropy between two similar language models as its detection metric. We use the official code and default models (Falcon 7B and Falcon 7B Instruct \cite{almazrouei2023falcon}), which, like FastDetectGPT, were not used for text generation in our dataset.

\paragraph{LLMDet} \cite{wu-etal-2023-llmdet} This detector uses the "proxy-perplexity" from 10 small language models as features. Proxy-perplexity approximates true perplexity by sampling n-grams, avoiding full model execution. None of the models used for proxy-perplexity calculation were involved in generating text for our dataset.
\section{Additional Statistical Tests}
\label{sec:appendix-tests}

To further explore whether potential bias arises under different conditions, we implement three complementary strategies:  (1) a \textbf{single-factor analysis} that does not control for confounding variables, (2) a \textbf{matched subset analysis} that fixes control features, and (3) a \textbf{down-sampled matched analysis} that enforces balanced sample sizes.  \footnote{We note that, due to the complete coverage of all attribute combinations in our dataset, the single-factor and matched-subset approaches yield identical results in this specific case; however, we describe both for conceptual clarity.}  Below, we describe each data manipulation strategy, followed by a description of the statistical tests applied to the resulting datasets.

\subsection{Data Subsetting and Sampling Strategies}

\paragraph{(1) Single-Factor (No Confound Control).}
In the simplest approach, we do not explicitly control for other attributes. We group the data by the \emph{focal} attribute (e.g., \emph{gender} = Female vs.\ Male) and compare detection outcomes.  This is a baseline approach that is straightforward but may be susceptible to confounding.

\paragraph{(2) Matched Subset (Fixing Control Features).}
In a more controlled approach, we fix all \emph{other} (control) attributes so that \emph{only} the \emph{main feature} varies among samples. Suppose our main feature has values \(\{f_1, f_2, \dots, f_n\}\). For each unique combination of the control features (e.g., \textit{CEFR} = B2, \textit{environment} = ESL), we identify those samples that share exactly those control-feature values but differ \emph{exclusively} in the main feature. We keep only the combinations that contain \emph{all} categories \(\{f_1,\dots,f_n\}\). Concatenating these subsets yields a matched dataset where any difference in detector outcomes is more plausibly attributed to the main feature rather than being confounded by other factors. Algorithm~\ref{alg:matched-subset} shows how we construct the matched subset.

Mathematically, let
\[
  \mathcal{C} = \{ c_1, \dots, c_r \}
  \quad\text{and}\quad
  \Omega = \bigtimes_{j=1}^r \mathrm{Levels}(c_j).
\]
For each combination \(\boldsymbol{\omega} \in \Omega\), we form the subset 
\[
  S_{\boldsymbol{\omega}}
  \;=\;
  \bigl\{
    x \in D \;\mid\; x[c_j] = \omega_j,\;\forall c_j \in \mathcal{C}
  \bigr\}.
\]
If \(S_{\boldsymbol{\omega}}\) spans all main feature categories, we retain it. We then unify these retained subsets to form 
\[
  \mathrm{Matched}
  \;=\;
  \bigcup_{\boldsymbol{\omega}\,\in\,\Omega :\, \{\!f_1,\dots,f_n\!\}\subseteq \text{MainFeature}(S_{\boldsymbol{\omega}})}
  S_{\boldsymbol{\omega}}.
\]

\begin{algorithm}[!h]
\renewcommand{\algorithmicrequire}{\textbf{Input:}}
\renewcommand{\algorithmicensure}{\textbf{Output:}}
\begin{algorithmic}[1]
    \REQUIRE Dataset $D$, main feature $F$ with categories $\{f_1,\ldots,f_n\}$, control features $\mathcal{C} = \{c_1,\dots,c_r\}$
    \ENSURE A subset $M$ where only $F$ varies in each control-feature combo
    \STATE $M \leftarrow \emptyset$
    \STATE $\Omega \leftarrow \mathrm{CartProd}\bigl(\mathrm{Levels}(c_1), \dots, \mathrm{Levels}(c_r)\bigr)$
    \FOR{each $\boldsymbol{\omega} \in \Omega$}
        \STATE $S_{\boldsymbol{\omega}} \leftarrow \bigl\{\,x \in D \;\mid\;\forall j,\, x[c_j] = \omega_j \bigr\}$
        \STATE $\mathcal{F}_{\boldsymbol{\omega}} \leftarrow \{\text{main feature value for } x \in S_{\boldsymbol{\omega}}\}$
        \IF{$\{f_1,\dots,f_n\} \subseteq \mathcal{F}_{\boldsymbol{\omega}}$}
            \STATE $M \leftarrow M \,\cup\, S_{\boldsymbol{\omega}}$
        \ENDIF
    \ENDFOR
    \RETURN $M$
\end{algorithmic}
\caption{Matched Subset (Fix Control Features)}
\label{alg:matched-subset}
\end{algorithm}

\paragraph{(3) Down-Sampled Matched (One-to-One Matching).}
Even if the matched dataset includes all main-feature categories, different groups may still have \emph{unequal} sample sizes. In this one-to-one variant, we require that for every record in the \emph{smallest} category (within each combination of control features), there is exactly \emph{one} matched record in every other category. Concretely, for each record in the smallest category, we randomly select \emph{one} record from each larger category that shares the same control-feature values. This ensures balanced sample sizes \emph{and} enforces a strict one-to-one matching across categories, though it may discard additional data from larger categories. Algorithm~\ref{alg:downsample-one-to-one} shows how to downsample to create matched data.

Formally, let
\[
   \Omega \;=\; \mathrm{CartProd}\bigl(\mathrm{Levels}(c_1), \dots, \mathrm{Levels}(c_r)\bigr),
\]
be the Cartesian product of all levels of control features $C = \{c_1, \dots, c_r\}$. For each $\boldsymbol{\omega} \in \Omega$, define
\[
   S_{\boldsymbol{\omega}}
   \;=\;
   \bigl\{\,
     x \in D
     \;\mid\;
     \forall c_j \in C,\;\;
     x[c_j] = \omega_j
   \bigr\},
\]
the subset of $D$ whose rows match the control-feature values in $\boldsymbol{\omega}$. Let $f_{k^*}$ be the category of the main feature $F$ with the fewest samples in $S_{\boldsymbol{\omega}}$. For each $x \in S_{\boldsymbol{\omega}}$ with $x[F] = f_{k^*}$, define a matched set

\[
\begin{aligned}
S_x 
  &= \{x\} 
     \;\cup\;
     \bigcup_{\substack{f_i \in \{f_1,\dots,f_n\} \\ f_i \neq f_{k^*}}}
     \Bigl\{\, 
       y \in S_{\boldsymbol{\omega}}
       \;\big|\;
       y[F] = f_i, \\[6pt]
  &\quad\quad
       \text{$y$ is sampled exactly once}
     \Bigr\}.
\end{aligned}
\]
where we pick exactly one such $y$ randomly from the available rows each time (removing it to prevent re-use).  
The overall matched subset for $\boldsymbol{\omega}$ is then
\[
   \tilde{S}_{\boldsymbol{\omega}}
   \;=\;
   \bigcup_{\,x : x[F] = f_{k^*}}
   S_x,
\]
and the final down-sampled matched dataset is
\[
   M
   \;=\;
   \bigcup_{\,\boldsymbol{\omega} \,\in\, \Omega}
   \tilde{S}_{\boldsymbol{\omega}}.
\]
This matching enforces that each record in the smallest category is paired one-to-one with exactly one record from each of the other categories.

\begin{algorithm}[!h]
\renewcommand{\algorithmicrequire}{\textbf{Input:}}
\renewcommand{\algorithmicensure}{\textbf{Output:}}
\begin{algorithmic}[1]
    \REQUIRE Dataset $D$, main feature $F$ with categories $\{f_1,\dots,f_n\}$, control features $C = \{c_1,\dots,c_r\}$
    \ENSURE A balanced, matched dataset $M$ where each sample in the smallest category is paired one-to-one with exactly one sample in every other category, for the same control-feature values
    \STATE $M \leftarrow \emptyset$
    \STATE $\Omega \leftarrow \mathrm{CartProd}\bigl(\mathrm{Levels}(c_1), \dots, \mathrm{Levels}(c_r)\bigr)$
    \FOR{each $\boldsymbol{\omega} \in \Omega$}
        \STATE $S_{\boldsymbol{\omega}} \leftarrow \bigl\{\,x \in D \;\mid\;\forall j,\, x[c_j] = \omega_j \bigr\}$
        \STATE Let $\{f_{k^*}\}$ be the category in $S_{\boldsymbol{\omega}}$ with the fewest samples (the ``smallest'' category)
        \FOR{each $x \in S_{\boldsymbol{\omega}}$ such that $x[F] = f_{k^*}$}
            \STATE $S_{x} \leftarrow \{x\}$ \COMMENT{Start a matched set with the sample from the smallest category}
            \FOR{each $f_i \in \{f_1,\dots,f_n\} \setminus \{f_{k^*}\}$}
                \STATE From $S_{\boldsymbol{\omega}}$ with $F = f_i$, randomly pick exactly one sample $y$ (if any remain)
                \STATE $S_{x} \leftarrow S_{x} \cup \{y\}$
                \STATE Remove $y$ from $S_{\boldsymbol{\omega}}$ to prevent re-use
            \ENDFOR
            \STATE $M \leftarrow M \cup S_{x}$
        \ENDFOR
    \ENDFOR
    \RETURN $M$
\end{algorithmic}
    \caption{Down-Sampled Matched Data Construction (One-to-One Matching)}
\label{alg:downsample-one-to-one}
\end{algorithm}

\subsection{Statistical Tests}
\label{sec:stats-focal-attr}

After applying one of the data strategies above (single-factor, matched subset, or down-sampled matched), we perform hypothesis tests to assess the significance of the focal attribute.  We use weighted statistical tests to account for the varying number of texts within each group.

\paragraph{Binary Attributes (Two Groups).}
If the focal attribute has exactly two categories (e.g., \emph{gender}: Female vs. Male), we employ Welch’s two-sample \emph{t}-test \citep{student1908probable}, modified to incorporate weights. The weighted t-test, with unequal variances, is calculated as follows.

Let \(w_i\) be the weight, and let \(X_i\) be its detection accuracy of each observation. Let two groups be \(G_1\) and \(G_2\)

The weighted means:
\[
\bar{X}_k = \frac{\sum_{i \in G_k} w_i X_i}{\sum_{i \in G_k} w_i}, \quad k \in \{1,2\}.
\]

Weighted variances:
\[
\widehat{\sigma}_k^2 = \frac{\sum_{i \in G_k} w_i (X_i - \bar{X}_k)^2}{\sum_{i \in G_k} w_i}, \quad k \in \{1, 2\}.
\]

Total weights per group:
\[
W_k = \sum_{i \in G_k} w_i, \quad k \in \{1, 2\}.
\]

Weighted Welch's t-statistic:
\[
t = \frac{\bar{X}_1 - \bar{X}_2}{\sqrt{\frac{\widehat{\sigma}^2_1}{W_1} + \frac{\widehat{\sigma}^2_2}{W_2}}}.
\]

Approximate degrees of freedom:
\[
\nu \approx \frac{\left(\frac{\widehat{\sigma}^2_1}{W_1} + \frac{\widehat{\sigma}^2_2}{W_2}\right)^2}
{\frac{\left(\frac{\widehat{\sigma}^2_1}{W_1}\right)^2}{n_1 - 1} + \frac{\left(\frac{\widehat{\sigma}^2_2}{W_2}\right)^2}{n_2 - 1}},
\]
where \(n_1, n_2\) denote the (unweighted) sample counts in each group.

We compare the resulting \(p\)-value to \(\alpha\) (e.g., 0.05) to assess significance.

\paragraph{Multi-Class Attributes (More Than Two Groups).}
When the focal attribute has more than two categories, we use a weighted one-way ANOVA\citep{fisher1925statistical}, implemented via a Weighted Least Squares (WLS) regression. The model is of the form:
\[
\text{accuracy}_i \sim C(\text{focal\_attribute})
\]
with weights \(w_i\). We fit this model using `statsmodels` and then perform a Type II ANOVA. The relevant \(F\)-statistic is:
\[
F = \frac{\text{MS}_{\text{Between}}}{\text{MS}_{\text{Within}}}
\]
A significant \(p\)-value indicates that at least one category differs from the others.

\section{Detailed Results}

\begin{table*}[ht]
    \centering
    \small
    \begin{tabular}{lcccccccc}
        \toprule
        \multirow{2}{*}{\textbf{Detector}} & 
        \multicolumn{2}{c}{\textbf{CEFR}} & 
        \multicolumn{2}{c}{\textbf{Sex}} & 
        \multicolumn{2}{c}{\textbf{Academic Genre}} & 
        \multicolumn{2}{c}{\textbf{Language Env.}} \\
        \cmidrule(lr){2-3}
        \cmidrule(lr){4-5}
        \cmidrule(lr){6-7}
        \cmidrule(lr){8-9}
        & $p$-value & Sig.? & $p$-value & Sig.? & $p$-value & Sig.? & $p$-value & Sig.? \\
        \midrule
        binoculars      & $1.3745\times 10^{-15}$ & Yes & 0.80451 & No  & 0.81889 & No  & $1.6875\times 10^{-17}$ & Yes \\
        chatgpt-roberta & 0.036675                & Yes & 0.61647 & No  & 0.41117 & No  & 0.79088                & No  \\
        detectgpt       & $3.4682\times 10^{-7}$  & Yes & 0.17331 & No  & 0.001146 & Yes & 0.0067062              & Yes \\
        fastdetectgpt   & $1.7303\times 10^{-5}$  & Yes & 0.95033 & No  & 0.22113  & No  & 0.0023632              & Yes \\
        fastdetectllm   & $2.3643\times 10^{-13}$ & Yes & 0.14879 & No  & $2.5190\times 10^{-6}$ & Yes & $1.0241\times 10^{-7}$  & Yes \\
        gltr            & $5.0894\times 10^{-17}$ & Yes & 0.37228 & No  & 0.79752  & No  & $4.2131\times 10^{-17}$ & Yes \\
        gpt2-base       & $3.5869\times 10^{-37}$ & Yes & 0.47449 & No  & $2.6296\times 10^{-10}$ & Yes & $1.4756\times 10^{-25}$ & Yes \\
        gpt2-large      & $1.7213\times 10^{-4}$  & Yes & 0.70402 & No  & $2.8182\times 10^{-7}$  & Yes & 0.030228               & Yes \\
        llmdet          & $1.1682\times 10^{-38}$ & Yes & 0.07170 & No  & $2.7679\times 10^{-4}$  & Yes & $8.4468\times 10^{-15}$ & Yes \\
        radar           & $2.0507\times 10^{-5}$  & Yes & 0.77951 & No  & 0.12957  & No  & 0.010882               & Yes \\
        \bottomrule
    \end{tabular}
    \caption{WLS results: ANOVA $p$-values for each factor across detectors. ``Yes'' indicates $p<0.05$ (significant), ``No'' otherwise.}
    \label{tab:bias_table}
\end{table*}

\begin{table*}[ht!]
\centering
\small
\begin{tabular}{lllllll}
\toprule
\textbf{Factor} & \textbf{Detector} & \textbf{Comparison} & \textbf{WaldStat} & \textbf{raw\_p} & \textbf{p\_corr} & \textbf{Sig.?} \\
\midrule
\multirow{9}{*}{CEFR} & \multirow{4}{*}{binoculars}
 & B2\_0 vs XX\_0 & 36.1784 & 2.7460e-09 & 1.9222e-08 & Yes \\
 &  & A2\_0 vs XX\_0 & 72.4739 & 8.4083e-17 & 8.4083e-16 & Yes \\
 &  & B1\_2 vs XX\_0 & 49.9436 & 3.4660e-12 & 3.1194e-11 & Yes \\
 &  & B1\_1 vs XX\_0 & 48.0481 & 8.6122e-12 & 6.8897e-11 & Yes \\
\cmidrule{2-7}
 & \multirow{2}{*}{chatgpt-roberta}
 & A2\_0 vs B2\_0 & 9.4951 & 2.1314e-03 & 2.1314e-02 & Yes \\
 &  & B1\_1 vs B2\_0 & 8.0170 & 4.7512e-03 & 4.2761e-02 & Yes \\
\cmidrule{2-7}
 & \multirow{3}{*}{detectgpt}
 & B2\_0 vs A2\_0 & 11.6703 & 6.6733e-04 & 4.0040e-03 & Yes \\
 &  & B2\_0 vs XX\_0 & 25.2118 & 6.3457e-07 & 5.7111e-06 & Yes \\
\midrule
\multirow{4}{*}{Language Env.} & \multirow{3}{*}{binoculars}
 & EFL vs ESL & 8.8010 & 3.1010e-03 & 3.1010e-03 & Yes \\
 &  & EFL vs NS  & 72.4739 & 8.4083e-17 & 2.5225e-16 & Yes \\
 &  & ESL vs NS  & 21.8296 & 3.4990e-06 & 6.9980e-06 & Yes \\
\cmidrule{2-7}
 & \multirow{1}{*}{detectgpt}
 & EFL vs NS & 8.5938 & 3.4700e-03 & 1.0410e-02 & Yes \\
 \midrule

\multirow{10}{*}{Academic Genre} & \multirow{3}{*}{detectgpt}
 & Humanities vs Sci\&Tech & 8.8179 & 3.0727e-03 & 1.5363e-02 & Yes \\
 &  & Humanities vs Life Sci  & 9.3496 & 2.3052e-03 & 1.3831e-02 & Yes \\
 &  & Social Sci vs Life Sci  & 6.7344 & 9.6317e-03 & 3.8527e-02 & Yes \\
\cmidrule{2-7}
 & \multirow{2}{*}{gpt2-base}
 & Sci\&Tech vs Humanities & 24.8324 & 7.6795e-07 & 3.8398e-06 & Yes \\
 &  & Sci\&Tech vs Social Sci & 41.8589 & 1.7148e-10 & 1.0289e-09 & Yes \\
\cmidrule{2-7}
 & \multirow{2}{*}{gpt2-large}
 & Sci\&Tech vs Humanities & 12.1460 & 5.1890e-04 & 2.5945e-03 & Yes \\
 &  & Sci\&Tech vs Social Sci & 32.5064 & 1.6753e-08 & 1.0052e-07 & Yes \\
\cmidrule{2-7}
 & \multirow{3}{*}{llmdet}
 & Life Sci vs Sci\&Tech & 11.0328 & 9.3605e-04 & 3.7442e-03 & Yes \\
 &  & Life Sci vs Social Sci & 12.6490 & 3.9804e-04 & 1.9902e-03 & Yes \\
 &  & Life Sci vs Humanities & 18.2247 & 2.2005e-05 & 1.3203e-04 & Yes \\
\bottomrule
\end{tabular}
\caption{WLS results: Post-hoc pairwise comparisons (Wald test, Holm correction) for CEFR Level, Language Environment, and Academic Genre.  Only showing significant pairs for detectors where the overall ANOVA for the corresponding factor was significant (see Table~\ref{tab:bias_table}).}
\label{tab:posthoc_combined}
\end{table*}

\section{Additional Results}

\begin{table*}[ht!]
\centering
\small
\begin{tabular}{llcccc}
\toprule
\textbf{Factor} & \textbf{Detector} & \textbf{Statistic} & \textbf{p-value} & \textbf{Sig?} & \textbf{Ranking (LS-means)} \\
\midrule
\multirow{10}{*}{CEFR} & binoculars & F = 19.0332 & 5.78e-15 & Yes & A2\_0 > B2\_0 > B1\_1 > B1\_2 > XX\_0 \\
 & chatgpt-roberta & F = 2.9443 & 1.97e-02 & Yes & A2\_0 > B1\_1 > B1\_2 > XX\_0 > B2\_0 \\
 & detectgpt & F = 9.3006 & 2.39e-07 & Yes & B2\_0 > B1\_2 > B1\_1 > A2\_0 > XX\_0 \\
 & fastdetectgpt & F = 7.4965 & 5.56e-06 & Yes & B2\_0 > B1\_2 > A2\_0 > B1\_1 > XX\_0 \\
 & fastdetectllm & F = 21.9696 & 3.19e-17 & Yes & XX\_0 > A2\_0 > B1\_1 > B2\_0 > B1\_2 \\
 & gltr & F = 21.8672 & 3.82e-17 & Yes & A2\_0 > B1\_1 > B1\_2 > B2\_0 > XX\_0 \\
 & gpt2-base & F = 50.8631 & 3.60e-38 & Yes & A2\_0 > B1\_1 > B1\_2 > B2\_0 > XX\_0 \\
 & gpt2-large & F = 6.0261 & 8.85e-05 & Yes & B1\_1 > B2\_0 > B1\_2 > A2\_0 > XX\_0 \\
 & llmdet & F = 35.2707 & 3.74e-27 & Yes & A2\_0 > XX\_0 > B1\_1 > B1\_2 > B2\_0 \\
 & radar & F = 6.6219 & 3.03e-05 & Yes & B1\_1 > B1\_2 > B2\_0 > A2\_0 > XX\_0 \\
\midrule
\multirow{10}{*}{Sex} & binoculars & t = -4.2320 & 2.32e-05 & Yes & F: 0.9412, M: 0.9425 (M > F) \\
 & chatgpt-roberta & t = -10.3064 & 6.76e-25 & Yes & F: 0.7433, M: 0.7489 (M > F) \\
 & detectgpt & t = 28.7863 & 1.39e-181 & Yes & F: 0.8000, M: 0.7885 (F > M) \\
 & fastdetectgpt & t = 7.9967 & 1.28e-15 & Yes & F: 0.8986, M: 0.8961 (F > M) \\
 & fastdetectllm & t = -12.2473 & 1.83e-34 & Yes & F: 0.4793, M: 0.4806 (M > F) \\
 & gltr & t = 7.2602 & 3.89e-13 & Yes & F: 0.8190, M: 0.8138 (F > M) \\
 & gpt2-base & t = -16.7136 & 1.24e-62 & Yes & F: 0.6193, M: 0.6282 (M > F) \\
 & gpt2-large & t = -11.7789 & 5.23e-32 & Yes & F: 0.5558, M: 0.5623 (M > F) \\
 & llmdet & t = -32.8895 & 3.45e-236 & Yes & F: 0.4968, M: 0.5020 (M > F) \\
 & radar & t = -0.9671 & 3.34e-01 & No & F vs. M not significant \\
\midrule
\multirow{10}{*}{Academic Genre} & binoculars & F = 0.5213 & 6.68e-01 & No & -- \\
 & chatgpt-roberta & F = 1.1469 & 3.29e-01 & No & -- \\
 & detectgpt & F = 6.4159 & 2.69e-04 & Yes & Humanities > Social Sci > Sci\&Tech > Life Sci \\
 & fastdetectgpt & F = 1.8218 & 1.41e-01 & No & -- \\
 & fastdetectllm & F = 8.1352 & 2.43e-05 & Yes & Humanities > Social Sci > Sci\&Tech > Life Sci \\
 & gltr & F = 0.2925 & 8.31e-01 & No & -- \\
 & gpt2-base & F = 14.7583 & 2.32e-09 & Yes & Sci\&Tech > Life Sci > Humanities > Social Sci \\
 & gpt2-large & F = 12.1231 & 9.15e-08 & Yes & Sci\&Tech > Life Sci > Humanities > Social Sci \\
 & llmdet & F = 7.7058 & 4.44e-05 & Yes & Life Sci > Sci\&Tech > Social Sci > Humanities \\
 & radar & F = 2.4177 & 6.51e-02 & No & -- \\
\midrule
\multirow{10}{*}{Language Env.} & binoculars & F = 42.2609 & 3.56e-18 & Yes & EFL > ESL > NS \\
 & chatgpt-roberta & F = 0.3534 & 7.02e-01 & No & -- \\
 & detectgpt & F = 9.1837 & 1.14e-04 & Yes & ESL > EFL > NS \\
 & fastdetectgpt & F = 7.2639 & 7.24e-04 & Yes & ESL > EFL > NS \\
 & fastdetectllm & F = 33.2769 & 1.31e-14 & Yes & NS > EFL > ESL \\
 & gltr & F = 40.7305 & 1.42e-17 & Yes & EFL > ESL > NS \\
 & gpt2-base & F = 49.4957 & 5.40e-21 & Yes & EFL > ESL > NS \\
 & gpt2-large & F = 4.6056 & 1.03e-02 & Yes & ESL > EFL > NS \\
 & llmdet & F = 6.2017 & 2.12e-03 & Yes & NS > ESL > EFL \\
 & radar & F = 7.1878 & 8.05e-04 & Yes & EFL > ESL > NS \\
\bottomrule
\end{tabular}
\caption{Single Factor and Matched Subset Results: Weighted ANOVA results for CEFR, Academic Genre, and Language Environment; Weighted T-test results for Sex.  "Statistic" indicates the test statistic (F for ANOVA, t for T-test). "Sig?" indicates omnibus significance. "Ranking" shows LS-means. A ``--'' indicates that the overall ANOVA for that factor and detector was not statistically significant, so LSMeans are not reported.}
\label{tab:combined_results}
\end{table*}

\begin{table*}[ht!]
\centering
\small
\resizebox{\textwidth}{!}{%
\begin{tabular}{lcccccccc}
\toprule
\textbf{Detector} & \textbf{N} & \textbf{F} & \textbf{p} & \textbf{Sig?} 
& \textbf{LS Mean (Hum)} & \textbf{LS Mean (SS)} & \textbf{LS Mean (ST)} 
& \textbf{LS Mean (LSci)} \\
\midrule
\textbf{binoculars}      & 624  & 0.4983  & 0.684  & No  &  --     &  --     &  --     & --      \\
\textbf{chatgpt-roberta} & 624  & 1.0527  & 0.369  & No  &  --     &  --     &  --     & --      \\
\textbf{detectgpt}       & 624  & 4.5679  & 0.0036 & Yes & 0.8066  & 0.7979  & 0.7857  & 0.7679  \\
\textbf{fastdetectgpt}   & 1248 & 1.2188  & 0.302  & No  &  --     &  --     &  --     & --      \\
\textbf{fastdetectllm}   & 624  & 6.3876  & 2.88e-04 & Yes & 0.4834  & 0.4813  & 0.4778  & 0.4697  \\
\textbf{gltr}            & 624  & 0.2762  & 0.843  & No  &  --     &  --     &  --     & --      \\
\textbf{gpt2-base}       & 624  & 11.3682 & 2.89e-07 & Yes & 0.6074  & 0.6011  & 0.6474  & 0.6216  \\
\textbf{gpt2-large}      & 624  & 10.3231 & 1.23e-06 & Yes & 0.5505  & 0.5310  & 0.5810  & 0.5568  \\
\textbf{llmdet}          & 624  & 5.9057  & 5.62e-04 & Yes & 0.4918  & 0.4995  & 0.5009  & 0.5109  \\
\textbf{radar}           & 624  & 1.9868  & 0.115  & No  &  --     &  --     &  --     & --      \\
\bottomrule
\end{tabular}}
\caption{Down-Sample Matched Results: Weighted ANOVA on ``Academic Genre'' for each detector. "Sig?" refers to overall significance of the ANOVA ($p<0.05$). Only detectors with \textbf{Yes} in the “Sig?” column have meaningful LS Means 
and potential post-hoc comparisons. A ``--'' indicates that the overall ANOVA for that factor and detector was not statistically significant, so LSMeans are not reported.}
\label{tab:genre-anova}
\end{table*}

\begin{table*}[ht!]
\centering
\small
\begin{tabular}{lcccccccc}
\toprule
\textbf{Detector} & \textbf{N} & \textbf{t} & \textbf{p} & \textbf{df} 
& \textbf{Sig?} & \textbf{LS Mean (M)} & \textbf{LS Mean (F)} 
& \textbf{Post-hoc p} \\
\midrule
\textbf{binoculars}      &  792 & -4.7853  & 1.71e-06   & 116529.44 & Yes & 0.9425 & 0.9410 & 0.701  \\
\textbf{chatgpt-roberta} &  792 & -10.3512 & 4.24e-25   & 114663.56 & Yes & 0.7489 & 0.7433 & 0.405  \\
\textbf{detectgpt}       &  792 & 29.4391  & 8.78e-190  & 115554.43 & Yes & 0.7885 & 0.8002 & 0.018\textsuperscript{*} \\
\textbf{fastdetectgpt}   & 1584 & 7.6643   & 1.81e-14   & 221280.83 & Yes & 0.8961 & 0.8985 & 0.534  \\
\textbf{fastdetectllm}   &  792 & -13.2158 & 7.60e-40   & 116208.70 & Yes & 0.4806 & 0.4792 & 0.289  \\
\textbf{gltr}            &  792 & 7.5210   & 5.48e-14   & 113707.20 & Yes & 0.8138 & 0.8192 & 0.544  \\
\textbf{gpt2-base}       &  792 & -16.5020 & 4.19e-61   & 112626.50 & Yes & 0.6282 & 0.6194 & 0.182  \\
\textbf{gpt2-large}      &  792 & -11.9331 & 8.32e-33   & 115088.05 & Yes & 0.5623 & 0.5557 & 0.337  \\
\textbf{llmdet}          &  792 & -31.0913 & 2.18e-211  & 120438.22 & Yes & 0.5020 & 0.4971 & 0.014\textsuperscript{*} \\
\textbf{radar}           &  792 & -1.0418  & 0.298      & 118146.32 & No  &  --    &  --    &  --    \\
\bottomrule
\end{tabular}
\caption{Down-Sample Matched Results: Weighted T-tests on ``Sex'' (M vs F) for each detector.“Sig?” refers to overall significance of the T-test ($p<0.05$). An asterisk (\textsuperscript{*}) in “Post-hoc p” indicates a significant difference (Holm-adjusted) specifically for M vs F. A ``--'' indicates that the overall T-test for that factor and detector was not statistically significant, so LSMeans are not reported.}
\label{tab:sex-test}
\end{table*}

\begin{table*}[ht!]
\centering
\small
\begin{tabular}{llll}
\toprule
\textbf{Factor} & \textbf{Detector} & \textbf{Comparison} & \textbf{Sig.?} \\
\midrule
\multirow{10}{*}{Acad. Genre} & \multirow{2}{*}{detectgpt}
 & Hum. vs S.\&T. & Yes \\
 &  & Hum. vs Life S.  & Yes \\
\cmidrule(lr){2-4}
 & \multirow{3}{*}{fastdetectllm}
 & Hum. vs S.\&T. & Yes \\
 &  & Hum. vs Life S.  & Yes \\
 & & Soc. S. vs Life S. & Yes \\
 \cmidrule(lr){2-4}
 & \multirow{2}{*}{gpt2-base}
 & S.\&T. vs Hum. & Yes \\
 &  & S.\&T. vs Soc. S. & Yes \\
\cmidrule(lr){2-4}
 & \multirow{2}{*}{gpt2-large}
 & S.\&T. vs Hum. & Yes \\
 &  & S.\&T. vs Soc. S. & Yes \\
\cmidrule(lr){2-4}
 & \multirow{3}{*}{llmdet}
 & Life S. vs Hum. & Yes \\
 &  & S.\&T. vs Hum. & Yes \\
 &  & Soc. S. vs Hum. & Yes \\

\bottomrule
\end{tabular}
\caption{Down-Sample Matched Results: Significant post-hoc pairs (Holm-adjusted) among academic genres. "Hum.", "Soc. S.", "S.\&T.", "Life S." abbreviate Humanities, Social Sciences, Sciences \& Technology, and Life Sciences, respectively. Only detectors that showed a significant overall ANOVA in Table~\ref{tab:genre-anova} are included.}
\label{tab:genre-posthoc}
\end{table*}

\section{Detailed Subset Results}

\begin{table*}[ht!]
\centering
\resizebox{\textwidth}{!}{%
\begin{tabular}{lcccccccccc}
\toprule
\textbf{Generator Model} 
& \textbf{binoculars} 
& \textbf{fastdetectgpt} 
& \textbf{detectgpt} 
& \textbf{gltr} 
& \textbf{chatgpt-roberta} 
& \textbf{radar} 
& \textbf{gpt2-base} 
& \textbf{gpt2-large} 
& \textbf{llmdet} 
& \textbf{fastdetectllm} \\
\midrule
\textbf{Qwen2.5-72B-Instruct} & 0.959 & 0.967 & 0.897 & 0.939 & 0.808 & 0.822 & 0.653 & 0.531 & 0.496 & 0.479 \\
\textbf{Qwen2.5-32B-Instruct} & 0.958 & 0.906 & 0.789 & 0.789 & 0.789 & 0.691 & 0.518 & 0.483 & 0.487 & 0.479 \\
\textbf{Qwen2.5-14B-Instruct} & 0.958 & 0.896 & 0.751 & 0.836 & 0.874 & 0.802 & 0.579 & 0.496 & 0.493 & 0.479 \\
\textbf{Qwen2.5-7B-Instruct} & 0.958 & 0.964 & 0.840 & 0.884 & 0.822 & 0.700 & 0.525 & 0.485 & 0.487 & 0.479 \\
\textbf{Qwen2.5-3B-Instruct} & 0.946 & 0.735 & 0.693 & 0.603 & 0.774 & 0.562 & 0.506 & 0.485 & 0.497 & 0.479 \\
\textbf{Qwen2.5-1.5B-Instruct} & 0.925 & 0.876 & 0.786 & 0.661 & 0.733 & 0.647 & 0.571 & 0.534 & 0.500 & 0.480 \\
\textbf{Qwen2.5-0.5B-Instruct} & 0.825 & 0.873 & 0.762 & 0.621 & 0.651 & 0.622 & 0.654 & 0.779 & 0.499 & 0.480 \\
\textbf{llama3.1-70b-instruct} & 0.930 & 0.905 & 0.802 & 0.871 & 0.759 & 0.721 & 0.637 & 0.540 & 0.531 & 0.483 \\
\textbf{llama3.1-8b-instruct} & 0.955 & 0.957 & 0.832 & 0.902 & 0.690 & 0.725 & 0.672 & 0.572 & 0.505 & 0.479 \\
\textbf{llama3.2-3b-instruct} & 0.924 & 0.907 & 0.776 & 0.831 & 0.649 & 0.728 & 0.635 & 0.559 & 0.519 & 0.483 \\
\textbf{llama3.2-1b-instruct} & 0.824 & 0.795 & 0.715 & 0.687 & 0.706 & 0.631 & 0.671 & 0.668 & 0.512 & 0.527 \\
\textbf{Mistral-Small-Instruct-2409} & 0.959 & 0.947 & 0.767 & 0.773 & 0.685 & 0.630 & 0.524 & 0.481 & 0.497 & 0.479 \\
\textbf{Average} & 0.927 & 0.894 & 0.784 & 0.783 & 0.745 & 0.690 & 0.595 & 0.551 & 0.502 & 0.484 \\
\bottomrule
\end{tabular}%
}
\caption{Overall Detector performance across various generator models without subsetting}
\label{tab:all}
\end{table*}

\begin{table*}[ht!]
\centering
\resizebox{\textwidth}{!}{%
\begin{tabular}{lcccccccccc}
\toprule
\textbf{Generator Model} 
& \textbf{binoculars}
& \textbf{fastdetectgpt}
& \textbf{detectgpt}
& \textbf{gltr}
& \textbf{chatgpt-roberta}
& \textbf{radar}
& \textbf{gpt2-base}
& \textbf{gpt2-large}
& \textbf{llmdet}
& \textbf{fastdetectllm} \\
\midrule
\textbf{Qwen2.5-72B-Instruct} & 0.990 & 0.965 & 0.990 & 0.892 & 0.866 & 0.834 & 0.814 & 0.588 & 0.511 & 0.488 \\
\textbf{Qwen2.5-32B-Instruct} & 0.990 & 0.915 & 0.930 & 0.801 & 0.858 & 0.759 & 0.620 & 0.480 & 0.497 & 0.488 \\
\textbf{Qwen2.5-14B-Instruct} & 0.989 & 0.918 & 0.933 & 0.781 & 0.940 & 0.814 & 0.691 & 0.504 & 0.499 & 0.488 \\
\textbf{Qwen2.5-7B-Instruct} & 0.990 & 0.965 & 0.945 & 0.857 & 0.837 & 0.757 & 0.551 & 0.470 & 0.494 & 0.488 \\
\textbf{Qwen2.5-3B-Instruct} & 0.976 & 0.759 & 0.732 & 0.720 & 0.777 & 0.609 & 0.568 & 0.486 & 0.521 & 0.488 \\
\textbf{Qwen2.5-1.5B-Instruct} & 0.959 & 0.894 & 0.750 & 0.836 & 0.794 & 0.677 & 0.594 & 0.524 & 0.511 & 0.489 \\
\textbf{Qwen2.5-0.5B-Instruct} & 0.870 & 0.878 & 0.683 & 0.802 & 0.671 & 0.626 & 0.673 & 0.772 & 0.513 & 0.489 \\
\textbf{llama3.1-70B-instruct} & 0.944 & 0.856 & 0.860 & 0.743 & 0.651 & 0.610 & 0.674 & 0.552 & 0.5757 & 0.491 \\
\textbf{llama3.1-8B-instruct} & 0.985 & 0.955 & 0.954 & 0.836 & 0.674 & 0.716 & 0.784 & 0.665 & 0.526 & 0.489 \\
\textbf{llama3.2-3B-instruct} & 0.944 & 0.889 & 0.870 & 0.753 & 0.644 & 0.692 & 0.679 & 0.572 & 0.583 & 0.492 \\
\textbf{llama3.2-1B-instruct} & 0.839 & 0.765 & 0.720 & 0.707 & 0.692 & 0.660 & 0.702 & 0.696 & 0.532 & 0.543 \\
\textbf{Mistral-Small-Instruct-2409} & 0.990 & 0.935 & 0.780 & 0.739 & 0.682 & 0.577 & 0.621 & 0.471 & 0.519 & 0.488 \\
\textbf{Average} & 0.956 & 0.891 & 0.845 & 0.789 & 0.757 & 0.694 & 0.664 & 0.565 & 0.520 & 0.493 \\
\bottomrule
\end{tabular}%
}
\caption{Detector performance across various generator models under subgroup CEFR = A2-0}
\label{tab:cefr_A2-0}
\end{table*}

\begin{table*}[ht!]
\centering
\resizebox{\textwidth}{!}{%
\begin{tabular}{lcccccccccccc}
\toprule
\textbf{Generator Model} 
& \textbf{binoculars} 
& \textbf{fastdetectgpt} 
& \textbf{gltr} 
& \textbf{detectgpt} 
& \textbf{chatgpt-roberta} 
& \textbf{radar} 
& \textbf{gpt2-base} 
& \textbf{gpt2-large} 
& \textbf{llmdet} 
& \textbf{fastdetectllm} \\
\midrule
\textbf{Qwen2.5-72B-Instruct} & 0.981 & 0.976 & 0.987 & 0.932 & 0.843 & 0.909 & 0.774 & 0.572 & 0.502 & 0.475 \\
\textbf{Qwen2.5-32B-Instruct} & 0.981 & 0.942 & 0.891 & 0.817 & 0.818 & 0.789 & 0.558 & 0.488 & 0.483 & 0.475 \\
\textbf{Qwen2.5-14B-Instruct} & 0.981 & 0.908 & 0.925 & 0.786 & 0.926 & 0.849 & 0.658 & 0.514 & 0.503 & 0.475 \\
\textbf{Qwen2.5-7B-Instruct} & 0.981 & 0.973 & 0.956 & 0.867 & 0.856 & 0.805 & 0.605 & 0.500 & 0.485 & 0.475 \\
\textbf{Qwen2.5-3B-Instruct} & 0.952 & 0.701 & 0.679 & 0.719 & 0.797 & 0.583 & 0.551 & 0.492 & 0.507 & 0.475 \\
\textbf{Qwen2.5-1.5B-Instruct} & 0.952 & 0.879 & 0.737 & 0.803 & 0.782 & 0.682 & 0.619 & 0.558 & 0.505 & 0.476 \\
\textbf{Qwen2.5-0.5B-Instruct} & 0.850 & 0.888 & 0.670 & 0.811 & 0.670 & 0.643 & 0.674 & 0.788 & 0.493 & 0.475 \\
\textbf{llama3.1-70b-Instruct} & 0.922 & 0.862 & 0.872 & 0.751 & 0.682 & 0.668 & 0.692 & 0.585 & 0.560 & 0.482 \\
\textbf{llama3.1-8b-Instruct} & 0.974 & 0.961 & 0.951 & 0.827 & 0.665 & 0.734 & 0.781 & 0.639 & 0.523 & 0.475 \\
\textbf{llama3.2-3b-Instruct} & 0.916 & 0.868 & 0.809 & 0.713 & 0.589 & 0.687 & 0.684 & 0.595 & 0.542 & 0.483 \\
\textbf{llama3.2-1b-Instruct} & 0.832 & 0.781 & 0.710 & 0.722 & 0.682 & 0.661 & 0.692 & 0.697 & 0.524 & 0.513 \\
\textbf{Mistral-Small-Instruct-2409} & 0.981 & 0.957 & 0.851 & 0.825 & 0.704 & 0.678 & 0.567 & 0.484 & 0.505 & 0.475 \\
\textbf{Average} & 0.942 & 0.891 & 0.837 & 0.798 & 0.751 & 0.724 & 0.655 & 0.576 & 0.511 & 0.479 \\
\bottomrule
\end{tabular}%
}
\caption{Detector performance across various generator models under subgroup CEFR = B1-1}
\label{tab:cefr_B1-1}
\end{table*}

\begin{table*}[ht!]
\centering
\resizebox{\textwidth}{!}{%
\begin{tabular}{lcccccccccccc}
\toprule
\textbf{Generator Model} 
& \textbf{binoculars} 
& \textbf{fastdetectgpt} 
& \textbf{gltr} 
& \textbf{detectgpt} 
& \textbf{chatgpt-roberta} 
& \textbf{radar} 
& \textbf{gpt2-base} 
& \textbf{gpt2-large} 
& \textbf{llmdet} 
& \textbf{fastdetectllm} \\
\midrule
\textbf{Qwen2.5-72B-Instruct} & 0.974 & 0.981 & 0.982 & 0.913 & 0.812 & 0.870 & 0.658 & 0.521 & 0.496 & 0.466 \\
\textbf{Qwen2.5-32B-Instruct} & 0.974 & 0.929 & 0.843 & 0.811 & 0.800 & 0.669 & 0.514 & 0.488 & 0.484 & 0.466 \\
\textbf{Qwen2.5-14B-Instruct} & 0.974 & 0.895 & 0.857 & 0.763 & 0.895 & 0.801 & 0.574 & 0.497 & 0.496 & 0.466 \\
\textbf{Qwen2.5-7B-Instruct} & 0.974 & 0.979 & 0.921 & 0.854 & 0.827 & 0.729 & 0.526 & 0.491 & 0.488 & 0.466 \\
\textbf{Qwen2.5-3B-Instruct} & 0.967 & 0.725 & 0.604 & 0.711 & 0.770 & 0.547 & 0.506 & 0.488 & 0.485 & 0.466 \\
\textbf{Qwen2.5-1.5B-Instruct} & 0.943 & 0.886 & 0.690 & 0.787 & 0.717 & 0.643 & 0.574 & 0.526 & 0.495 & 0.466 \\
\textbf{Qwen2.5-0.5B-Instruct} & 0.838 & 0.889 & 0.649 & 0.774 & 0.647 & 0.624 & 0.652 & 0.772 & 0.493 & 0.468 \\
\textbf{Llama3.1-70b-Instruct} & 0.949 & 0.923 & 0.919 & 0.838 & 0.795 & 0.750 & 0.663 & 0.548 & 0.534 & 0.471 \\
\textbf{Llama3.1-8b-Instruct} & 0.971 & 0.972 & 0.954 & 0.852 & 0.697 & 0.761 & 0.679 & 0.563 & 0.506 & 0.466 \\
\textbf{Llama3.2-3b-Instruct} & 0.936 & 0.905 & 0.850 & 0.778 & 0.657 & 0.745 & 0.662 & 0.572 & 0.519 & 0.473 \\
\textbf{Llama3.2-1b-Instruct} & 0.852 & 0.822 & 0.715 & 0.743 & 0.724 & 0.623 & 0.678 & 0.671 & 0.509 & 0.516 \\
\textbf{Mistral-Small-Instruct-2409} & 0.974 & 0.964 & 0.841 & 0.788 & 0.685 & 0.657 & 0.523 & 0.488 & 0.492 & 0.466 \\
\textbf{Average} & 0.944 & 0.906 & 0.819 & 0.801 & 0.752 & 0.702 & 0.601 & 0.552 & 0.500 & 0.471 \\
\bottomrule
\end{tabular}%
}
\caption{Detector performance across various generator models under subgroup CEFR = B1-2}
\label{tab:cefr_B1-2}
\end{table*}

\begin{table*}[ht!]
\centering
\resizebox{\textwidth}{!}{%
\begin{tabular}{lcccccccccccc}
\toprule
\textbf{Generator Model} 
& \textbf{binoculars} 
& \textbf{fastdetectgpt} 
& \textbf{gltr} 
& \textbf{detectgpt} 
& \textbf{chatgpt-roberta} 
& \textbf{radar} 
& \textbf{gpt2-base} 
& \textbf{gpt2-large} 
& \textbf{llmdet} 
& \textbf{fastdetectllm} \\
\midrule
\textbf{Qwen2.5-72B-Instruct} & 0.961 & 0.980 & 0.950 & 0.902 & 0.705 & 0.781 & 0.538 & 0.507 & 0.467 & 0.474 \\
\textbf{Qwen2.5-32B-Instruct} & 0.960 & 0.899 & 0.716 & 0.794 & 0.693 & 0.632 & 0.490 & 0.500 & 0.467 & 0.474 \\
\textbf{Qwen2.5-14B-Instruct} & 0.961 & 0.900 & 0.807 & 0.744 & 0.818 & 0.794 & 0.531 & 0.500 & 0.468 & 0.474 \\
\textbf{Qwen2.5-7B-Instruct} & 0.961 & 0.970 & 0.873 & 0.816 & 0.780 & 0.635 & 0.513 & 0.501 & 0.470 & 0.474 \\
\textbf{Qwen2.5-3B-Instruct} & 0.956 & 0.800 & 0.590 & 0.667 & 0.753 & 0.548 & 0.496 & 0.500 & 0.468 & 0.474 \\
\textbf{Qwen2.5-1.5B-Instruct} & 0.936 & 0.897 & 0.653 & 0.775 & 0.684 & 0.629 & 0.559 & 0.544 & 0.476 & 0.474 \\
\textbf{Qwen2.5-0.5B-Instruct} & 0.854 & 0.889 & 0.647 & 0.752 & 0.642 & 0.623 & 0.682 & 0.806 & 0.481 & 0.474 \\
\textbf{Llama3.1-70b-Instruct} & 0.958 & 0.966 & 0.925 & 0.868 & 0.795 & 0.777 & 0.606 & 0.525 & 0.481 & 0.476 \\
\textbf{Llama3.1-8b-Instruct} & 0.960 & 0.971 & 0.903 & 0.860 & 0.697 & 0.743 & 0.607 & 0.523 & 0.469 & 0.474 \\
\textbf{Llama3.2-3b-Instruct} & 0.952 & 0.960 & 0.895 & 0.849 & 0.649 & 0.762 & 0.634 & 0.553 & 0.481 & 0.474 \\
\textbf{Llama3.2-1b-Instruct} & 0.848 & 0.837 & 0.736 & 0.729 & 0.705 & 0.634 & 0.692 & 0.679 & 0.486 & 0.523 \\
\textbf{Mistral-Small-Instruct-2409} & 0.961 & 0.956 & 0.776 & 0.771 & 0.663 & 0.628 & 0.505 & 0.500 & 0.470 & 0.474 \\
\textbf{Average} & 0.939 & 0.919 & 0.789 & 0.794 & 0.715 & 0.682 & 0.571 & 0.553 & 0.474 & 0.478 \\
\bottomrule
\end{tabular}%
}
\caption{Detector performance across various generator models under subgroup CEFR = B2-0}
\label{tab:cefr_B2-0}
\end{table*}

\begin{table*}[ht!]
\centering
\resizebox{\textwidth}{!}{
\begin{tabular}{lcccccccccccc}
\toprule
\textbf{Generator Model} 
& \textbf{binoculars} 
& \textbf{fastdetectgpt} 
& \textbf{gltr} 
& \textbf{detectgpt} 
& \textbf{chatgpt-roberta} 
& \textbf{radar} 
& \textbf{gpt2-base} 
& \textbf{gpt2-large} 
& \textbf{llmdet} 
& \textbf{fastdetectllm} \\
\midrule
\textbf{Qwen2.5-72B-Instruct} & 0.875 & 0.923 & 0.756 & 0.837 & 0.809 & 0.685 & 0.453 & 0.460 & 0.503 & 0.500 \\
\textbf{Qwen2.5-32B-Instruct} & 0.875 & 0.823 & 0.527 & 0.711 & 0.773 & 0.598 & 0.395 & 0.457 & 0.509 & 0.500 \\
\textbf{Qwen2.5-14B-Instruct} & 0.875 & 0.857 & 0.632 & 0.675 & 0.773 & 0.741 & 0.421 & 0.464 & 0.500 & 0.500 \\
\textbf{Qwen2.5-7B-Instruct} & 0.875 & 0.922 & 0.694 & 0.799 & 0.809 & 0.545 & 0.410 & 0.457 & 0.500 & 0.500 \\
\textbf{Qwen2.5-3B-Instruct} & 0.873 & 0.693 & 0.394 & 0.645 & 0.766 & 0.537 & 0.399 & 0.457 & 0.506 & 0.500 \\
\textbf{Qwen2.5-1.5B-Instruct} & 0.824 & 0.815 & 0.451 & 0.730 & 0.689 & 0.603 & 0.500 & 0.521 & 0.521 & 0.503 \\
\textbf{Qwen2.5-0.5B-Instruct} & 0.699 & 0.806 & 0.436 & 0.664 & 0.625 & 0.594 & 0.587 & 0.761 & 0.522 & 0.500 \\
\textbf{Llama3.1-70b-Instruct} & 0.873 & 0.915 & 0.755 & 0.810 & 0.877 & 0.804 & 0.522 & 0.476 & 0.503 & 0.502 \\
\textbf{Llama3.1-8b-Instruct} & 0.872 & 0.917 & 0.716 & 0.782 & 0.721 & 0.664 & 0.485 & 0.462 & 0.503 & 0.500 \\
\textbf{Llama3.2-3b-Instruct} & 0.870 & 0.913 & 0.726 & 0.797 & 0.716 & 0.759 & 0.488 & 0.491 & 0.509 & 0.500 \\
\textbf{Llama3.2-1b-Instruct} & 0.739 & 0.763 & 0.531 & 0.667 & 0.728 & 0.573 & 0.581 & 0.586 & 0.510 & 0.548 \\
\textbf{Mistral-Small-Instruct-2409} & 0.875 & 0.914 & 0.576 & 0.697 & 0.693 & 0.599 & 0.394 & 0.457 & 0.500 & 0.500 \\
\textbf{Average} & 0.844 & 0.855 & 0.600 & 0.735 & 0.748 & 0.642 & 0.470 & 0.504 & 0.507 & 0.504 \\
\bottomrule
\end{tabular}
}
\caption{Detector performance across various generator models under subgroup CEFR = XX-0}
\label{tab:cefr_XX-o}
\end{table*}

\begin{table*}[ht!]
\centering
\resizebox{\textwidth}{!}{%
\begin{tabular}{lcccccccccccc}
\toprule
\textbf{Generator Model} 
& \textbf{binoculars} 
& \textbf{fastdetectgpt} 
& \textbf{gltr} 
& \textbf{detectgpt} 
& \textbf{chatgpt-roberta} 
& \textbf{radar} 
& \textbf{gpt2-base} 
& \textbf{gpt2-large} 
& \textbf{llmdet} 
& \textbf{fastdetectllm} \\
\midrule
\textbf{Qwen2.5-72B-Instruct} & 0.983 & 0.981 & 0.978 & 0.917 & 0.823 & 0.862 & 0.716 & 0.554 & 0.489 & 0.474 \\
\textbf{Qwen2.5-32B-Instruct} & 0.983 & 0.932 & 0.866 & 0.800 & 0.817 & 0.725 & 0.547 & 0.478 & 0.477 & 0.474 \\
\textbf{Qwen2.5-14B-Instruct} & 0.983 & 0.913 & 0.883 & 0.763 & 0.903 & 0.820 & 0.615 & 0.497 & 0.487 & 0.474 \\
\textbf{Qwen2.5-7B-Instruct} & 0.983 & 0.978 & 0.929 & 0.860 & 0.829 & 0.764 & 0.546 & 0.481 & 0.478 & 0.474 \\
\textbf{Qwen2.5-3B-Instruct} & 0.967 & 0.743 & 0.643 & 0.710 & 0.791 & 0.574 & 0.520 & 0.480 & 0.490 & 0.474 \\
\textbf{Qwen2.5-1.5B-Instruct} & 0.956 & 0.900 & 0.717 & 0.803 & 0.749 & 0.667 & 0.587 & 0.531 & 0.489 & 0.475 \\
\textbf{Qwen2.5-0.5B-Instruct} & 0.861 & 0.895 & 0.670 & 0.803 & 0.670 & 0.649 & 0.669 & 0.773 & 0.484 & 0.474 \\
\textbf{llama3.1-70b-instruct} & 0.943 & 0.897 & 0.885 & 0.787 & 0.716 & 0.696 & 0.664 & 0.547 & 0.540 & 0.480 \\
\textbf{llama3.1-8b-instruct} & 0.979 & 0.969 & 0.941 & 0.845 & 0.682 & 0.737 & 0.726 & 0.603 & 0.503 & 0.474 \\
\textbf{llama3.2-3b-instruct} & 0.936 & 0.899 & 0.842 & 0.761 & 0.627 & 0.714 & 0.660 & 0.573 & 0.521 & 0.481 \\
\textbf{llama3.2-1b-instruct} & 0.846 & 0.796 & 0.705 & 0.724 & 0.695 & 0.658 & 0.678 & 0.681 & 0.506 & 0.516 \\
\textbf{Mistral-Small-Instruct-2409} & 0.983 & 0.957 & 0.816 & 0.794 & 0.675 & 0.644 & 0.552 & 0.477 & 0.491 & 0.474 \\
\textbf{Average} & 0.950 & 0.905 & 0.823 & 0.797 & 0.748 & 0.709 & 0.623 & 0.556 & 0.496 & 0.479 \\
\bottomrule
\end{tabular}%
}
\caption{Detector performance across various generator models under subgroup Language Environment = EFL}
\label{tab:language environment_EFL}
\end{table*}

\begin{table*}[ht!]
\centering
\resizebox{\textwidth}{!}{
\begin{tabular}{lcccccccccccc}
\toprule
\textbf{Generator Model} 
& \textbf{binoculars} 
& \textbf{fastdetectgpt} 
& \textbf{gltr} 
& \textbf{detectgpt} 
& \textbf{chatgpt-roberta} 
& \textbf{radar} 
& \textbf{gpt2-base} 
& \textbf{gpt2-large} 
& \textbf{llmdet} 
& \textbf{fastdetectllm} \\
\midrule
\textbf{Qwen2.5-72B-Instruct} & 0.967 & 0.967 & 0.977 & 0.904 & 0.787 & 0.836 & 0.665 & 0.534 & 0.501 & 0.477 \\
\textbf{Qwen2.5-32B-Instruct} & 0.966 & 0.907 & 0.813 & 0.816 & 0.757 & 0.691 & 0.535 & 0.505 & 0.490 & 0.477 \\
\textbf{Qwen2.5-14B-Instruct} & 0.967 & 0.890 & 0.874 & 0.777 & 0.886 & 0.805 & 0.605 & 0.514 & 0.498 & 0.477 \\
\textbf{Qwen2.5-7B-Instruct} & 0.967 & 0.962 & 0.915 & 0.833 & 0.823 & 0.686 & 0.555 & 0.507 & 0.494 & 0.477 \\
\textbf{Qwen2.5-3B-Instruct} & 0.957 & 0.739 & 0.654 & 0.698 & 0.750 & 0.567 & 0.544 & 0.509 & 0.501 & 0.477 \\
\textbf{Qwen2.5-1.5B-Instruct} & 0.935 & 0.869 & 0.690 & 0.793 & 0.732 & 0.643 & 0.585 & 0.549 & 0.507 & 0.477 \\
\textbf{Qwen2.5-0.5B-Instruct} & 0.839 & 0.873 & 0.648 & 0.758 & 0.640 & 0.600 & 0.670 & 0.804 & 0.510 & 0.478 \\
\textbf{llama3.1-70b-Instruct} & 0.942 & 0.908 & 0.911 & 0.824 & 0.764 & 0.718 & 0.652 & 0.563 & 0.533 & 0.479 \\
\textbf{llama3.1-8b-Instruct} & 0.963 & 0.958 & 0.940 & 0.843 & 0.690 & 0.746 & 0.691 & 0.584 & 0.510 & 0.477 \\
\textbf{llama3.2-3b-Instruct} & 0.936 & 0.911 & 0.871 & 0.788 & 0.648 & 0.737 & 0.673 & 0.576 & 0.517 & 0.479 \\
\textbf{llama3.2-1b-Instruct} & 0.841 & 0.814 & 0.740 & 0.732 & 0.712 & 0.623 & 0.710 & 0.692 & 0.521 & 0.533 \\
\textbf{Mistral-Small-Instruct-2409} & 0.967 & 0.948 & 0.740 & 0.732 & 0.712 & 0.623 & 0.710 & 0.692 & 0.521 & 0.533 \\
\textbf{Average} & 0.937 & 0.896 & 0.821 & 0.795 & 0.741 & 0.690 & 0.620 & 0.570 & 0.507 & 0.482 \\
\bottomrule
\end{tabular}
}\caption{Detector performance across various generator models under subgroup Language Environment = ESL}
\label{tab:language environment_ESL}
\end{table*}

\begin{table*}[ht]
\centering
\resizebox{\textwidth}{!}{%
\begin{tabular}{lcccccccccccc}
\toprule
\textbf{Generator Model} 
& \textbf{binoculars} 
& \textbf{fastdetectgpt} 
& \textbf{chatgpt-roberta} 
& \textbf{detectgpt} 
& \textbf{radar} 
& \textbf{gltr} 
& \textbf{gpt2-large} 
& \textbf{gpt2-base} 
& \textbf{llmdet} 
& \textbf{fastdetectllm} \\
\midrule
\textbf{Qwen2.5-72B-Instruct} & 0.875 & 0.923 & 0.809 & 0.837 & 0.685 & 0.756 & 0.460 & 0.453 & 0.503 & 0.500 \\
\textbf{Qwen2.5-32B-Instruct} & 0.875 & 0.823 & 0.773 & 0.711 & 0.598 & 0.527 & 0.457 & 0.395 & 0.509 & 0.500 \\
\textbf{Qwen2.5-14B-Instruct} & 0.875 & 0.857 & 0.773 & 0.675 & 0.741 & 0.632 & 0.464 & 0.421 & 0.500 & 0.500 \\
\textbf{Qwen2.5-7B-Instruct} & 0.875 & 0.922 & 0.809 & 0.799 & 0.545 & 0.694 & 0.457 & 0.410 & 0.500 & 0.500 \\
\textbf{Qwen2.5-3B-Instruct} & 0.873 & 0.693 & 0.766 & 0.645 & 0.537 & 0.394 & 0.457 & 0.399 & 0.506 & 0.500 \\
\textbf{Qwen2.5-1.5B-Instruct} & 0.824 & 0.815 & 0.689 & 0.730 & 0.603 & 0.451 & 0.521 & 0.500 & 0.521 & 0.503 \\
\textbf{Qwen2.5-0.5B-Instruct} & 0.699 & 0.806 & 0.625 & 0.664 & 0.594 & 0.436 & 0.761 & 0.587 & 0.522 & 0.500 \\
\textbf{Llama3.1-70b-Instruct} & 0.873 & 0.915 & 0.877 & 0.810 & 0.804 & 0.755 & 0.476 & 0.522 & 0.503 & 0.502 \\
\textbf{Llama3.1-8b-Instruct} & 0.872 & 0.917 & 0.721 & 0.782 & 0.664 & 0.716 & 0.462 & 0.485 & 0.503 & 0.500 \\
\textbf{Llama3.2-3b-Instruct} & 0.870 & 0.913 & 0.716 & 0.797 & 0.759 & 0.726 & 0.491 & 0.488 & 0.509 & 0.500 \\
\textbf{Llama3.2-1b-Instruct} & 0.739 & 0.763 & 0.728 & 0.667 & 0.573 & 0.531 & 0.586 & 0.581 & 0.510 & 0.548 \\
\textbf{Mistral-Small-Instruct-2409} & 0.875 & 0.914 & 0.693 & 0.697 & 0.599 & 0.576 & 0.457 & 0.394 & 0.500 & 0.500 \\
\textbf{Average} & 0.844 & 0.855 & 0.748 & 0.735 & 0.642 & 0.600 & 0.504 & 0.470 & 0.507 & 0.504 \\
\bottomrule
\end{tabular}%
}
\caption{Detector performance across various generator models under subgroup Language Environment = NS}
\label{tab:language environment_NS}
\end{table*}

\begin{table*}[ht!]
\centering
\resizebox{\textwidth}{!}{
\begin{tabular}{lcccccccccccc}
\toprule
\textbf{Generator Model} 
& \textbf{binoculars} 
& \textbf{fastdetectgpt} 
& \textbf{gltr} 
& \textbf{detectgpt} 
& \textbf{chatgpt-roberta} 
& \textbf{radar} 
& \textbf{gpt2-base} 
& \textbf{gpt2-large} 
& \textbf{llmdet} 
& \textbf{fastdetectllm} \\
\midrule
\textbf{Qwen2.5-72B-Instruct} & 0.951 & 0.963 & 0.937 & 0.903 & 0.820 & 0.814 & 0.649 & 0.538 & 0.495 & 0.479 \\
\textbf{Qwen2.5-32B-Instruct} & 0.950 & 0.892 & 0.788 & 0.793 & 0.783 & 0.681 & 0.513 & 0.484 & 0.484 & 0.479 \\
\textbf{Qwen2.5-14B-Instruct} & 0.951 & 0.895 & 0.839 & 0.757 & 0.877 & 0.805 & 0.574 & 0.500 & 0.492 & 0.479 \\
\textbf{Qwen2.5-7B-Instruct} & 0.951 & 0.959 & 0.880 & 0.841 & 0.807 & 0.705 & 0.521 & 0.485 & 0.484 & 0.479 \\
\textbf{Qwen2.5-3B-Instruct} & 0.938 & 0.733 & 0.603 & 0.690 & 0.775 & 0.558 & 0.510 & 0.490 & 0.494 & 0.479 \\
\textbf{Qwen2.5-1.5B-Instruct} & 0.919 & 0.871 & 0.676 & 0.785 & 0.736 & 0.645 & 0.579 & 0.544 & 0.498 & 0.480 \\
\textbf{Qwen2.5-0.5B-Instruct} & 0.817 & 0.867 & 0.626 & 0.770 & 0.650 & 0.620 & 0.652 & 0.779 & 0.499 & 0.480 \\
\textbf{Llama3.1-70b-instruct} & 0.926 & 0.903 & 0.882 & 0.813 & 0.764 & 0.722 & 0.639 & 0.547 & 0.523 & 0.483 \\
\textbf{Llama3.1-8b-instruct} & 0.946 & 0.951 & 0.900 & 0.836 & 0.679 & 0.714 & 0.671 & 0.571 & 0.500 & 0.479 \\
\textbf{Llama3.2-3b-instruct} & 0.921 & 0.903 & 0.832 & 0.786 & 0.642 & 0.727 & 0.628 & 0.562 & 0.517 & 0.483 \\
\textbf{Llama3.2-1b-instruct} & 0.834 & 0.805 & 0.699 & 0.716 & 0.708 & 0.630 & 0.670 & 0.671 & 0.510 & 0.521 \\
\textbf{Mistral-Small-Instruct-2409} & 0.951 & 0.941 & 0.776 & 0.787 & 0.675 & 0.619 & 0.518 & 0.482 & 0.494 & 0.479 \\
\textbf{Average} & 0.921 & 0.890 & 0.787 & 0.790 & 0.743 & 0.687 & 0.594 & 0.554 & 0.499 & 0.483 \\
\bottomrule
\end{tabular}
}
\caption{Detector performance across various generator models under subgroup Sex = Female}
\label{tab:sec_female}
\end{table*}

\begin{table*}[ht!]
\centering
\resizebox{\textwidth}{!}{
\begin{tabular}{lcccccccccccccc}
\toprule
\textbf{Generator Model} 
& \textbf{binoculars} 
& \textbf{fastdetectgpt} 
& \textbf{gltr} 
& \textbf{detectgpt} 
& \textbf{chatgpt-roberta} 
& \textbf{radar} 
& \textbf{gpt2-base} 
& \textbf{gpt2-large} 
& \textbf{llmdet} 
& \textbf{fastdetectllm} \\
\midrule
\textbf{Qwen2.5-72B-Instruct} & 0.968 & 0.972 & 0.942 & 0.891 & 0.793 & 0.834 & 0.657 & 0.522 & 0.496 & 0.479 \\
\textbf{Qwen2.5-32B-Instruct} & 0.968 & 0.921 & 0.791 & 0.788 & 0.797 & 0.704 & 0.523 & 0.482 & 0.492 & 0.479 \\
\textbf{Qwen2.5-14B-Instruct} & 0.967 & 0.899 & 0.834 & 0.747 & 0.868 & 0.798 & 0.584 & 0.493 & 0.496 & 0.479 \\
\textbf{Qwen2.5-7B-Instruct} & 0.968 & 0.969 & 0.890 & 0.837 & 0.839 & 0.694 & 0.529 & 0.485 & 0.491 & 0.479 \\
\textbf{Qwen2.5-3B-Instruct} & 0.956 & 0.736 & 0.606 & 0.699 & 0.769 & 0.570 & 0.503& 0.480 & 0.501 & 0.479 \\
\textbf{Qwen2.5-1.5B-Instruct} & 0.933 & 0.881 & 0.644 & 0.788 & 0.729 & 0.651 & 0.560 & 0.524 & 0.504 & 0.480 \\
\textbf{Qwen2.5-0.5B-Instruct} & 0.834 & 0.880 & 0.617 & 0.754 & 0.653 & 0.629 & 0.658 & 0.777 & 0.501 & 0.479 \\
\textbf{Llama3.1-70B-Instruct} & 0.936 & 0.909 & 0.858 & 0.788 & 0.750 & 0.721 & 0.637 & 0.532 & 0.541 & 0.483 \\
\textbf{Llama3.1-8b-Instruct} & 0.965 & 0.964 & 0.906 & 0.828 & 0.703 & 0.741 & 0.674 & 0.575 & 0.512 & 0.479 \\
\textbf{Llama3.2-3b-Instruct} & 0.929 & 0.911 & 0.832 & 0.766 & 0.654 & 0.730 & 0.641 & 0.555 & 0.520 & 0.483 \\
\textbf{Llama3.2-1b-Instruct} & 0.811 & 0.783 & 0.671 & 0.715 & 0.702 & 0.633 & 0.671 & 0.666 & 0.514 & 0.535 \\
\textbf{Mistral-Small-Instruct-2409} & 0.968 & 0.954 & 0.771 & 0.742 & 0.699 & 0.644 & 0.533 & 0.480 & 0.502 & 0.479 \\
\textbf{Average} & 0.934 & 0.898 & 0.780 & 0.779 & 0.746 & 0.696 & 0.597 & 0.548 & 0.506 & 0.485 \\
\bottomrule
\end{tabular}
}
\caption{Detector performance across various generator models under subgroup Sex = Male}
\label{tab:sex_male}
\end{table*}

\begin{table*}[ht!]
\centering
\resizebox{\textwidth}{!}{
\begin{tabular}{lcccccccccccc}
\toprule
\textbf{Generator Model} 
& \textbf{binoculars} 
& \textbf{fastdetectgpt} 
& \textbf{gltr} 
& \textbf{detectgpt} 
& \textbf{chatgpt-roberta} 
& \textbf{radar} 
& \textbf{gpt2-base} 
& \textbf{gpt2-large} 
& \textbf{llmdet} 
& \textbf{fastdetectllm} \\
\midrule
\textbf{Qwen2.5-72B-Instruct} & 0.962 & 0.960 & 0.948 & 0.921 & 0.794 & 0.823 & 0.619 & 0.510 & 0.491 & 0.482 \\
\textbf{Qwen2.5-32B-Instruct} & 0.962 & 0.889 & 0.771 & 0.799 & 0.759 & 0.669 & 0.489 & 0.483 & 0.485 & 0.482 \\
\textbf{Qwen2.5-14B-Instruct} & 0.961 & 0.910 & 0.859 & 0.789 & 0.873 & 0.811 & 0.565 & 0.496 & 0.490 & 0.482 \\
\textbf{Qwen2.5-7B-Instruct} & 0.962 & 0.957 & 0.890 & 0.845 & 0.806 & 0.705 & 0.525 & 0.487 & 0.484 & 0.482 \\
\textbf{Qwen2.5-3B-Instruct} & 0.953 & 0.729 & 0.605 & 0.716 & 0.770 & 0.571 & 0.496 & 0.486 & 0.490 & 0.482 \\
\textbf{Qwen2.5-1.5B-Instruct} & 0.927 & 0.854 & 0.640 & 0.777 & 0.703 & 0.629 & 0.556 & 0.531 & 0.498 & 0.482 \\
\textbf{Qwen2.5-0.5B-Instruct} & 0.811 & 0.851 & 0.627 & 0.770 & 0.637 & 0.612 & 0.660 & 0.796 & 0.503 & 0.482 \\
\textbf{llama3.1-70b-instruct} & 0.934 & 0.890 & 0.885 & 0.799 & 0.755 & 0.723 & 0.613 & 0.528 & 0.528 & 0.487 \\
\textbf{llama3.1-8b-instruct} & 0.957 & 0.946 & 0.894 & 0.852 & 0.679 & 0.732 & 0.643 & 0.546 & 0.498 & 0.482 \\
\textbf{llama3.2-3b-instruct} & 0.938 & 0.906 & 0.841 & 0.810 & 0.651 & 0.743 & 0.624 & 0.552 & 0.508 & 0.486 \\
\textbf{llama3.2-1b-instruct} & 0.838 & 0.790 & 0.694 & 0.732 & 0.695 & 0.638 & 0.659 & 0.665 & 0.511 & 0.529 \\
\textbf{Mistral-Small-Instruct-2409} & 0.962 & 0.935 & 0.757 & 0.791 & 0.677 & 0.626 & 0.497 & 0.484 & 0.499 & 0.482 \\
\textbf{Average} & 0.931 & 0.885 & 0.784 & 0.800 & 0.733 & 0.690 & 0.579 & 0.547 & 0.499 & 0.486 \\
\bottomrule
\end{tabular}
}
\caption{Detector performance across various generator models under subgroup Academic Genre = Humanities}
\label{tab:academic genre_humanities}
\end{table*}

\begin{table*}[ht!]
\centering
\resizebox{\textwidth}{!}{
\begin{tabular}{lcccccccccccc}
\toprule
\textbf{Generator Model} 
& \textbf{binoculars} 
& \textbf{fastdetectgpt} 
& \textbf{gltr} 
& \textbf{detectgpt} 
& \textbf{chatgpt-roberta} 
& \textbf{radar} 
& \textbf{gpt2-base} 
& \textbf{gpt2-large} 
& \textbf{llmdet} 
& \textbf{fastdetectllm} \\
\midrule
\textbf{Qwen2.5-72B-Instruct} & 0.951 & 0.976 & 0.931 & 0.868 & 0.852 & 0.777 & 0.683 & 0.562 & 0.496 & 0.476 \\
\textbf{Qwen2.5-32B-Instruct} & 0.951 & 0.911 & 0.797 & 0.751 & 0.824 & 0.690 & 0.553 & 0.490 & 0.493 & 0.476 \\
\textbf{Qwen2.5-14B-Instruct} & 0.951 & 0.890 & 0.810 & 0.697 & 0.864 & 0.775 & 0.589 & 0.502 & 0.497 & 0.476 \\
\textbf{Qwen2.5-7B-Instruct} & 0.951 & 0.968 & 0.880 & 0.810 & 0.834 & 0.689 & 0.511 & 0.495 & 0.495 & 0.476 \\
\textbf{Qwen2.5-3B-Instruct} & 0.939 & 0.751 & 0.611 & 0.668 & 0.756 & 0.516 & 0.518 & 0.494 & 0.498 & 0.476 \\
\textbf{Qwen2.5-1.5B-Instruct} & 0.921 & 0.908 & 0.648 & 0.784 & 0.744 & 0.665 & 0.561 & 0.519 & 0.496 & 0.476 \\
\textbf{Qwen2.5-0.5B-Instruct} & 0.819 & 0.863 & 0.613 & 0.723 & 0.660 & 0.642 & 0.671 & 0.781 & 0.495 & 0.479 \\
\textbf{Llama3.1-70B-Instruct} & 0.926 & 0.932 & 0.878 & 0.787 & 0.756 & 0.724 & 0.664 & 0.556 & 0.530 & 0.478 \\
\textbf{Llama3.1-8B-Instruct} & 0.948 & 0.972 & 0.909 & 0.816 & 0.703 & 0.694 & 0.683 & 0.593 & 0.504 & 0.476 \\
\textbf{Llama3.2-3B-Instruct} & 0.912 & 0.913 & 0.844 & 0.760 & 0.648 & 0.697 & 0.640 & 0.571 & 0.531 & 0.483 \\
\textbf{Llama3.2-1B-Instruct} & 0.806 & 0.785 & 0.681 & 0.683 & 0.737 & 0.600 & 0.673 & 0.671 & 0.513 & 0.529 \\
\textbf{Mistral-Small-Instruct-2409} & 0.951 & 0.963 & 0.794 & 0.714 & 0.717 & 0.602 & 0.528 & 0.486 & 0.499 & 0.476 \\
\textbf{Average} & 0.919 & 0.903 & 0.783 & 0.755 & 0.758 & 0.673 & 0.606 & 0.560 & 0.504 & 0.481 \\
\bottomrule
\end{tabular}
}
\caption{Detector performance across various generator models under subgroup Academic Genre = Life Science}
\label{tab:academic genre_life science}
\end{table*}

\begin{table*}[ht!]
\centering
\resizebox{\textwidth}{!}{
\begin{tabular}{lcccccccccccc}
\toprule
\textbf{Generator Model} 
& \textbf{binoculars} 
& \textbf{fastdetectgpt} 
& \textbf{gltr} 
& \textbf{detectgpt} 
& \textbf{chatgpt-roberta} 
& \textbf{radar} 
& \textbf{gpt2-base} 
& \textbf{gpt2-large} 
& \textbf{llmdet} 
& \textbf{fastdetectllm} \\
\midrule
\textbf{Qwen2.5-72B-Instruct} & 0.962 & 0.958 & 0.941 & 0.887 & 0.817 & 0.826 & 0.694 & 0.559 & 0.501 & 0.477 \\
\textbf{Qwen2.5-32B-Instruct} & 0.962 & 0.913 & 0.814 & 0.796 & 0.797 & 0.717 & 0.543 & 0.497 & 0.486 & 0.477 \\
\textbf{Qwen2.5-14B-Instruct} & 0.962 & 0.883 & 0.853 & 0.761 & 0.878 & 0.806 & 0.599 & 0.514 & 0.492 & 0.477 \\
\textbf{Qwen2.5-7B-Instruct} & 0.962 & 0.954 & 0.893 & 0.844 & 0.847 & 0.702 & 0.543 & 0.493 & 0.484 & 0.477 \\
\textbf{Qwen2.5-3B-Instruct} & 0.942 & 0.706 & 0.609 & 0.673 & 0.784 & 0.570 & 0.519 & 0.496 & 0.504 & 0.477 \\
\textbf{Qwen2.5-1.5B-Instruct} & 0.935 & 0.868 & 0.674 & 0.797 & 0.752 & 0.647 & 0.574 & 0.544 & 0.496 & 0.479 \\
\textbf{Qwen2.5-0.5B-Instruct} & 0.841 & 0.872 & 0.626 & 0.775 & 0.658 & 0.631 & 0.649 & 0.490 & 0.490 & 0.477 \\
\textbf{llama3.1-70b-Instruct} & 0.931 & 0.894 & 0.868 & 0.802 & 0.756 & 0.708 & 0.668 & 0.568 & 0.537 & 0.480 \\
\textbf{llama3.1-8b-Instruct} & 0.960 & 0.950 & 0.913 & 0.812 & 0.681 & 0.712 & 0.717 & 0.612 & 0.511 & 0.477 \\
\textbf{llama3.2-3b-Instruct} & 0.928 & 0.895 & 0.827 & 0.751 & 0.642 & 0.733 & 0.644 & 0.580 & 0.521 & 0.489 \\
\textbf{llama3.2-1b-Instruct} & 0.816 & 0.776 & 0.674 & 0.698 & 0.702 & 0.634 & 0.680 & 0.653 & 0.518 & 0.529 \\
\textbf{Mistral-Small-Instruct-2409} & 0.962 & 0.937 & 0.784 & 0.766 & 0.683 & 0.649 & 0.547 & 0.489 & 0.494 & 0.477 \\
\textbf{Average} & 0.930 & 0.884 & 0.790 & 0.780 & 0.750 & 0.695 & 0.615 & 0.568 & 0.503 & 0.482 \\
\bottomrule
\end{tabular}
}
\caption{Detector performance across various generator models under subgroup Academic Genre = Science Technology}
\label{tab:academic genre_science technology}
\end{table*}

\begin{table*}[ht!]
\centering
\resizebox{\textwidth}{!}{
\begin{tabular}{lcccccccccccc}
\toprule
\textbf{Generator Model} 
& \textbf{binoculars} 
& \textbf{fastdetectgpt} 
& \textbf{gltr} 
& \textbf{detectgpt} 
& \textbf{chatgpt-roberta} 
& \textbf{radar} 
& \textbf{gpt2-base} 
& \textbf{gpt2-large} 
& \textbf{llmdet} 
& \textbf{fastdetectllm} \\
\midrule
\textbf{Qwen2.5-72B-Instruct} & 0.955 & 0.977 & 0.932 & 0.906 & 0.780 & 0.847 & 0.619 & 0.500 & 0.495 & 0.481 \\
\textbf{Qwen2.5-32B-Instruct} & 0.954 & 0.909 & 0.774 & 0.798 & 0.784 & 0.686 & 0.494 & 0.462 & 0.487 & 0.481 \\
\textbf{Qwen2.5-14B-Instruct} & 0.955 & 0.901 & 0.813 & 0.742 & 0.876 & 0.806 & 0.563 & 0.472 & 0.495 & 0.481 \\
\textbf{Qwen2.5-7B-Instruct} & 0.955 & 0.976 & 0.870 & 0.849 & 0.806 & 0.699 & 0.515 & 0.465 & 0.488 & 0.481 \\
\textbf{Qwen2.5-3B-Instruct} & 0.948 & 0.760 & 0.590 & 0.711 & 0.780 & 0.579 & 0.495 & 0.466 & 0.495 & 0.481 \\
\textbf{Qwen2.5-1.5B-Instruct} & 0.916 & 0.881 & 0.676 & 0.784 & 0.735 & 0.651 & 0.587 & 0.537 & 0.510 & 0.481 \\
\textbf{Qwen2.5-0.5B-Instruct} & 0.823 & 0.901 & 0.616 & 0.768 & 0.652 & 0.607 & 0.645 & 0.762 & 0.509 & 0.481 \\
\textbf{llama3.1-70b-Instruct} & 0.929 & 0.911 & 0.857 & 0.814 & 0.767 & 0.728 & 0.609 & 0.510 & 0.531 & 0.486 \\
\textbf{llama3.1-8b-Instruct} & 0.950 & 0.964 & 0.891 & 0.844 & 0.701 & 0.753 & 0.644 & 0.540 & 0.507 & 0.481 \\
\textbf{llama3.2-3b-Instruct} & 0.916 & 0.913 & 0.817 & 0.780 & 0.656 & 0.730 & 0.632 & 0.535 & 0.518 & 0.484 \\
\textbf{llama3.2-1b-Instruct} & 0.832 & 0.827 & 0.694 & 0.740 & 0.700 & 0.643 & 0.673 & 0.653 & 0.507 & 0.523 \\
\textbf{Mistral-Small-Instruct-2409} & 0.955 & 0.956 & 0.760 & 0.782 & 0.674 & 0.632 & 0.522 & 0.464 & 0.499 & 0.481 \\
\textbf{Average} & 0.924 & 0.906 & 0.774 & 0.793 & 0.743 & 0.697 & 0.583 & 0.531 & 0.504 & 0.485 \\
\bottomrule
\end{tabular}
}
\caption{Detector performance across various generator models under subgroup Academic Genre = Social Science}
\label{tab:academic genre_social science}
\end{table*}

\clearpage
\clearpage
\section{Statistics of Generated Texts}

\begin{table*}[h]
\small
\centering
\begin{tabular}{c|c|c}
\toprule
\textbf{Category} & \textbf{Feature} & \textbf{Explanation} \\
\midrule
\multirow{4}{*}{Basic}
 & Word Count         & Total words \\
 & Unique Words       & Distinct words \\
 & Character Count    & Total characters \\
 & Sentence Count     & Number of sentences \\
\midrule
\multirow{2}{*}{Lexical}
 & Avg Word Length    & (Characters - special chars) / words \\
 & Type-Token Ratio   & Distinct words / total words \\
\midrule
Syntactic / POS
 & N/V/Adj Ratio      & Share of nouns, verbs, adjectives \\
\midrule
\multirow{2}{*}{Readability}
 & Flesch Reading Ease & Formula-based readability score \\
 & Gunning Fog Index   & Index using complex words \& sentence length \\
\bottomrule
\end{tabular}
\caption{Features and Explanations}
\label{tab:features_explanations}
\end{table*}

\begin{figure*}[h]
    \centering
    \includegraphics[width=\linewidth]{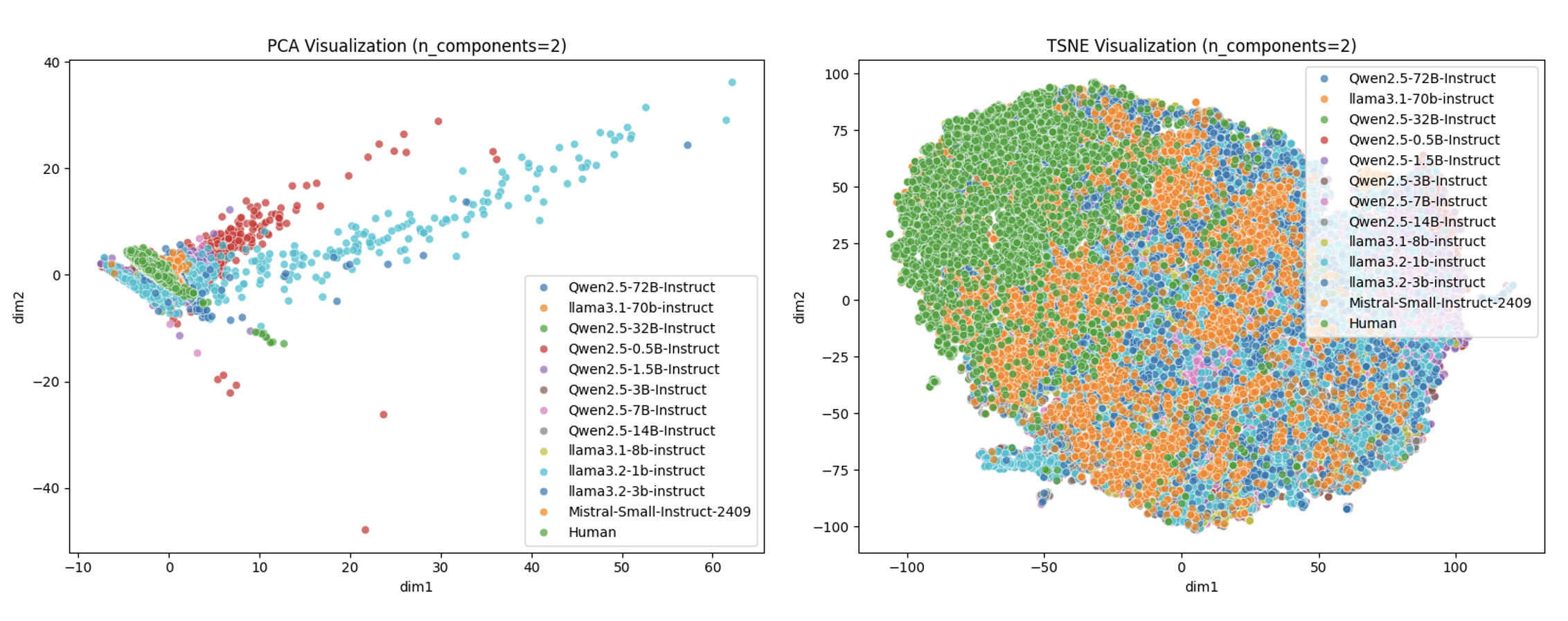}
    \caption{Visualization results of PCA (left) and t-SNE (right)}
    \label{fig:pca_tsne}
\end{figure*}

We performed dimensionality reduction to visualize the text feature distributions from different models alongside human-written texts. Figure~\ref{fig:pca_tsne} shows PCA (left) and t-SNE (right) results in 2D space. Under PCA, texts generated by each model family often cluster together, indicating relatively similar language patterns within a family. Under t-SNE, the overall point cloud is more scattered, with each model forming a denser cluster but still overlapping somewhat with other models. Human texts, meanwhile, appear distinct from model-generated texts, implying that the two groups remain separable in feature space. Interestingly, points from the same model family tend to reside near each other, reflecting shared output styles, while the gap between human and AI texts suggests potential discriminative power for detection tasks.

\begin{figure}[H]
    \centering
    \includegraphics[width=\linewidth]{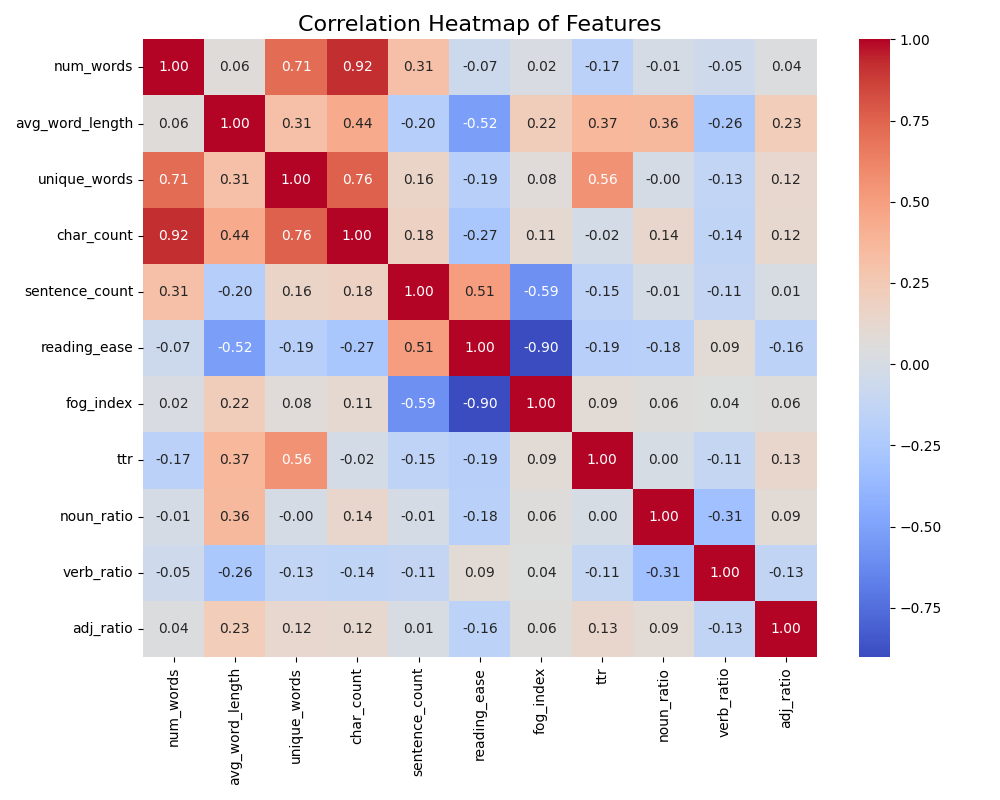}
    \caption{Human text features heatmap}
    \label{fig:human_heatmap}
\end{figure}

\begin{figure}[H]
    \centering
    \includegraphics[width=\linewidth]{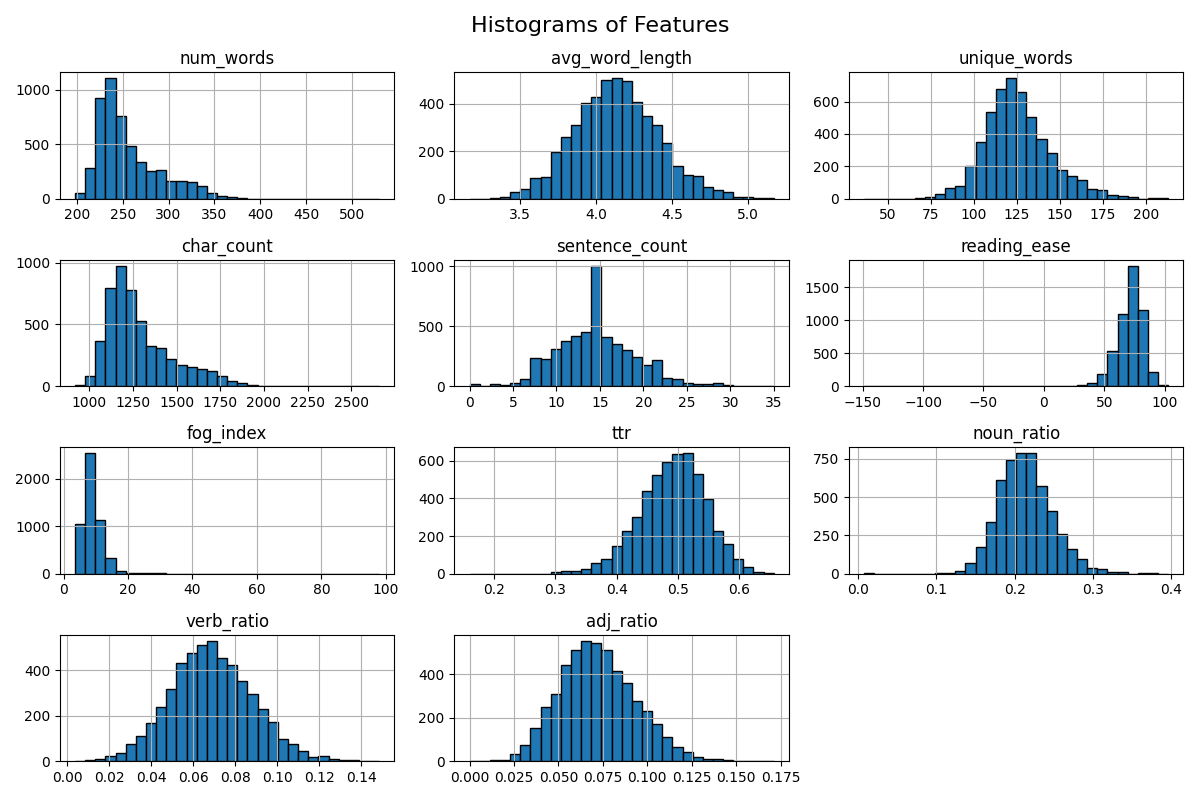}
    \caption{Human text features histgrams}
    \label{fig:human_hist}
\end{figure}

\begin{figure}[H]
    \centering
    \includegraphics[width=\linewidth]{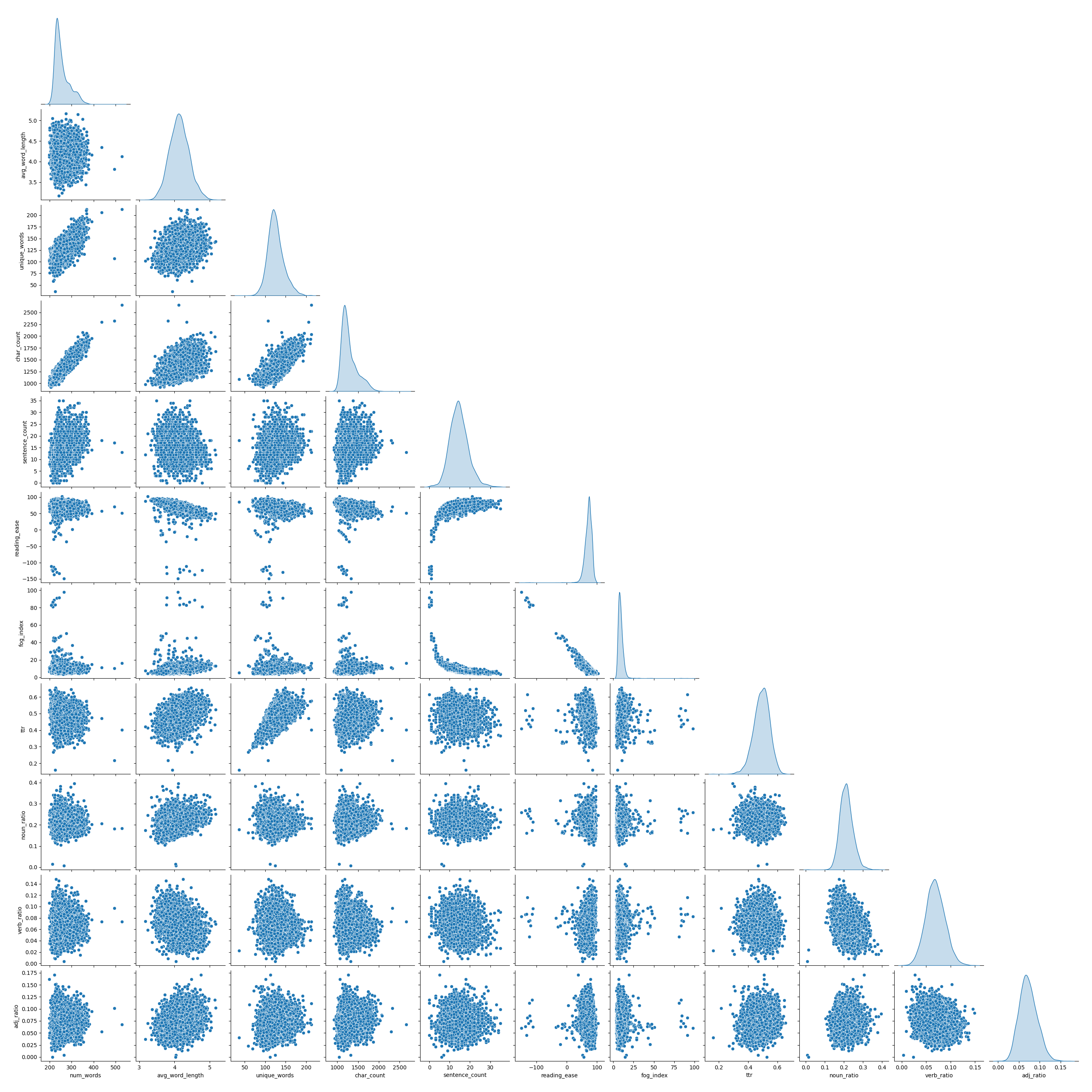}
    \caption{Human text features pairplot}
    \label{fig:human_pairplot}
\end{figure}


\begin{figure}[H]
    \centering
    \includegraphics[width=\linewidth]{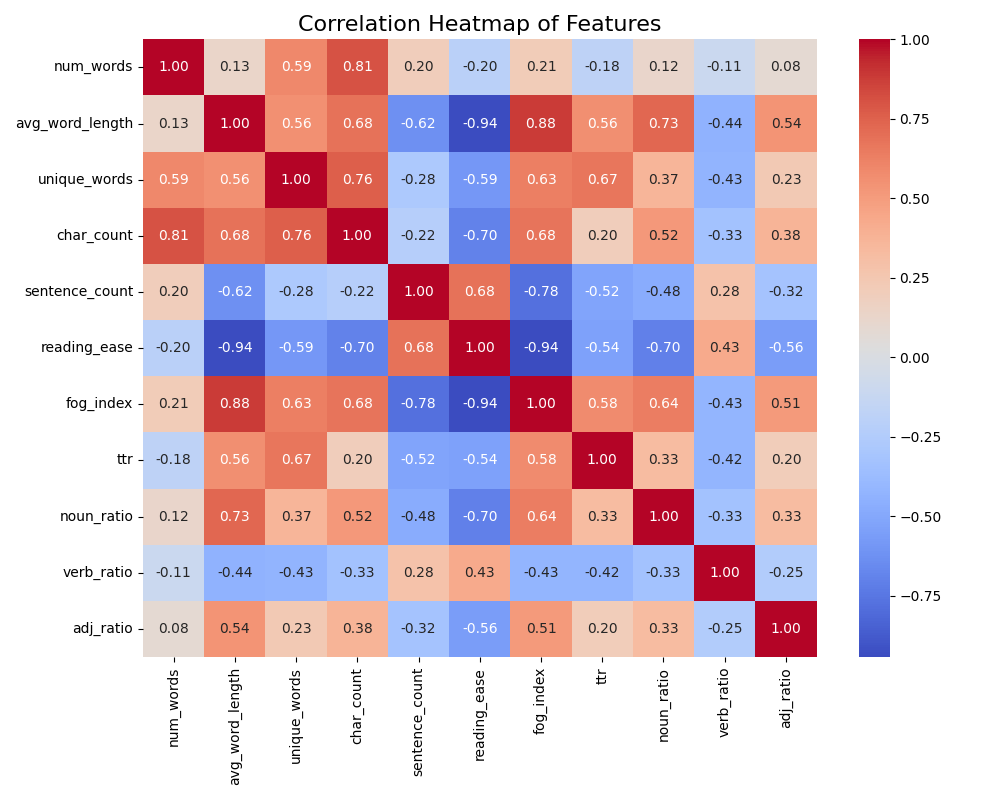}
    \caption{Llama3.1-8B-Instruct text features heatmap}
    \label{fig:human_heatmap}
\end{figure}

\begin{figure}[H]
    \centering
    \includegraphics[width=\linewidth]{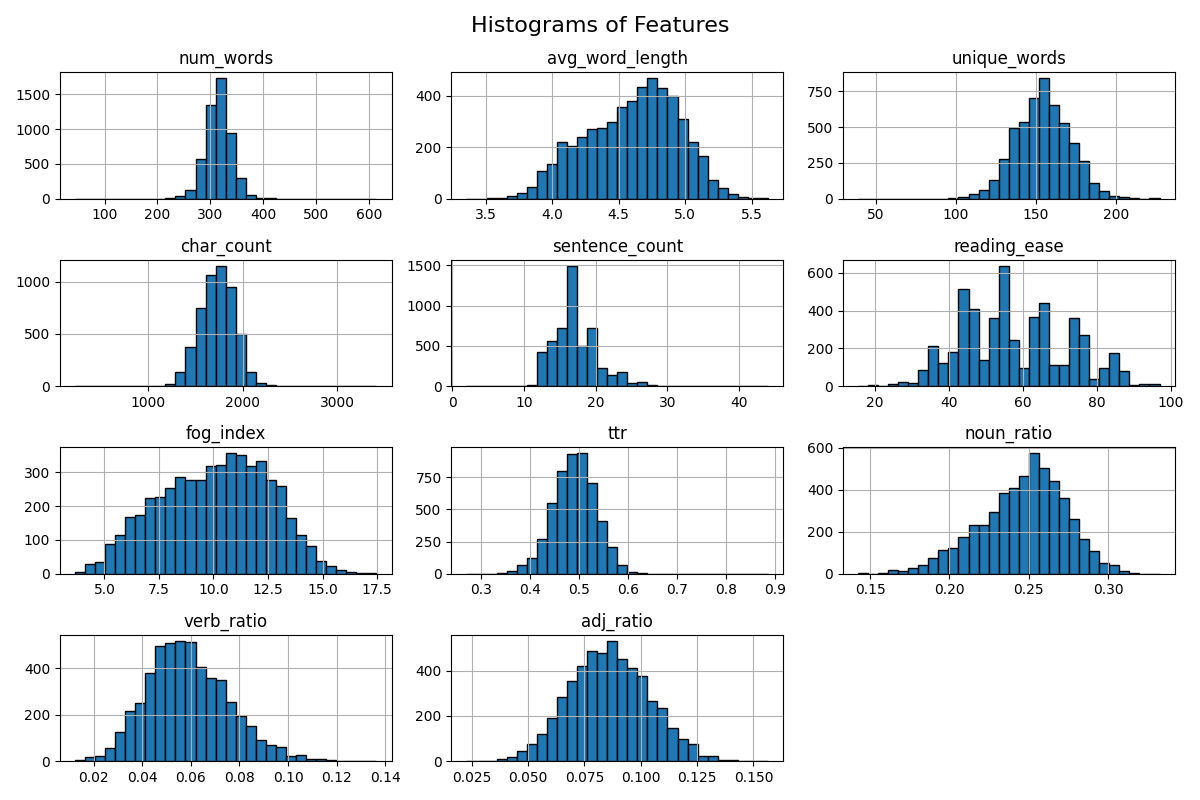}
    \caption{Llama3.1-8B-Instruct text features histgrams}
    \label{fig:human_hist}
\end{figure}

\begin{figure}[H]
    \centering
    \includegraphics[width=\linewidth]{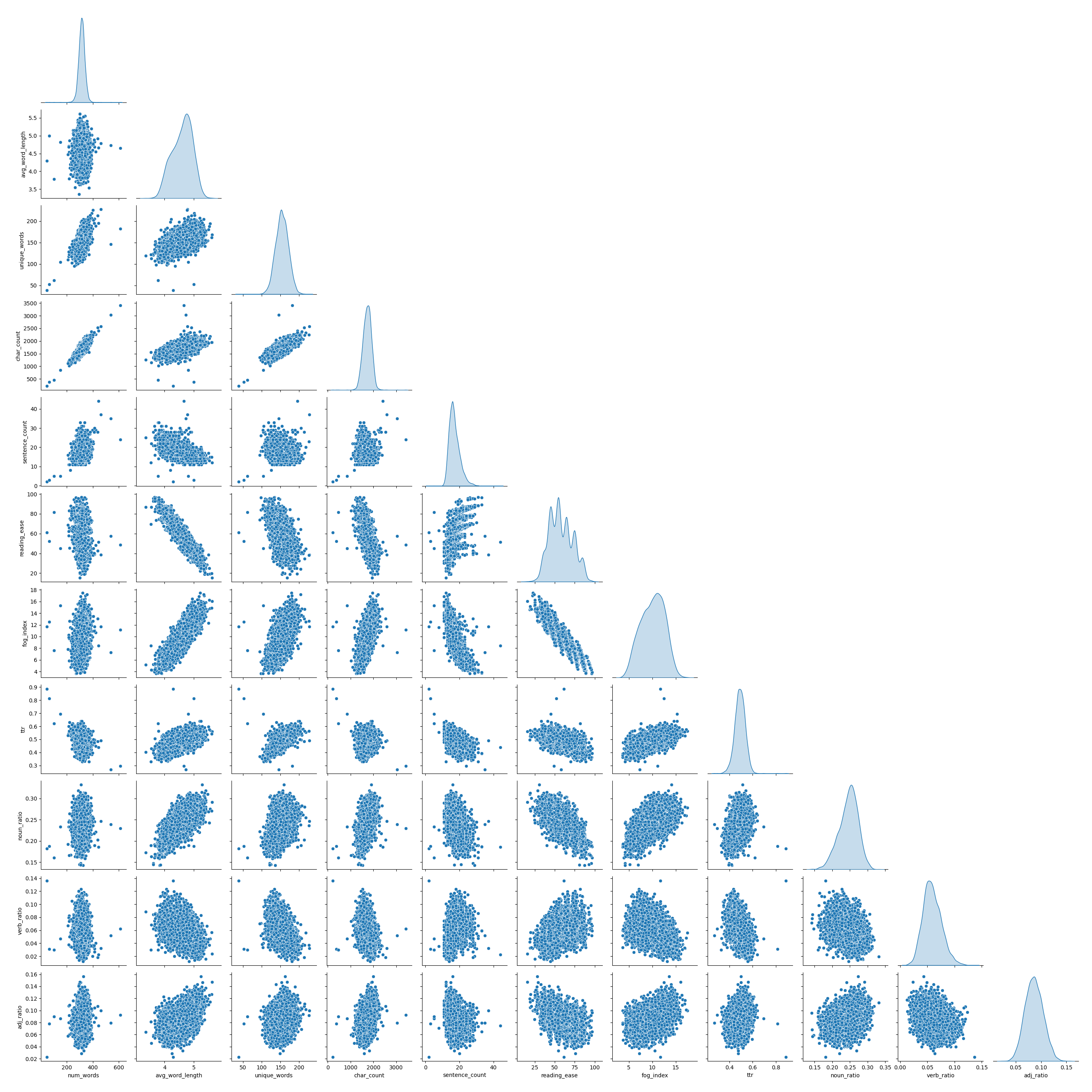}
    \caption{Llama3.1-8B-Instruct text features pairplot}
    \label{fig:human_pairplot}
\end{figure}


\begin{figure}[H]
    \centering
    \includegraphics[width=\linewidth]{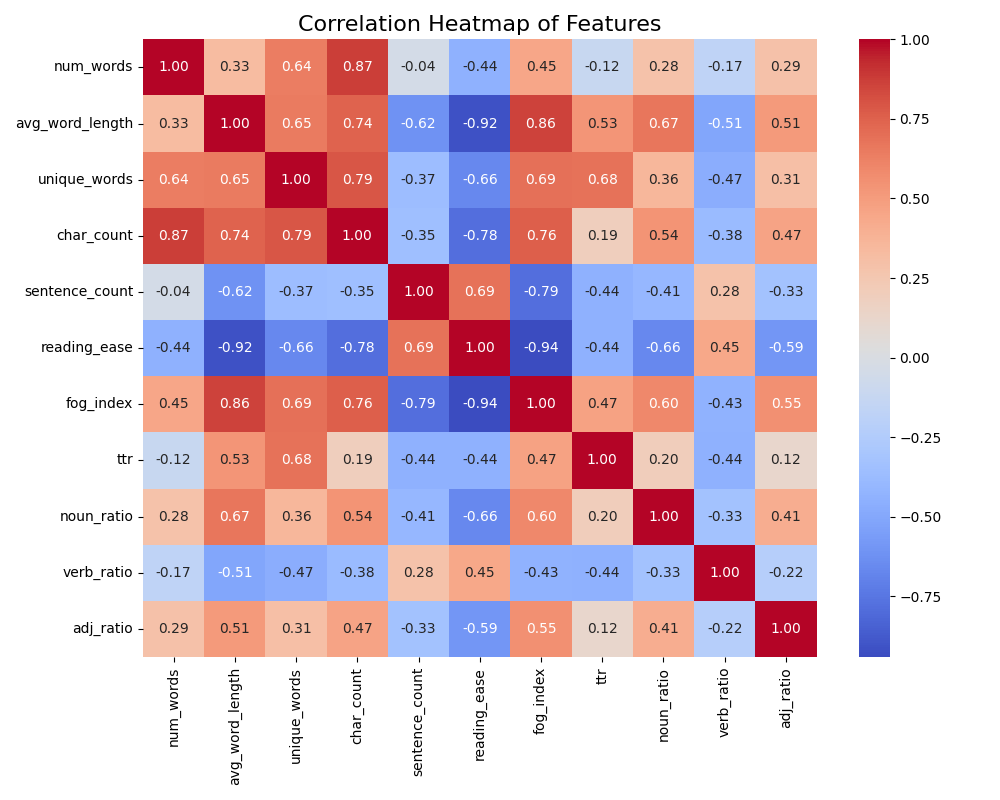}
    \caption{Llama3.1-70B-Instruct text features heatmap}
    \label{fig:human_heatmap}
\end{figure}

\begin{figure}[H]
    \centering
    \includegraphics[width=\linewidth]{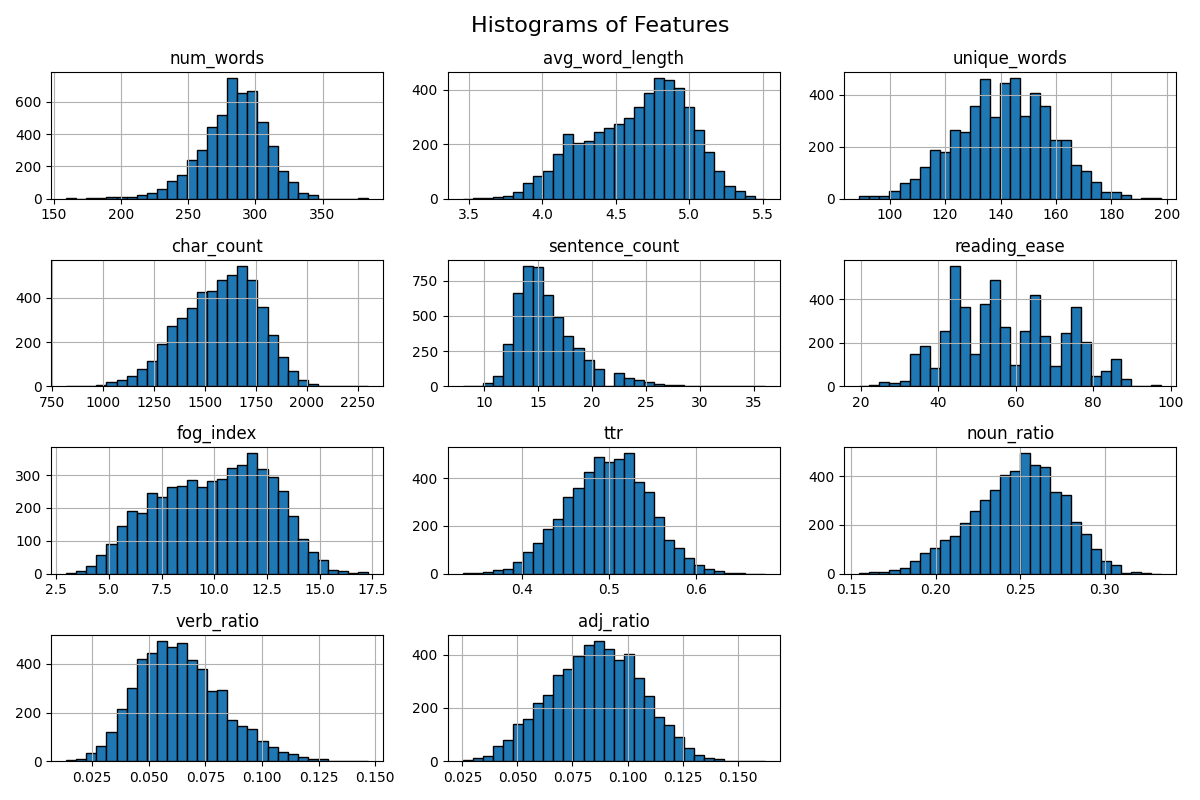}
    \caption{Llama3.1-70B-Instruct text features histgrams}
    \label{fig:human_hist}
\end{figure}

\begin{figure}[H]
    \centering
    \includegraphics[width=\linewidth]{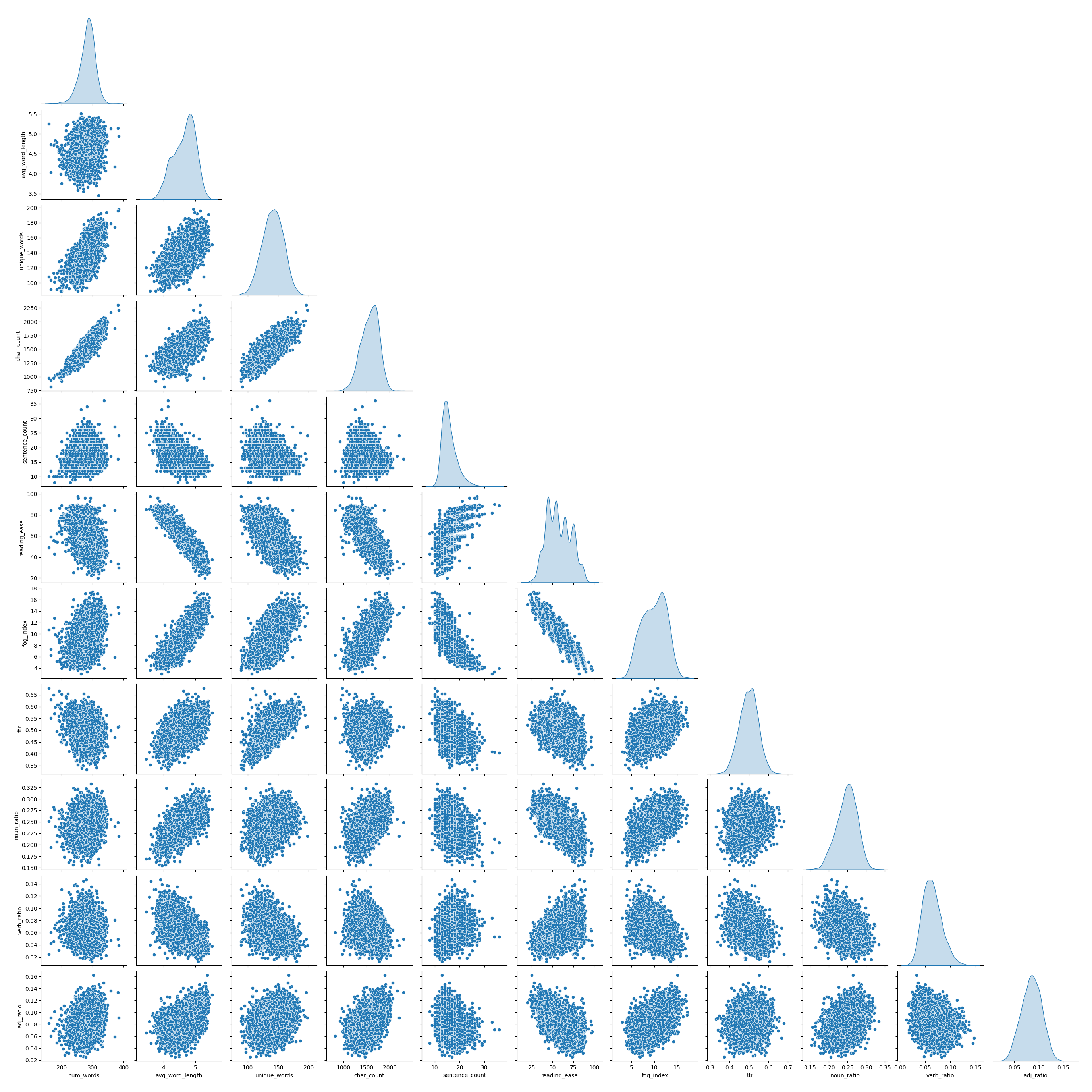}
    \caption{Llama3.1-70B-Instruct text features pairplot}
    \label{fig:human_pairplot}
\end{figure}


\begin{figure}[H]
    \centering
    \includegraphics[width=\linewidth]{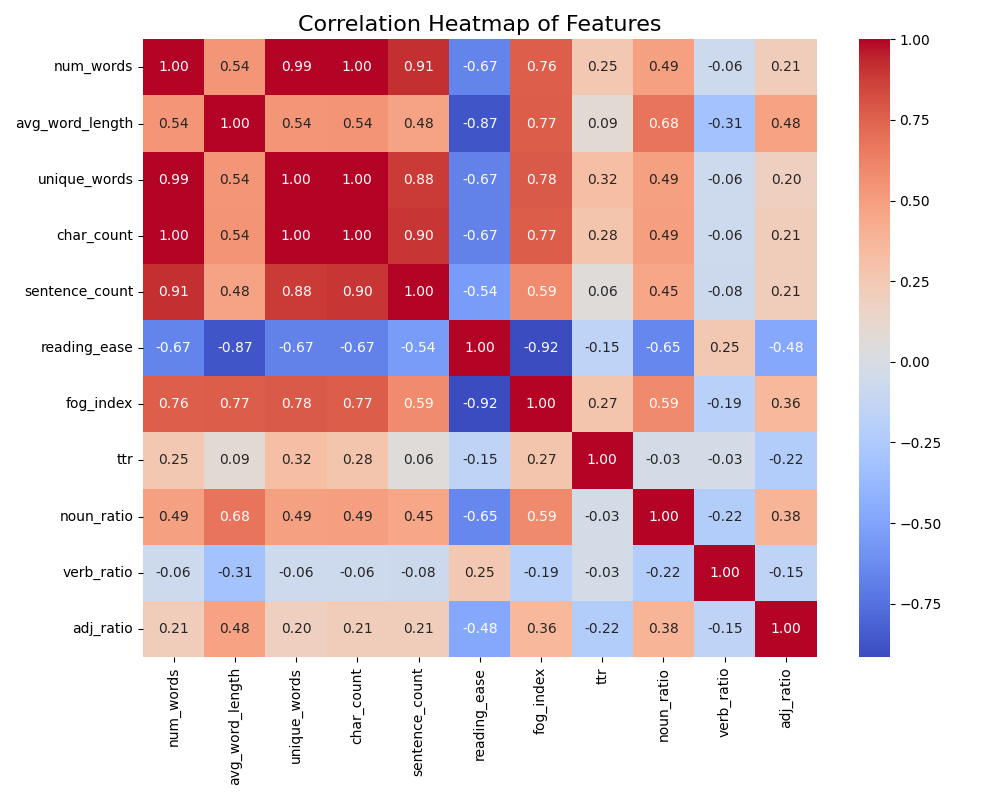}
    \caption{Llama3.2-1B-Instruct text features heatmap}
    \label{fig:human_heatmap}
\end{figure}

\begin{figure}[H]
    \centering
    \includegraphics[width=\linewidth]{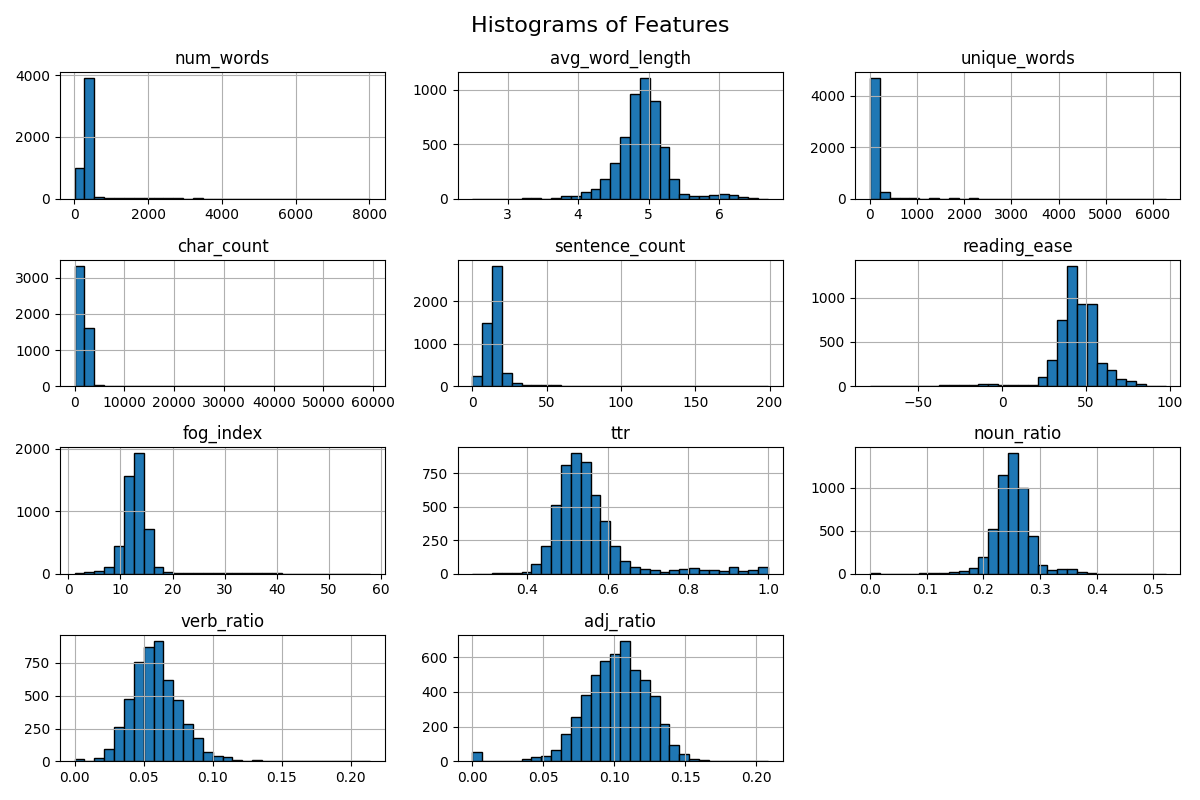}
    \caption{Llama3.2-1B-Instruct text features histgrams}
    \label{fig:human_hist}
\end{figure}

\begin{figure}[H]
    \centering
    \includegraphics[width=\linewidth]{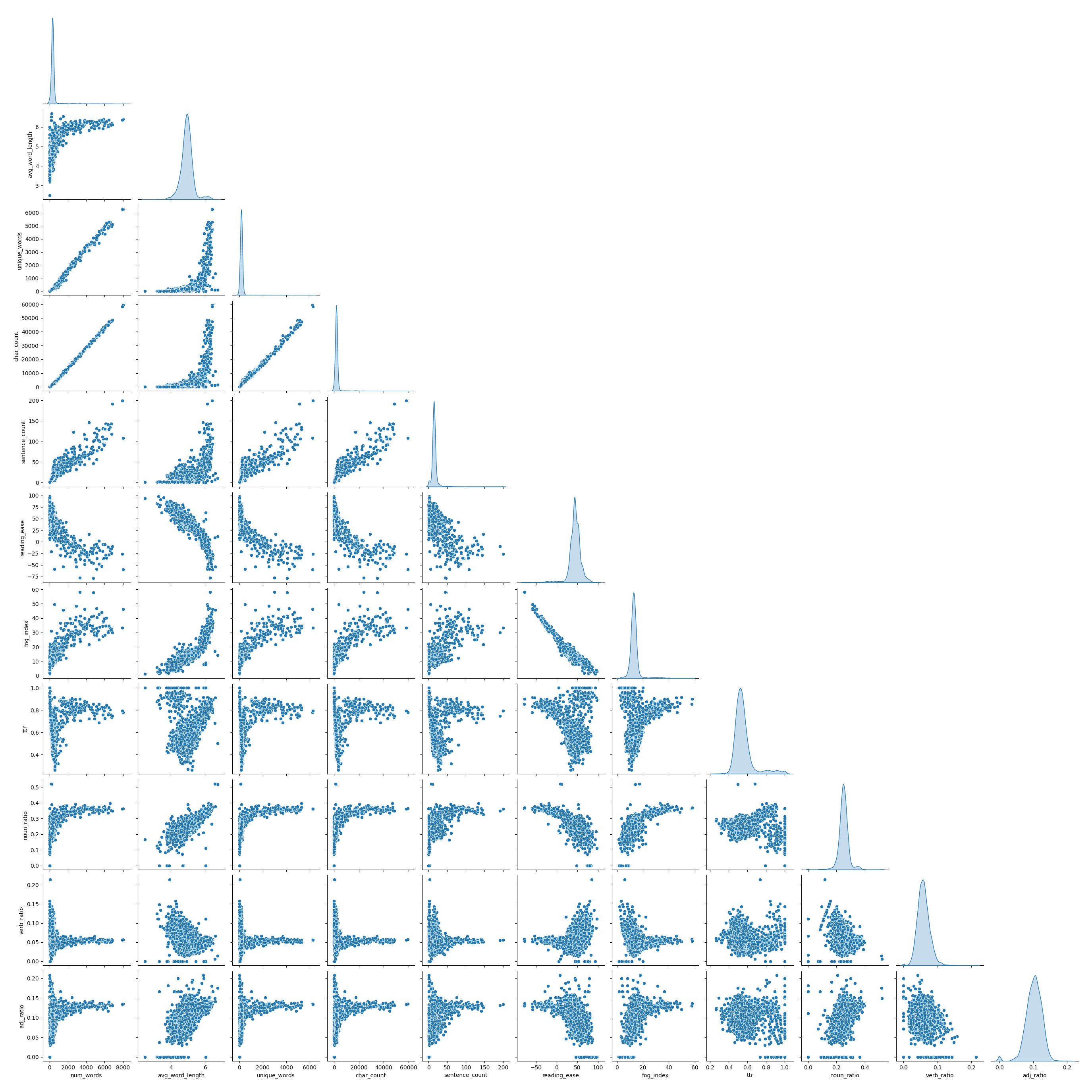}
    \caption{Llama3.2-1B-Instruct text features pairplot}
    \label{fig:human_pairplot}
\end{figure}


\begin{figure}[H]
    \centering
    \includegraphics[width=\linewidth]{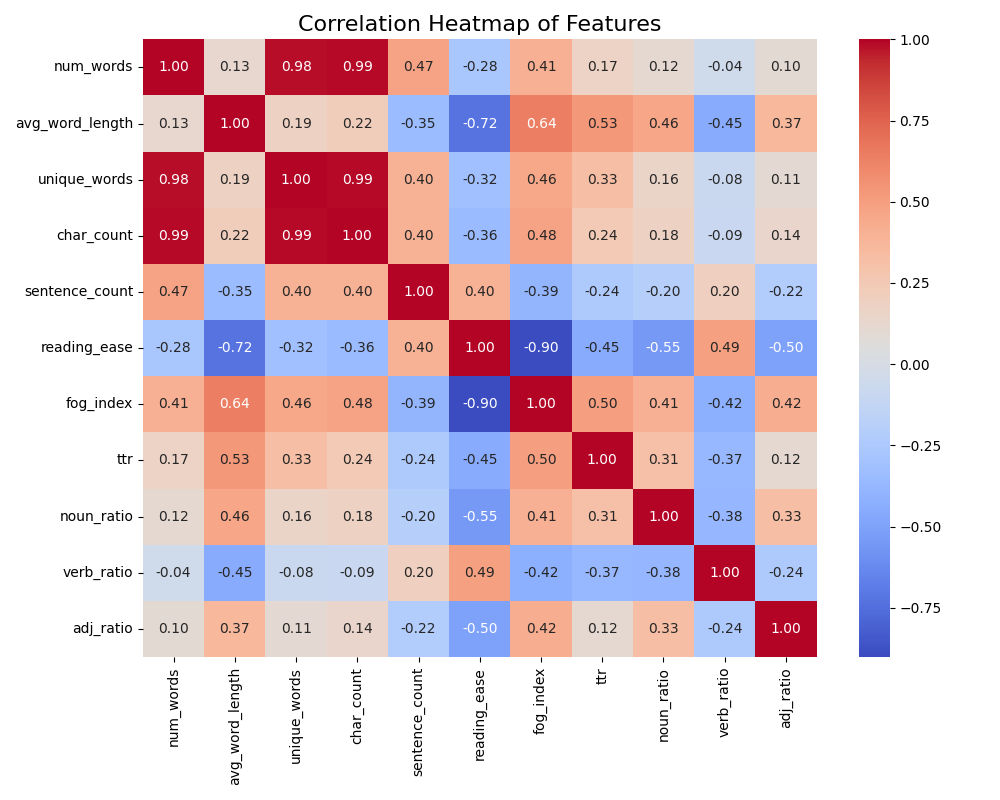}
    \caption{Llama3.2-3B-Instruct text features heatmap}
    \label{fig:human_heatmap}
\end{figure}

\begin{figure}[H]
    \centering
    \includegraphics[width=\linewidth]{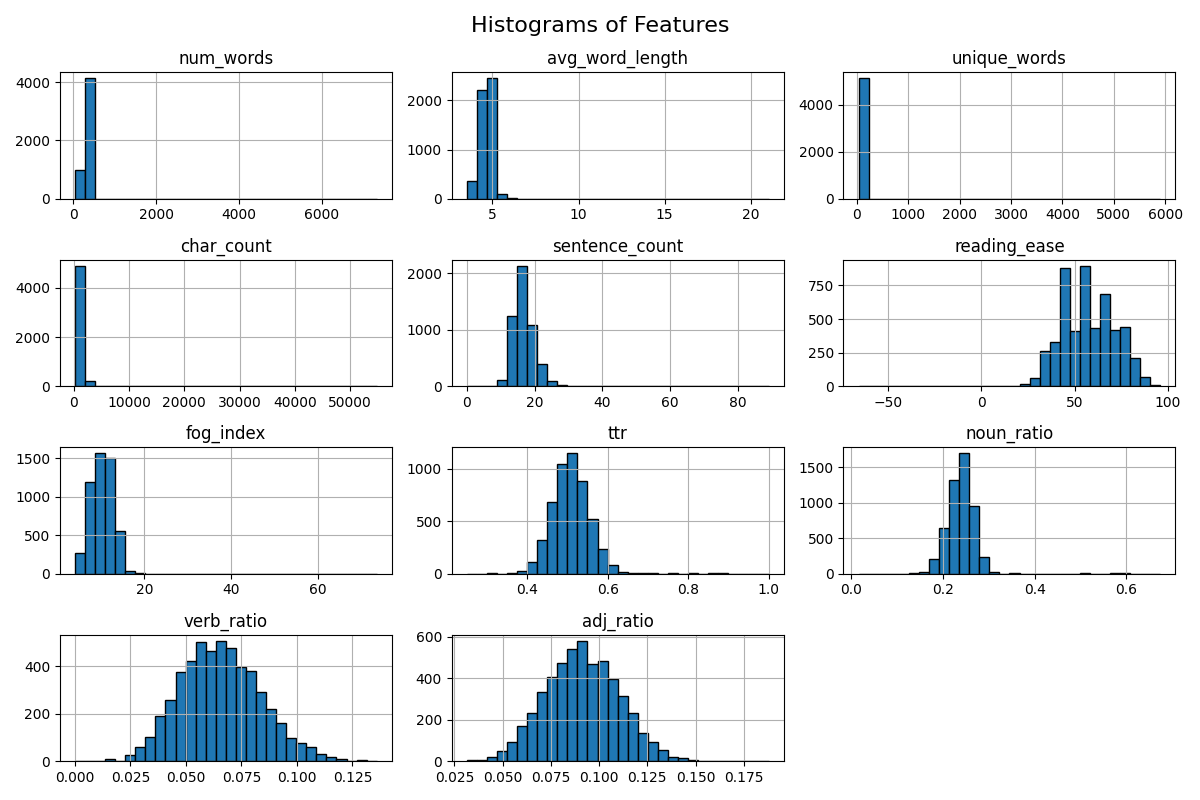}
    \caption{Llama3.2-3B-Instruct text features histgrams}
    \label{fig:human_hist}
\end{figure}

\begin{figure}[H]
    \centering
    \includegraphics[width=\linewidth]{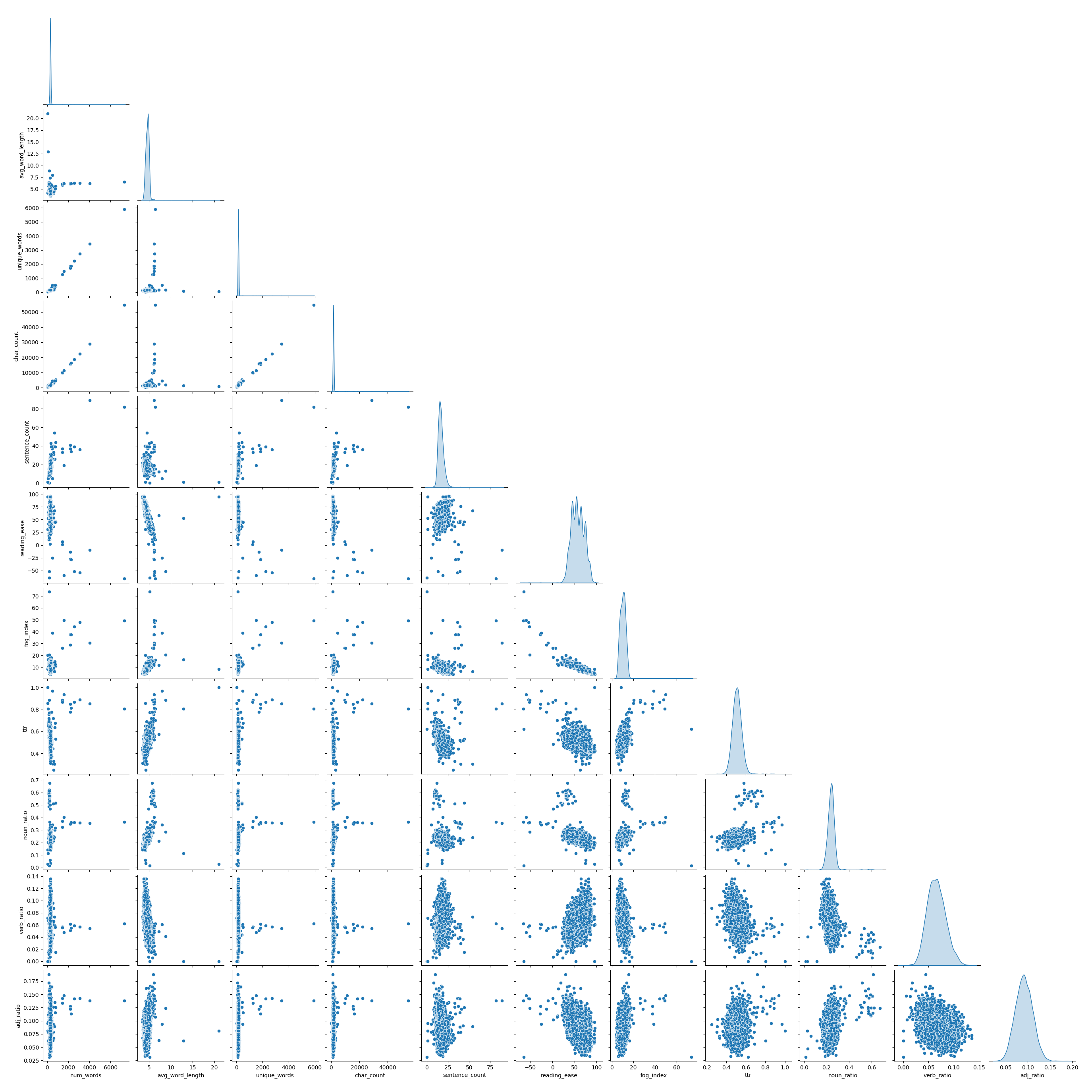}
    \caption{Llama3.2-3B-Instruct text features pairplot}
    \label{fig:human_pairplot}
\end{figure}


\begin{figure}[H]
    \centering
    \includegraphics[width=\linewidth]{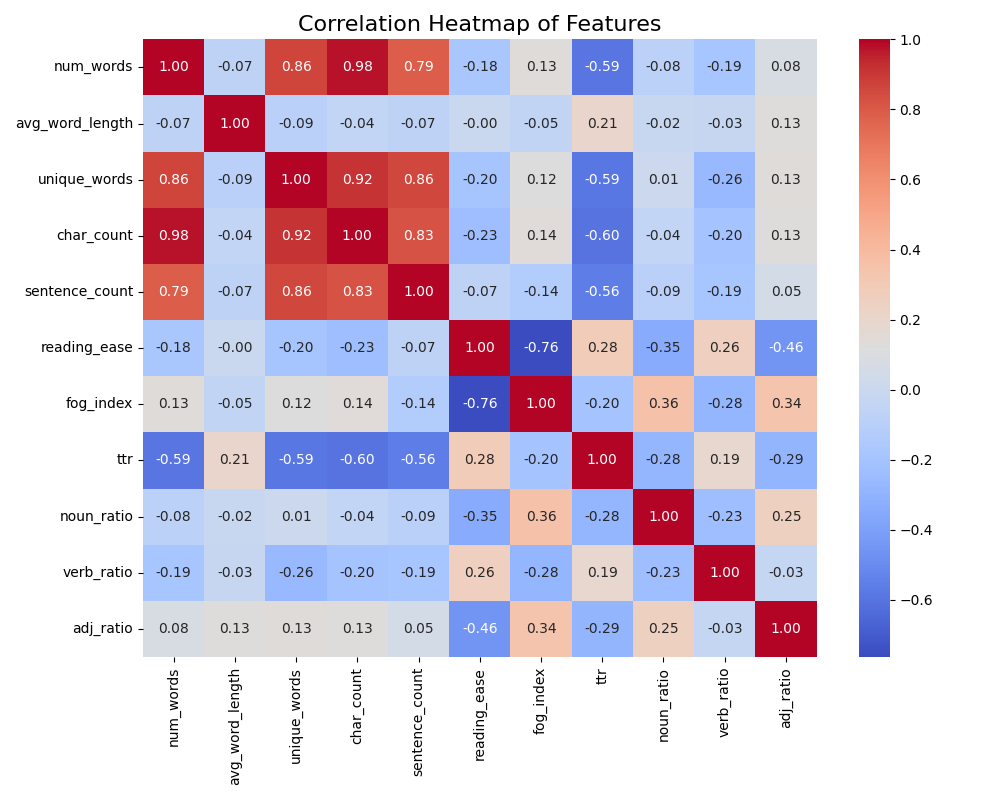}
    \caption{Qwen2.5-0.5B-Instruct text features heatmap}
    \label{fig:human_heatmap}
\end{figure}

\begin{figure}[H]
    \centering
    \includegraphics[width=\linewidth]{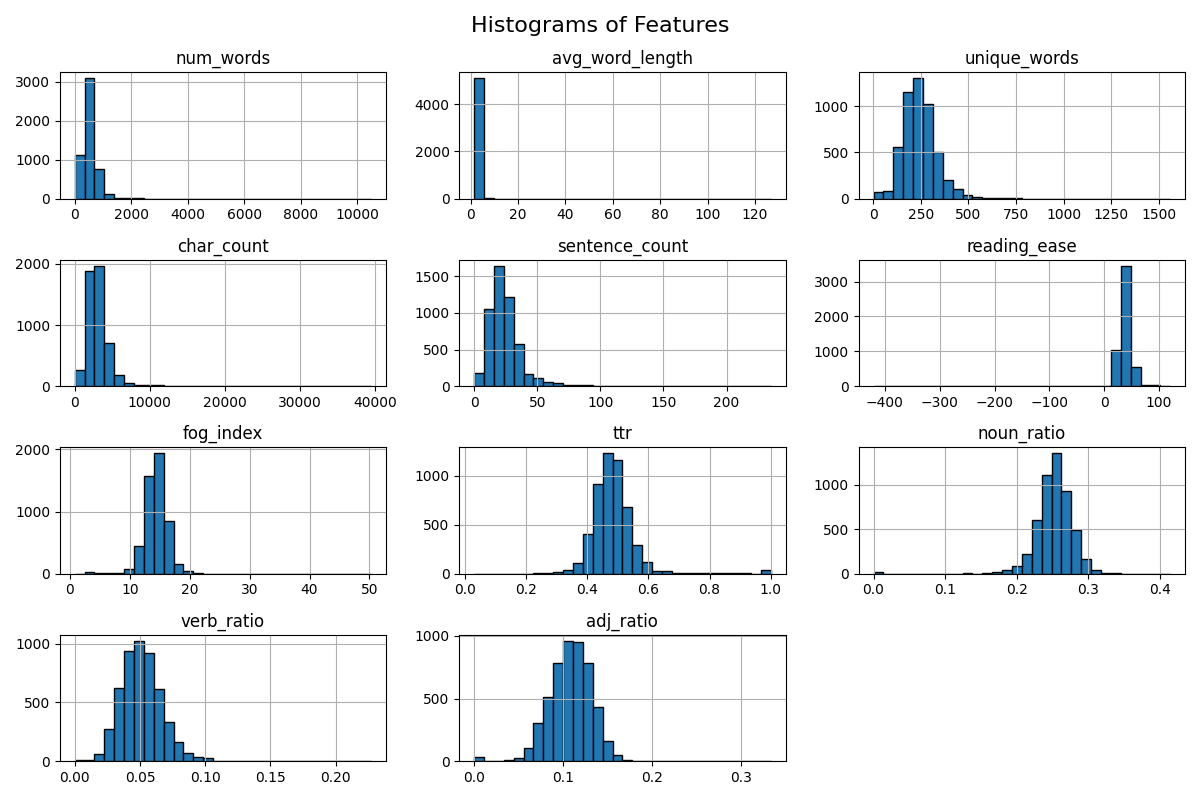}
    \caption{Qwen2.5-0.5B-Instruct text features histgrams}
    \label{fig:human_hist}
\end{figure}

\begin{figure}[H]
    \centering
    \includegraphics[width=\linewidth]{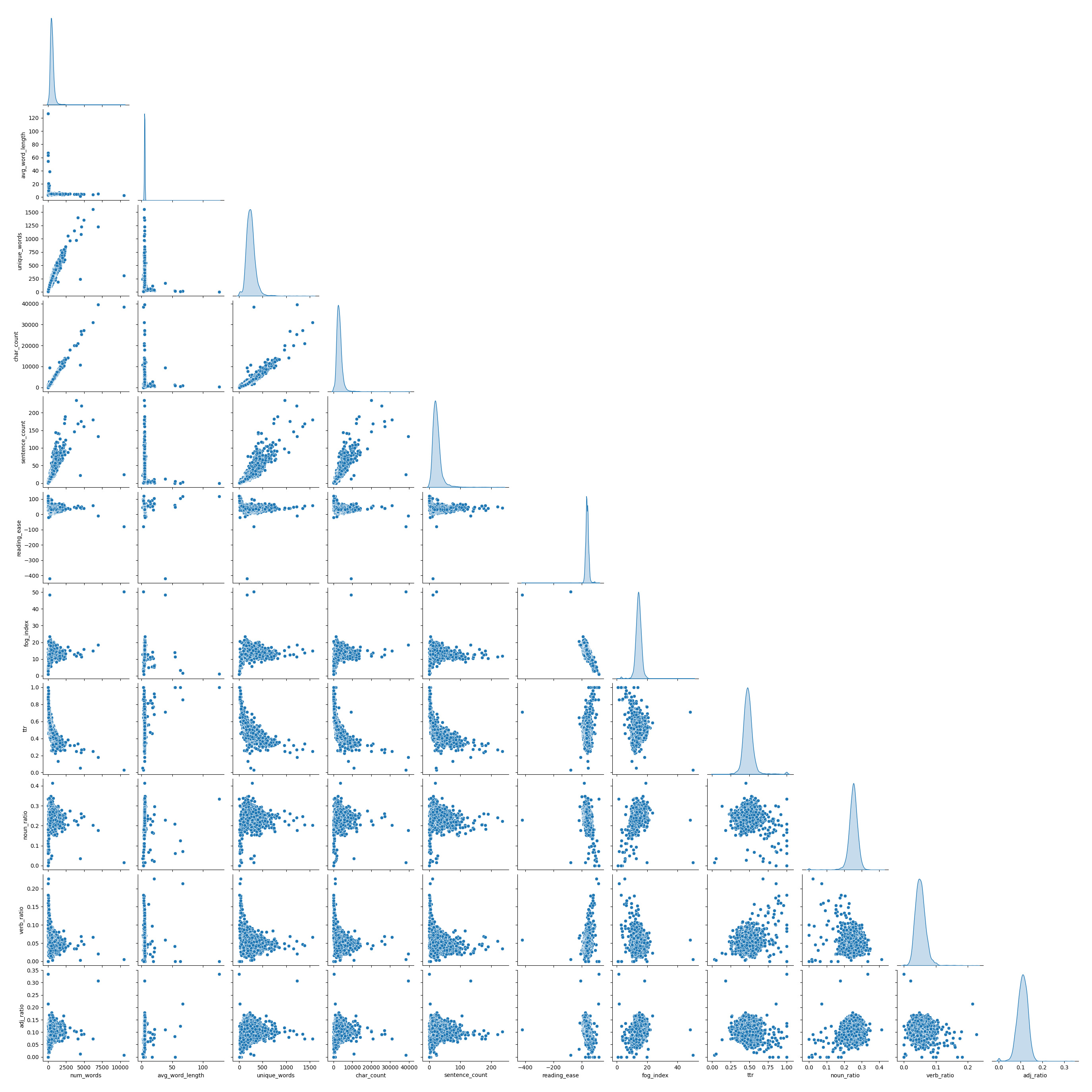}
    \caption{Qwen2.5-0.5B-Instruct text features pairplot}
    \label{fig:human_pairplot}
\end{figure}


\begin{figure}[H]
    \centering
    \includegraphics[width=\linewidth]{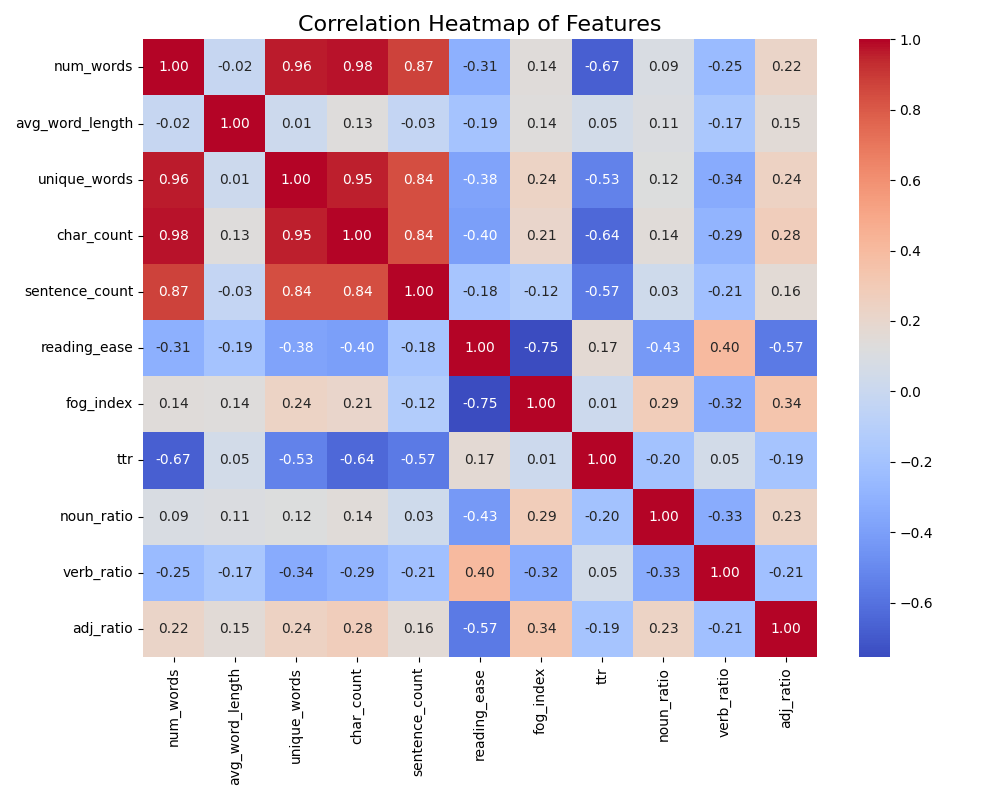}
    \caption{Qwen2.5-1.5B-Instruct text features heatmap}
    \label{fig:human_heatmap}
\end{figure}

\begin{figure}[H]
    \centering
    \includegraphics[width=\linewidth]{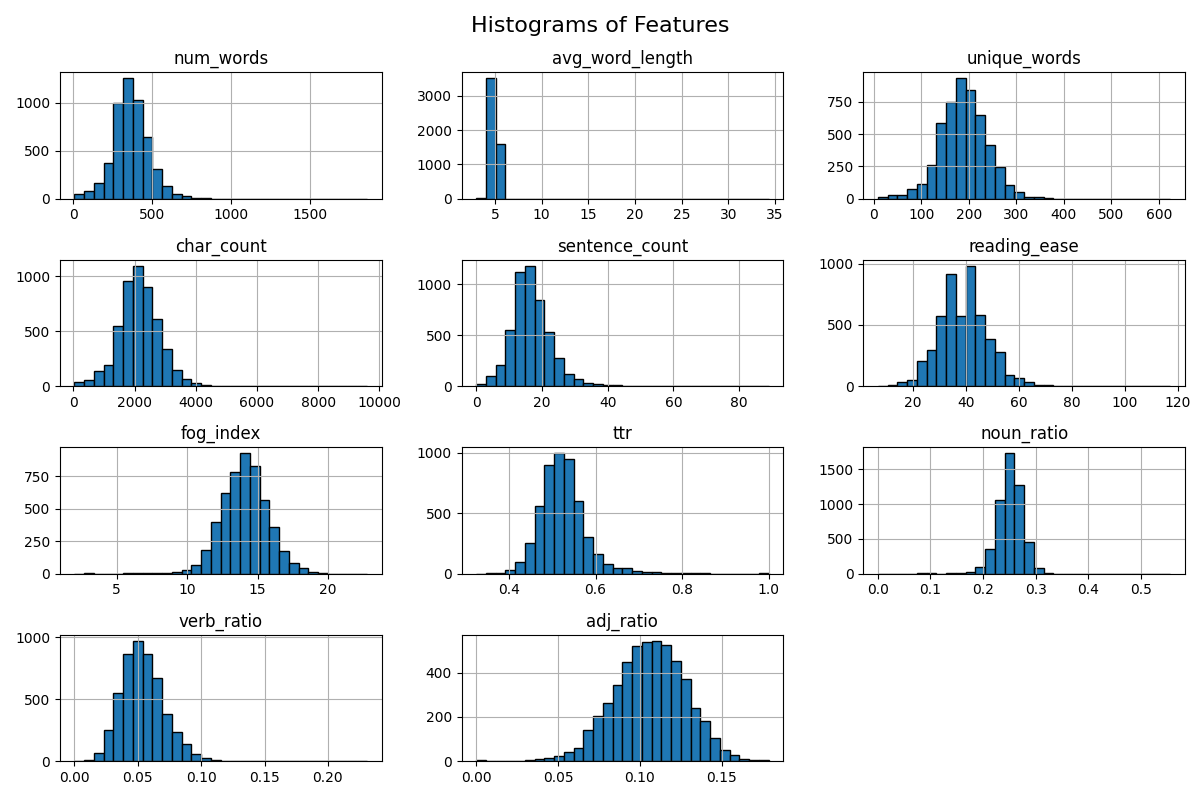}
    \caption{Qwen2.5-1.5B-Instruct text features histgrams}
    \label{fig:human_hist}
\end{figure}

\begin{figure}[H]
    \centering
    \includegraphics[width=\linewidth]{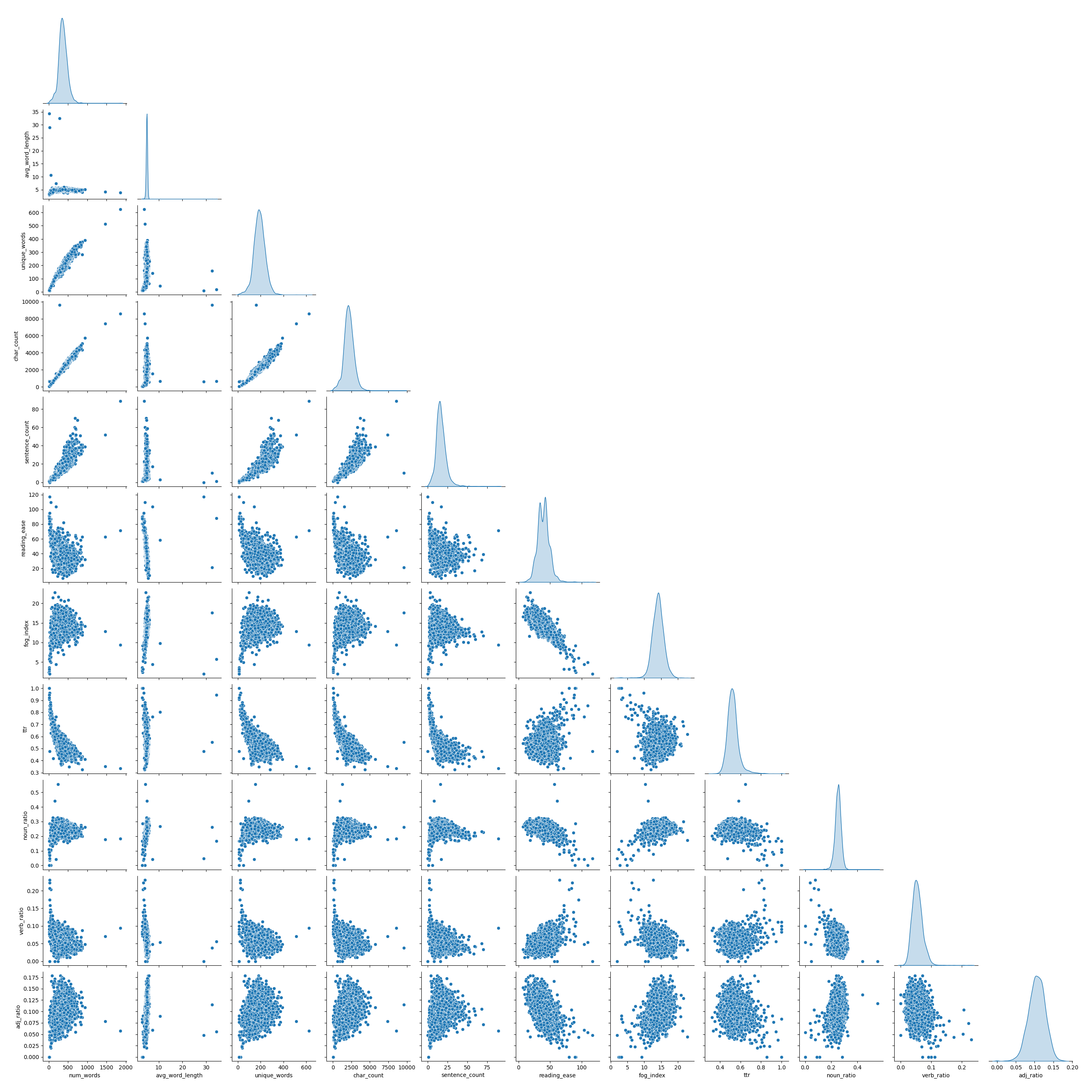}
    \caption{Qwen2.5-1.5B-Instruct text features pairplot}
    \label{fig:human_pairplot}
\end{figure}


\begin{figure}[H]
    \centering
    \includegraphics[width=\linewidth]{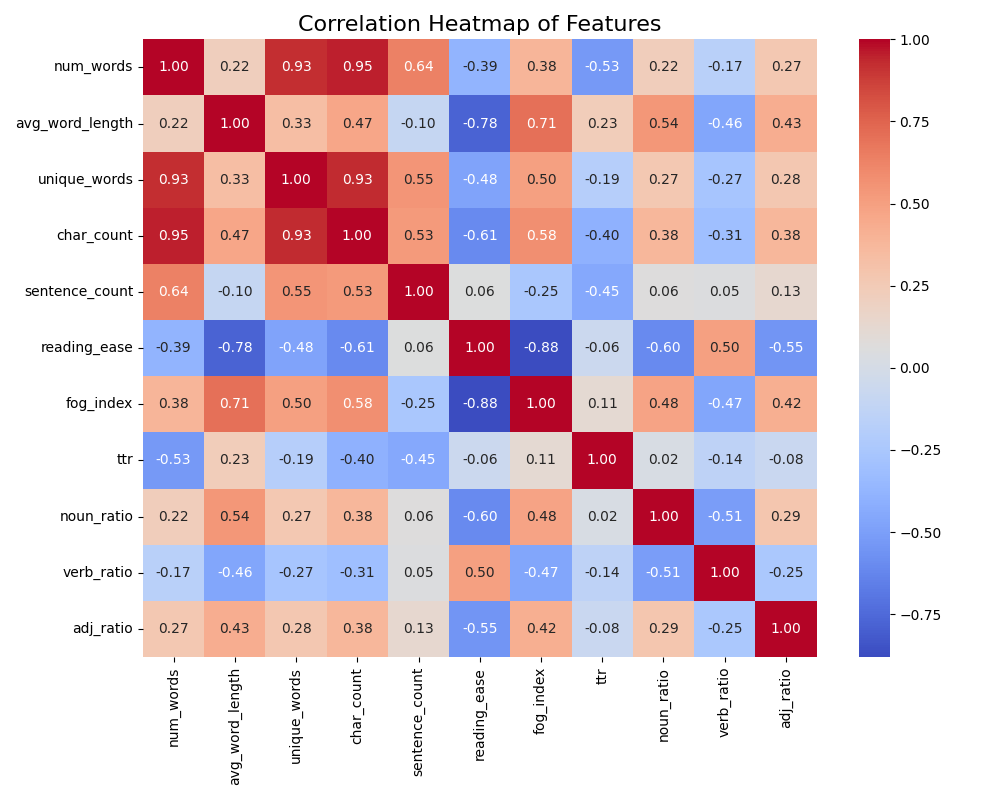}
    \caption{Qwen2.5-3B-Instruct text features heatmap}
    \label{fig:human_heatmap}
\end{figure}

\begin{figure}[H]
    \centering
    \includegraphics[width=\linewidth]{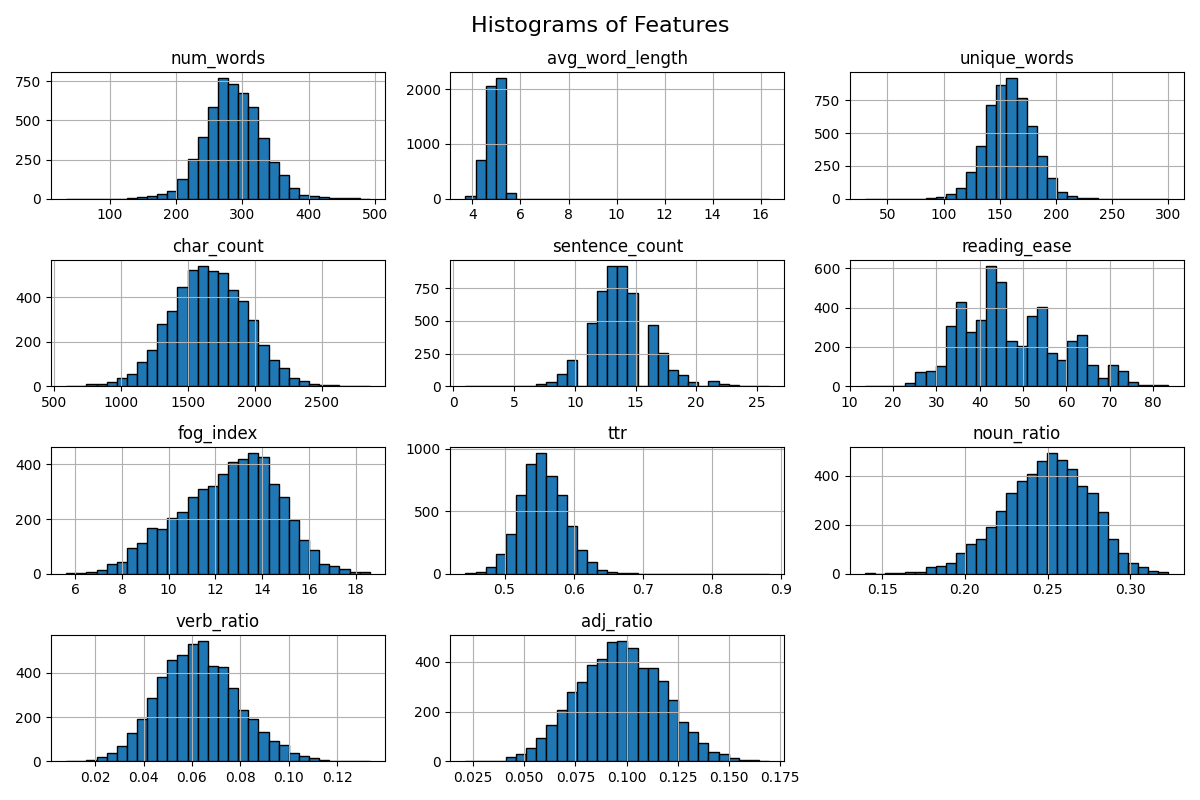}
    \caption{Qwen2.5-3B-Instruct text features histgrams}
    \label{fig:human_hist}
\end{figure}

\begin{figure}[H]
    \centering
    \includegraphics[width=\linewidth]{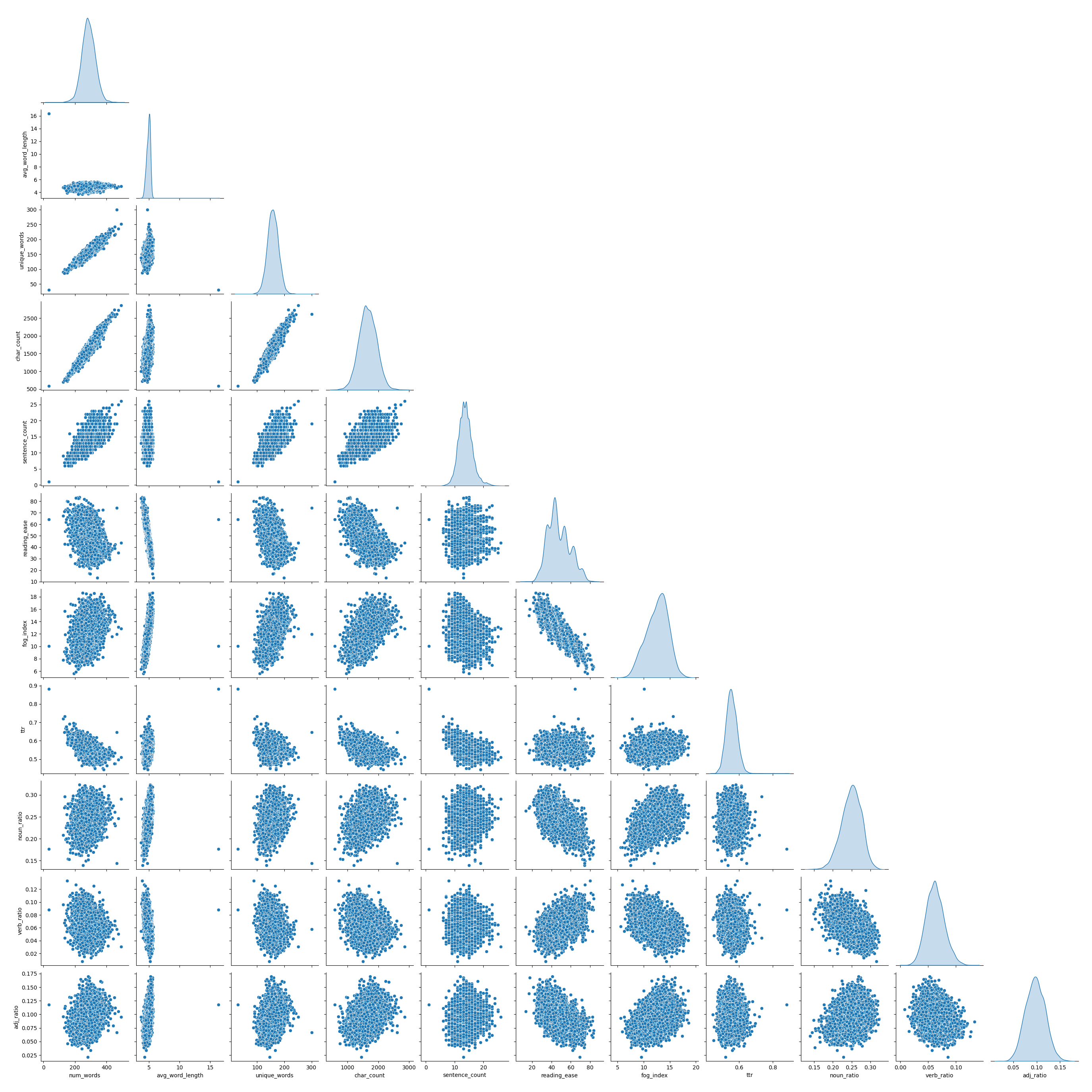}
    \caption{Qwen2.5-3B-Instruct text features pairplot}
    \label{fig:human_pairplot}
\end{figure}


\begin{figure}[H]
    \centering
    \includegraphics[width=\linewidth]{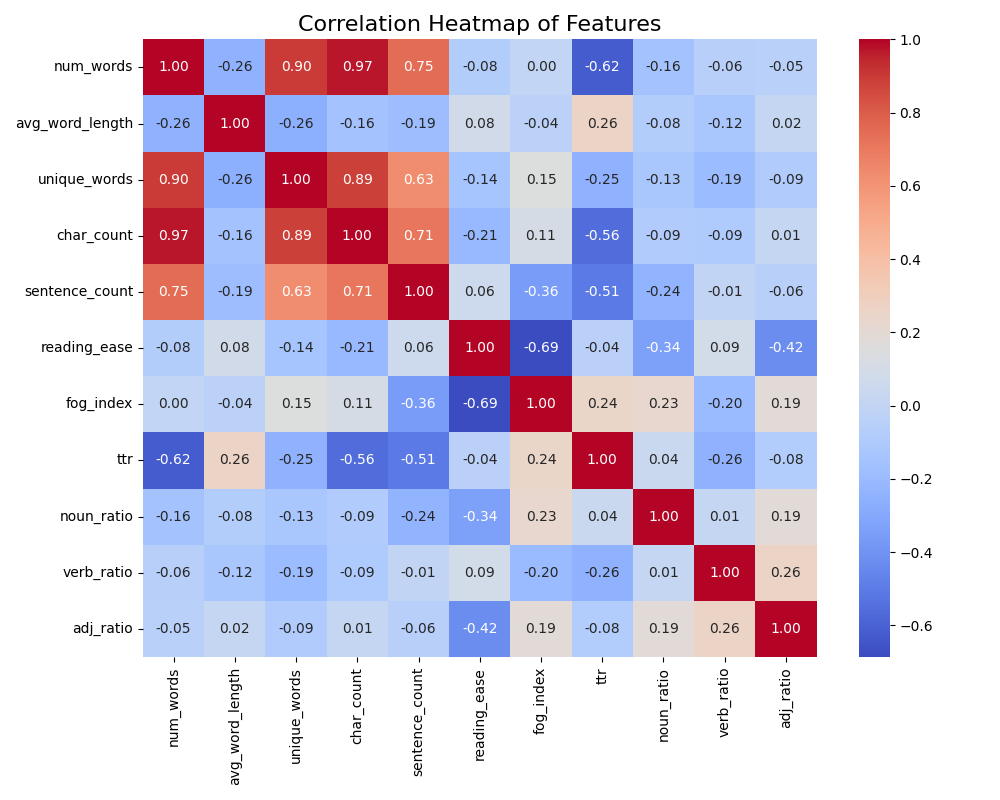}
    \caption{Qwen2.5-7B-Instruct text features heatmap}
    \label{fig:human_heatmap}
\end{figure}

\begin{figure}[H]
    \centering
    \includegraphics[width=\linewidth]{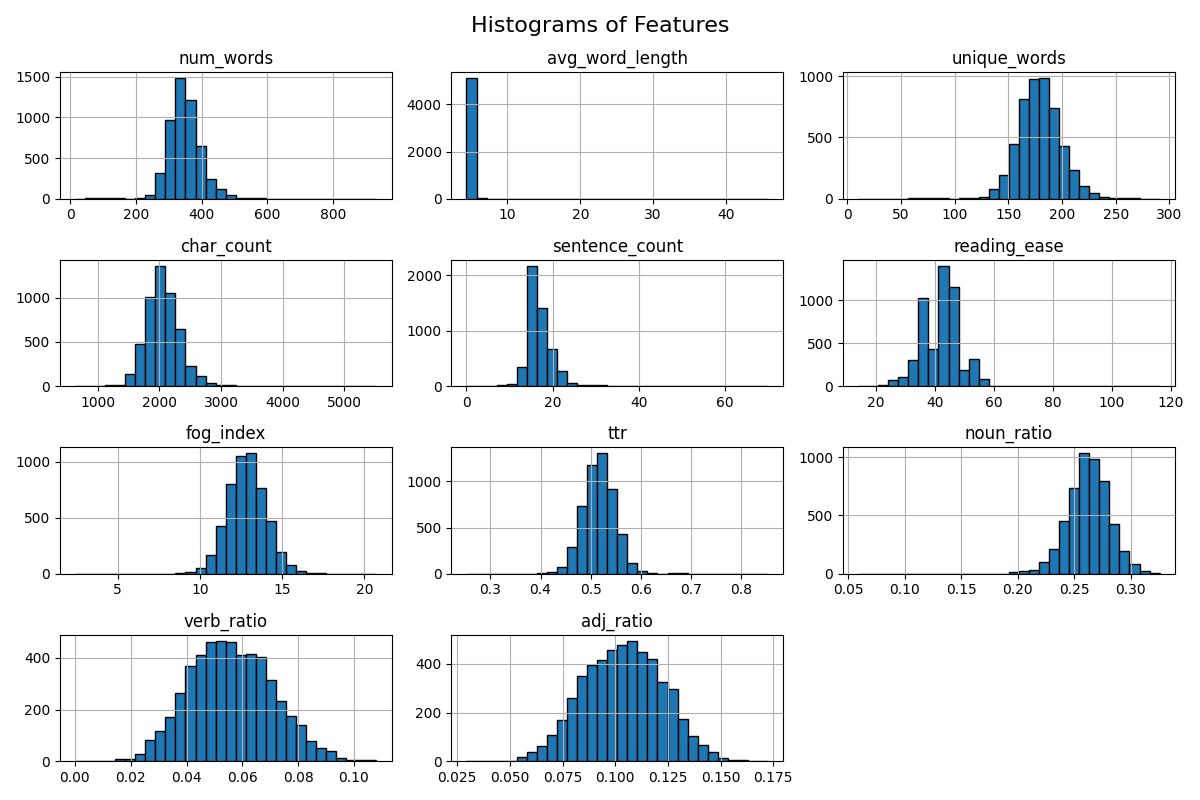}
    \caption{Qwen2.5-7B-Instruct text features histgrams}
    \label{fig:human_hist}
\end{figure}

\begin{figure}[H]
    \centering
    \includegraphics[width=\linewidth]{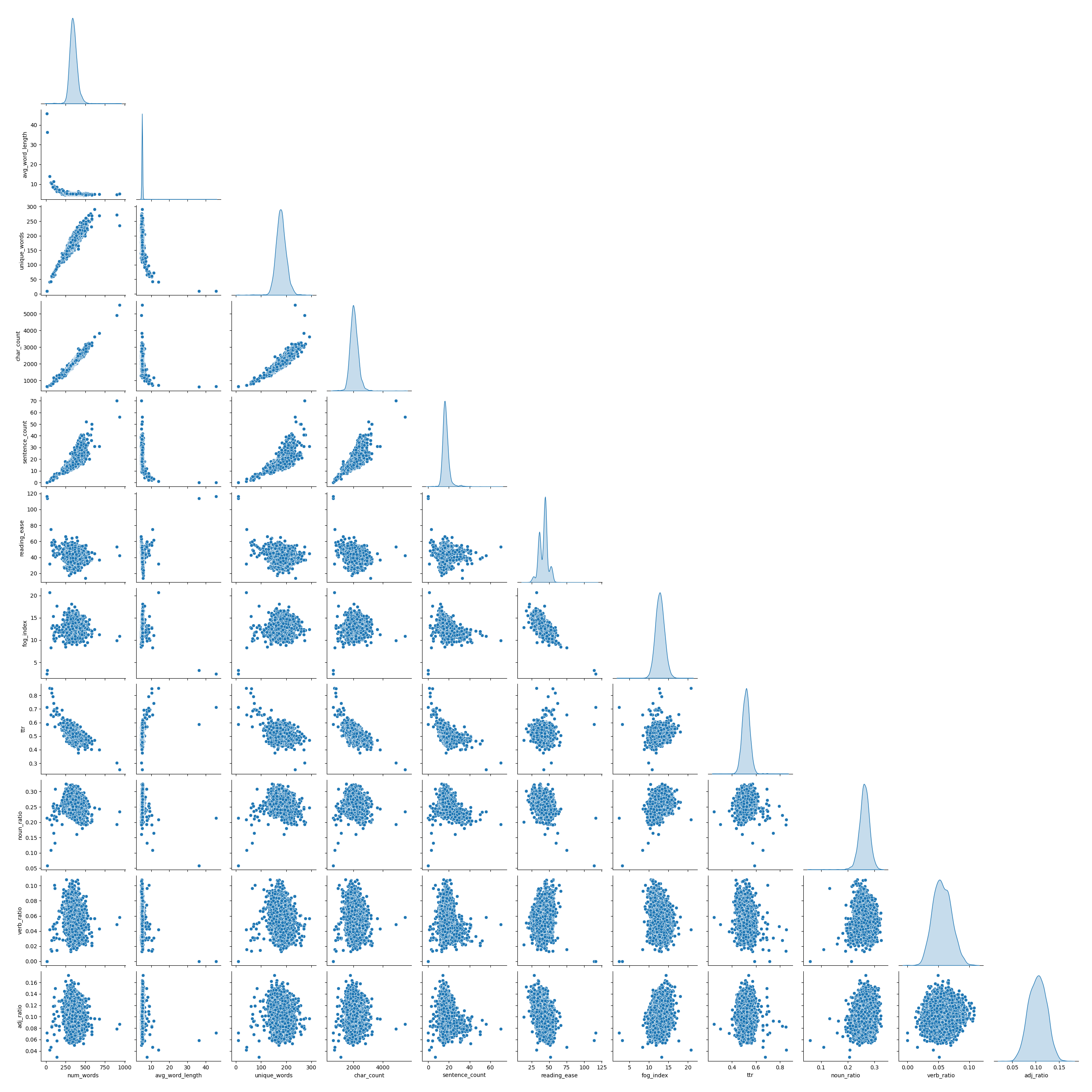}
    \caption{Qwen2.5-7B-Instruct text features pairplot}
    \label{fig:human_pairplot}
\end{figure}


\begin{figure}[H]
    \centering
    \includegraphics[width=\linewidth]{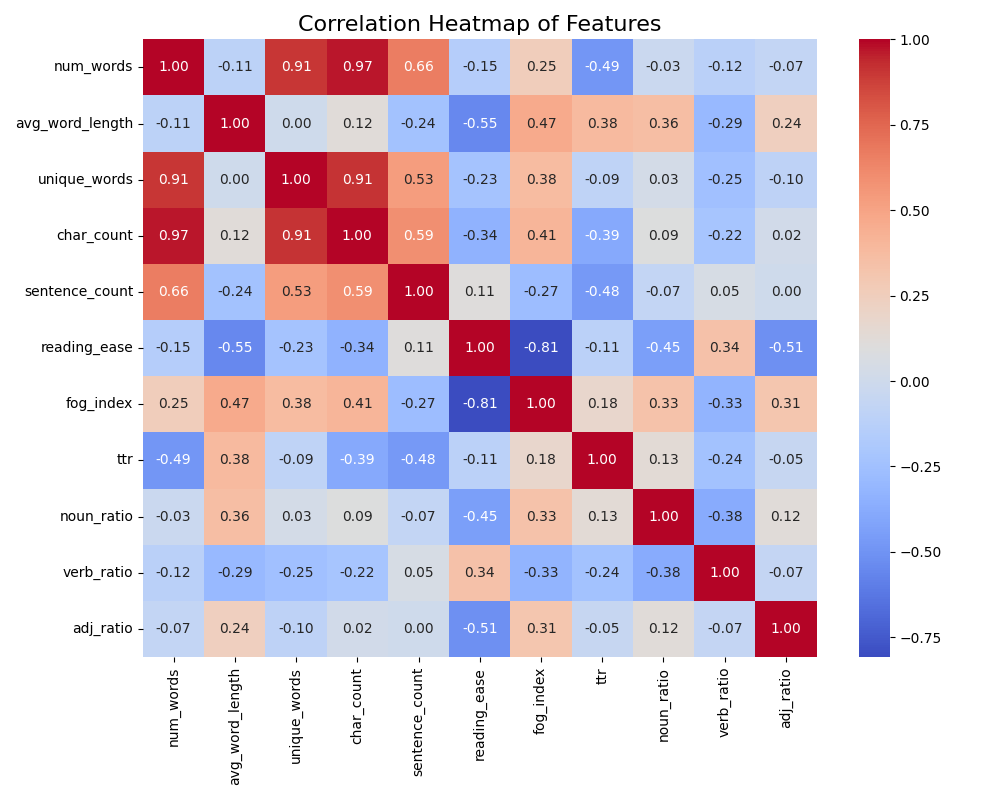}
    \caption{Qwen2.5-14B-Instruct text features heatmap}
    \label{fig:human_heatmap}
\end{figure}

\begin{figure}[H]
    \centering
    \includegraphics[width=\linewidth]{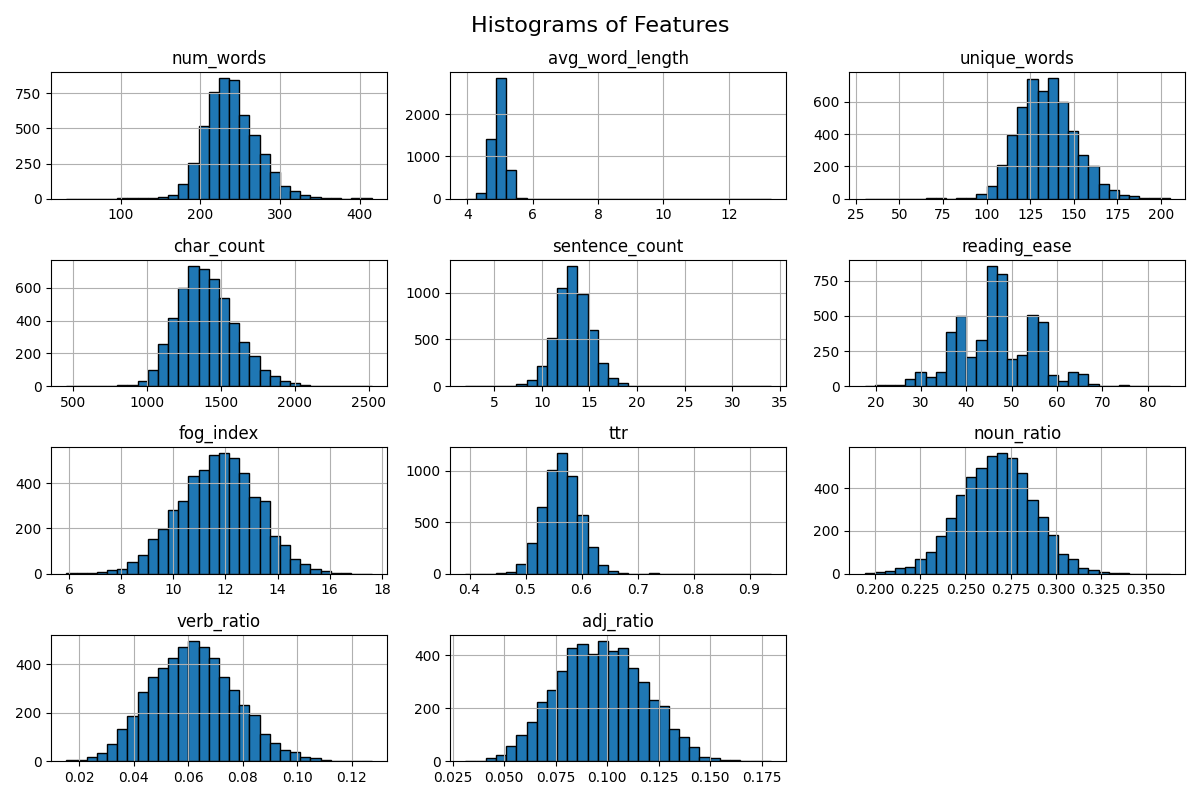}
    \caption{Qwen2.5-14B-Instruct text features histgrams}
    \label{fig:human_hist}
\end{figure}

\begin{figure}[H]
    \centering
    \includegraphics[width=\linewidth]{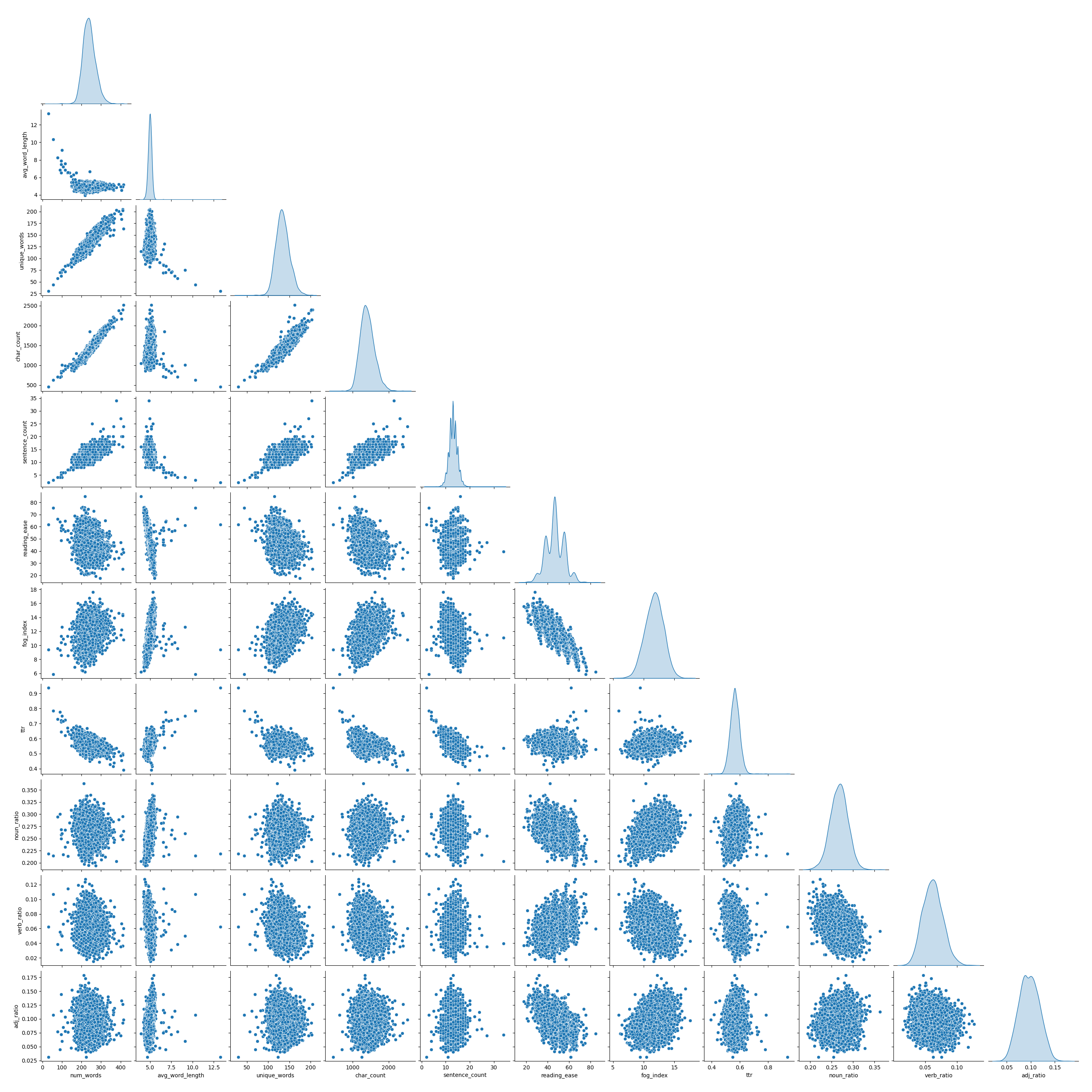}
    \caption{Qwen2.5-14B-Instruct text features pairplot}
    \label{fig:human_pairplot}
\end{figure}


\begin{figure}[H]
    \centering
    \includegraphics[width=\linewidth]{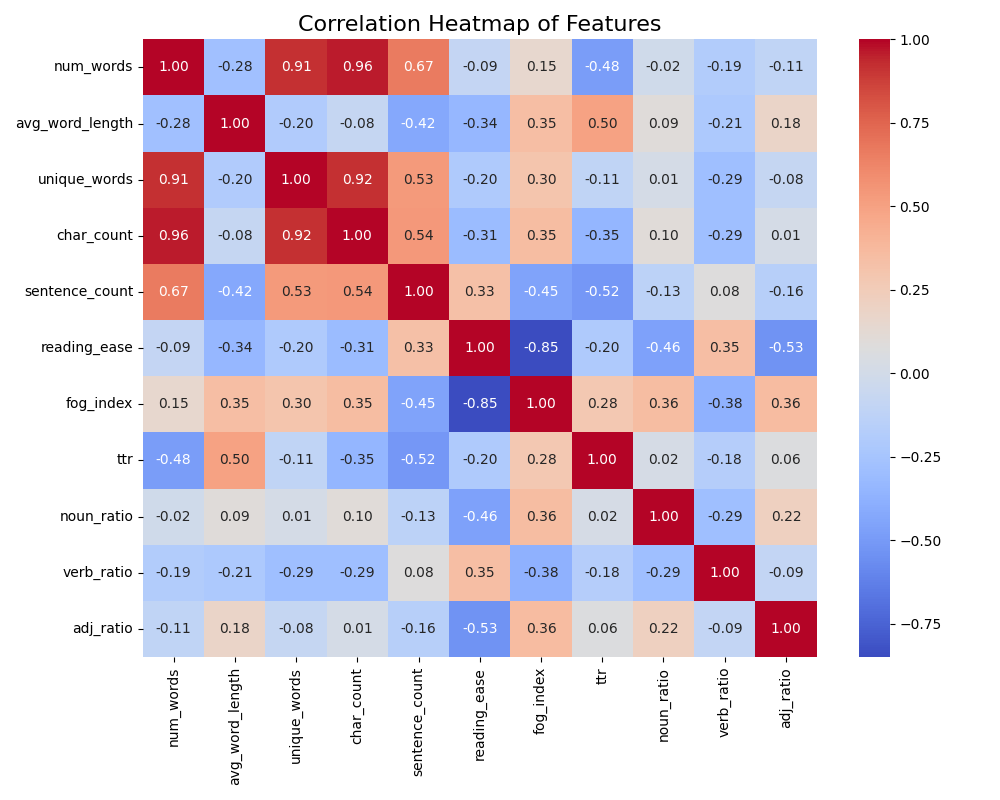}
    \caption{Qwen2.5-32B-Instruct text features heatmap}
    \label{fig:human_heatmap}
\end{figure}

\begin{figure}[H]
    \centering
    \includegraphics[width=\linewidth]{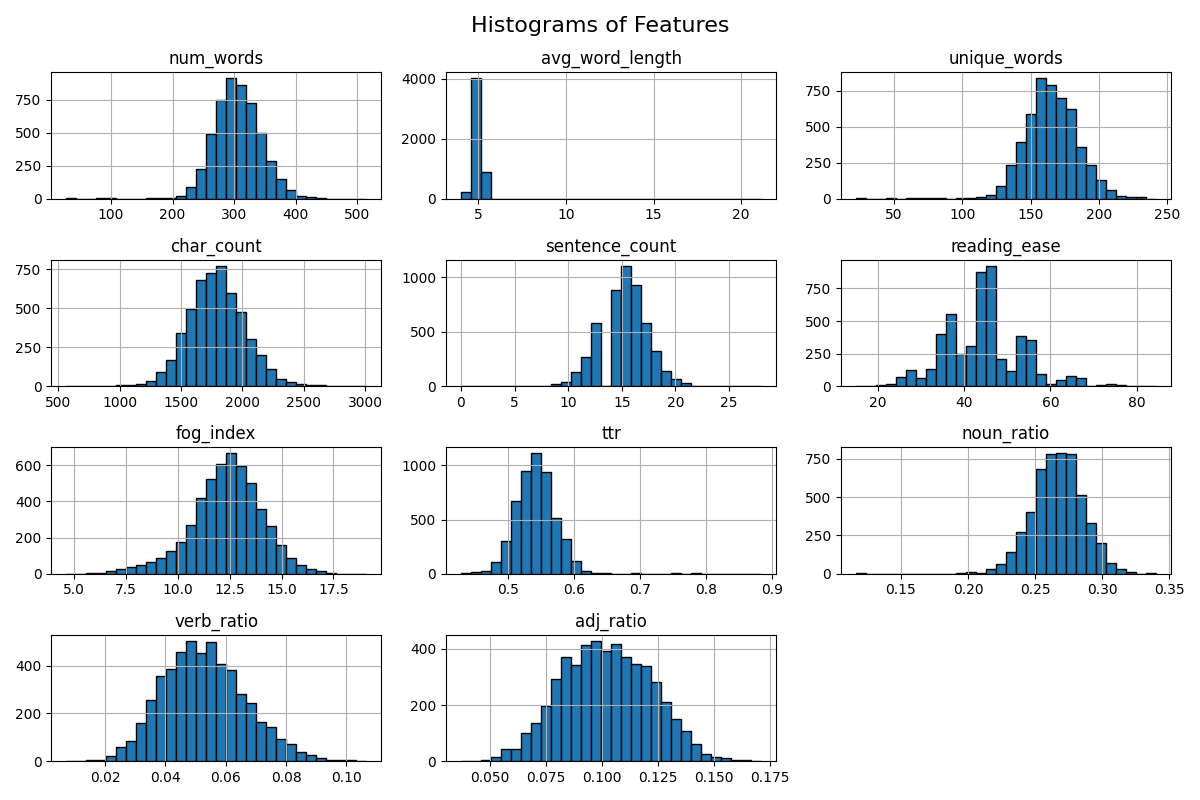}
    \caption{Qwen2.5-32B-Instruct text features histgrams}
    \label{fig:human_hist}
\end{figure}

\begin{figure}[H]
    \centering
    \includegraphics[width=\linewidth]{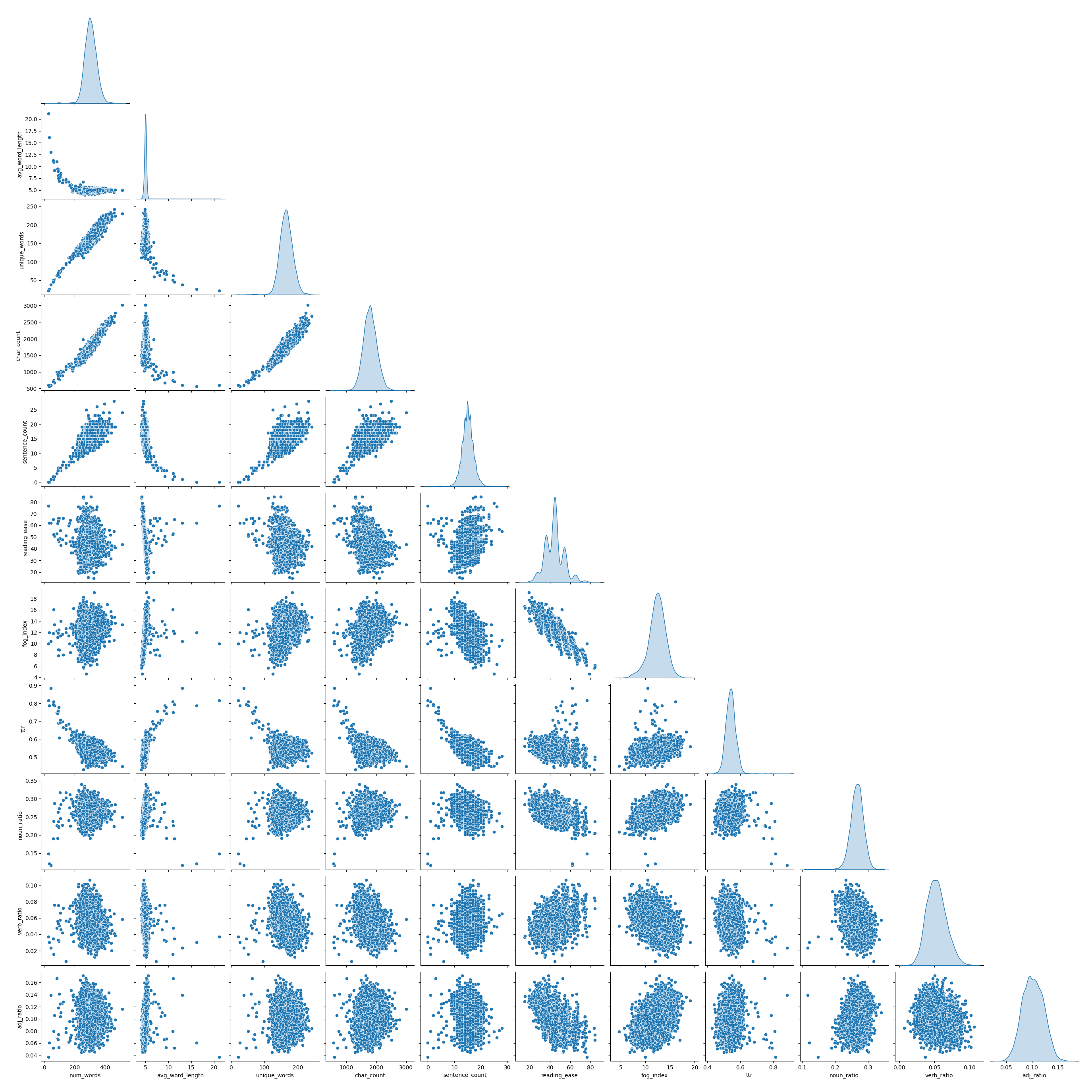}
    \caption{Qwen2.5-32B-Instruct text features pairplot}
    \label{fig:human_pairplot}
\end{figure}


\begin{figure}[H]
    \centering
    \includegraphics[width=\linewidth]{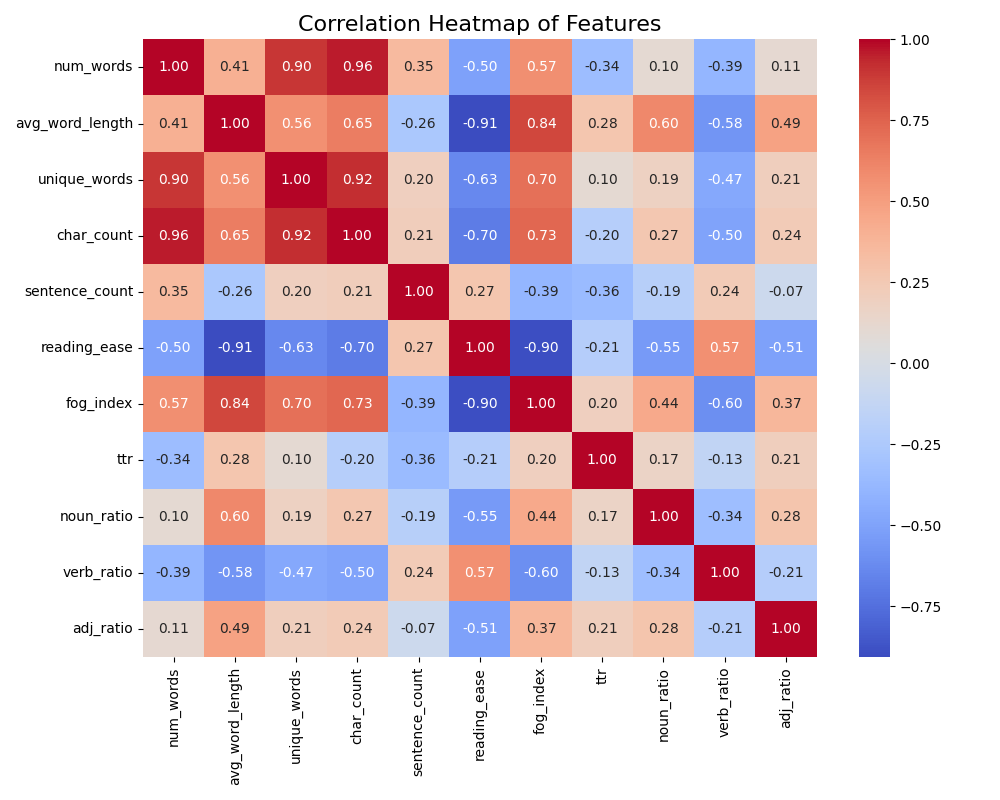}
    \caption{Qwen2.5-72B-Instruct text features heatmap}
    \label{fig:human_heatmap}
\end{figure}

\begin{figure}[H]
    \centering
    \includegraphics[width=\linewidth]{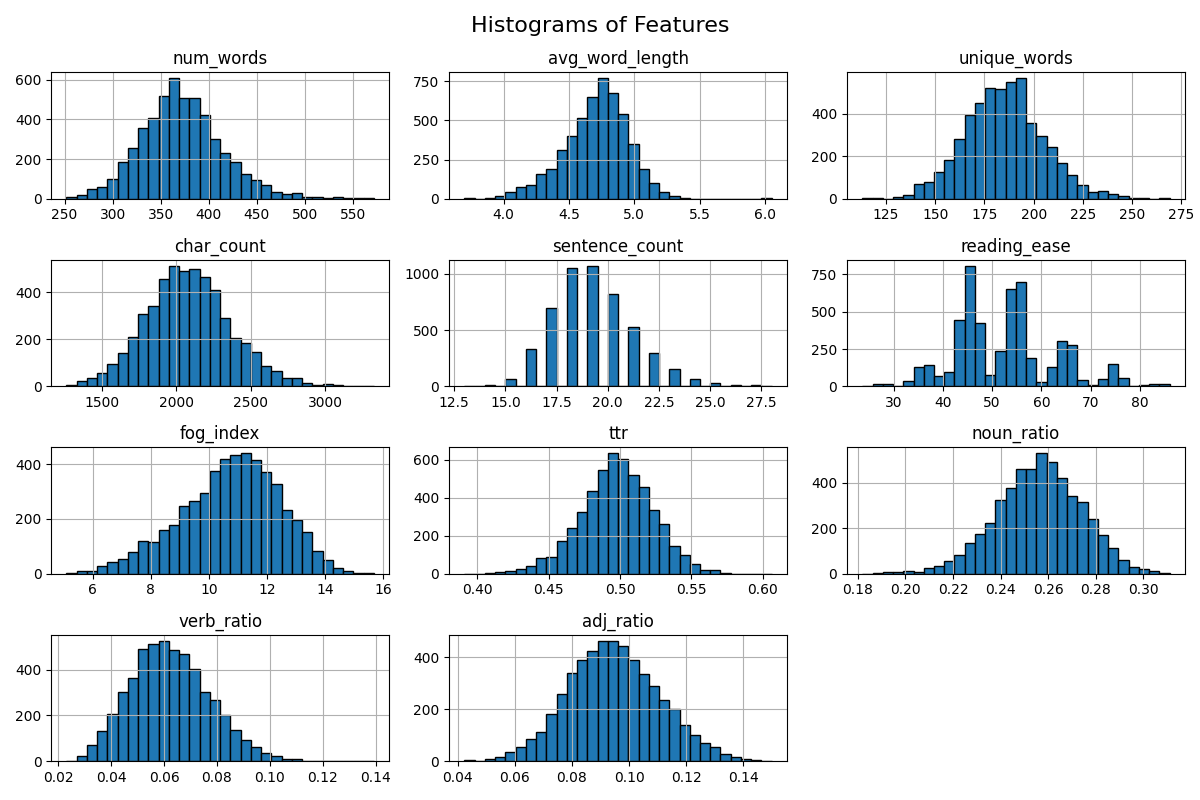}
    \caption{Qwen2.5-72B-Instruct text features histgrams}
    \label{fig:human_hist}
\end{figure}

\begin{figure}[H]
    \centering
    \includegraphics[width=\linewidth]{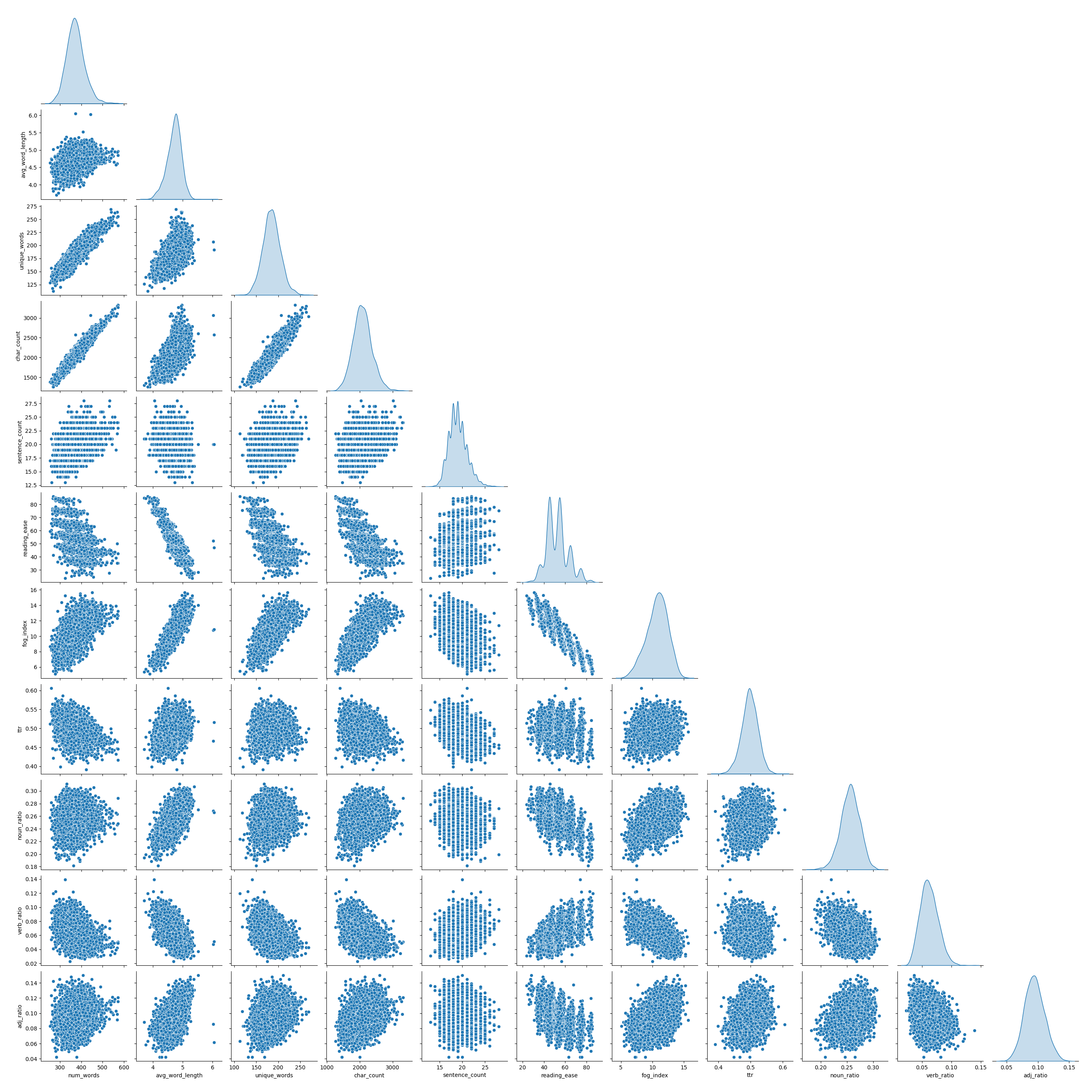}
    \caption{Qwen2.5-72B-Instruct text features pairplot}
    \label{fig:human_pairplot}
\end{figure}


\begin{figure}[H]
    \centering
    \includegraphics[width=\linewidth]{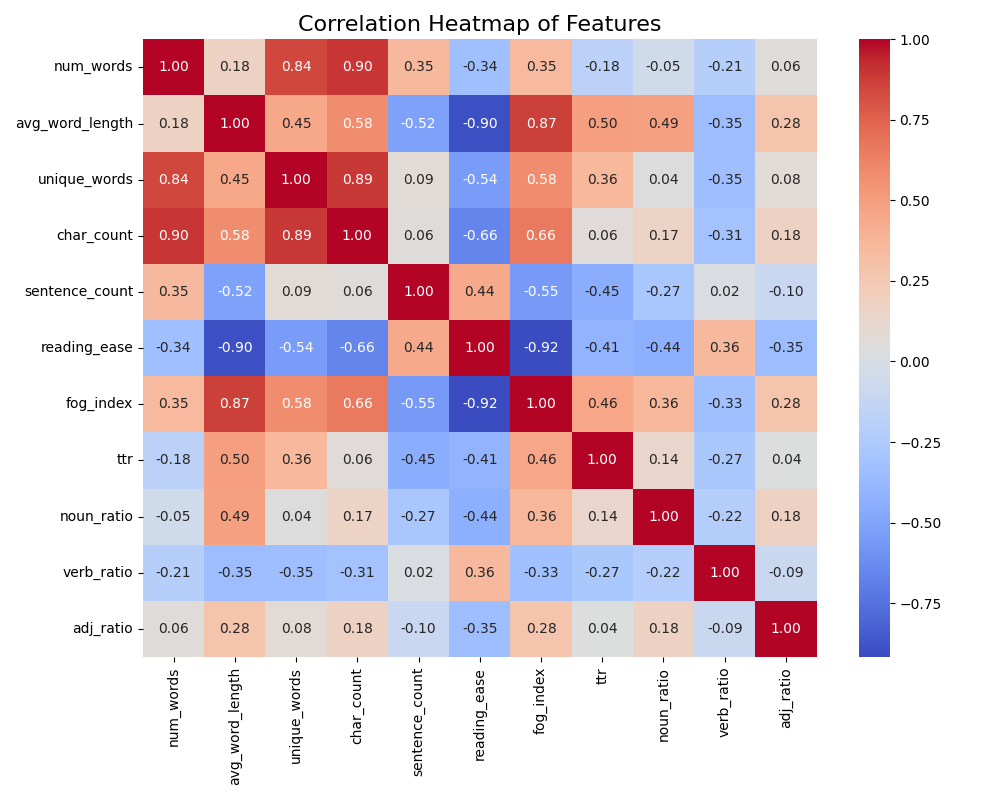}
    \caption{Mistral-Small-Instruct-2409 text features heatmap}
    \label{fig:human_heatmap}
\end{figure}

\begin{figure}[H]
    \centering
    \includegraphics[width=\linewidth]{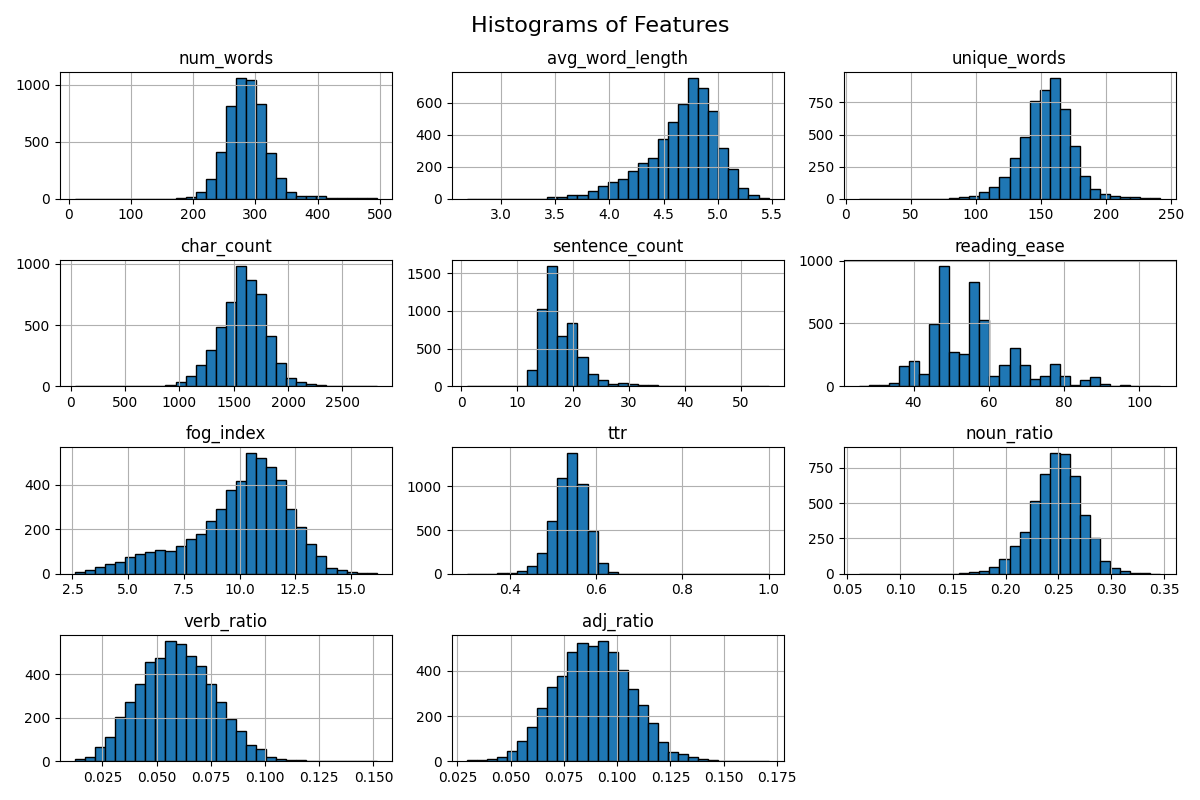}
    \caption{Mistral-Small-Instruct-2409 text features histgrams}
    \label{fig:human_hist}
\end{figure}

\begin{figure}[H]
    \centering
    \includegraphics[width=\linewidth]{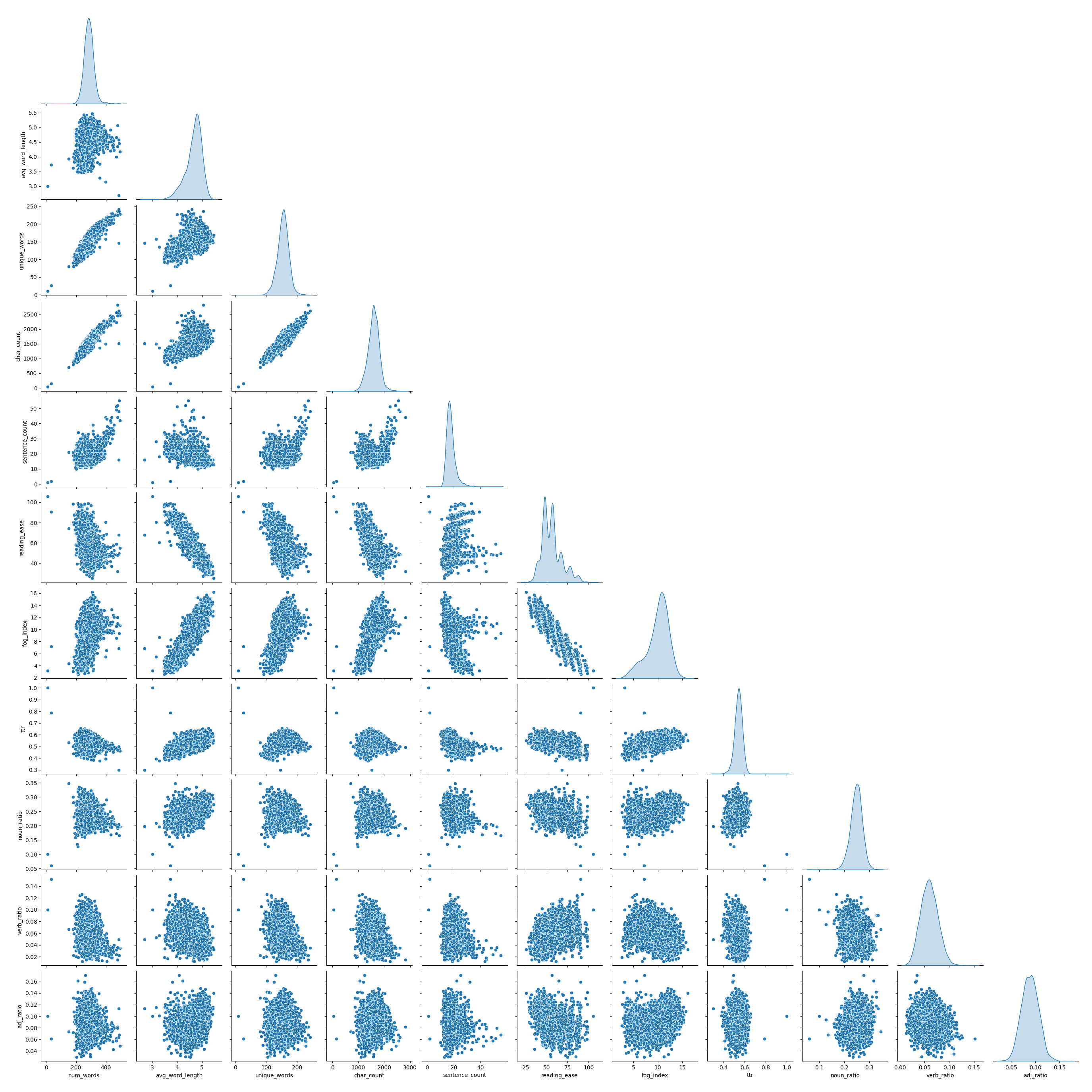}
    \caption{Mistral-Small-Instruct-2409 text features pairplot}
    \label{fig:human_pairplot}
\end{figure}
 
\section{Prompt}
\label{sec:prompt-content}
In this section, we present two types of prompts used in our study. The \textbf{System Prompt} is generated based on each learner’s attributes, including sex, academic background, English proficiency level, and country of origin, to construct a more targeted background persona. The \textbf{Task Prompt} provides the specific writing instructions, topic, length requirements, format, and spell-check reminders for the learners.

\subsection{System Prompt}
\label{sec:system_prompt}

We construct the \textbf{System Prompt} by concatenating four short statements reflecting the following dimensions in order: ``Sex,'' ``Acad.~Genre,'' ``CEFR,'' and ``Country.'' Each dimension--value pair maps to a predefined prompt string. For instance, if ``Sex'' = Male, then the prompt part is ``You are a male author.'' The final prompt strings for the other attributes are similarly selected from the respective lists. 

\vspace{1em}
\noindent
\begin{showcase}[title=Example]


|\classbg{Sex = Male}|
--> "You are a male author."

|\classbg{Acad. Genre = Science \& Technology}|
--> "Your academic background is in Science \& Technology."

|\classbg{CEFR = A2\_0}| 
--> "Your English proficiency level is CEFR A2."

|\classbg{Country = CHN}| 
--> "You are from China, an EFL environment."


\end{showcase}

When concatenated, these form the complete system prompt:

\begin{prompt}[title={Prompt \thetcbcounter: Male, Sci\_Tech, A2\_0, CHN}]
You are a male author. 
Your academic background is in Science \& Technology. 
Your English proficiency level is CEFR A2. 
You are from China, an EFL environment (English as a Foreign Language). 
\end{prompt}

\noindent
The JSON representation below details all possible dimension--value pairs and their corresponding prompt strings, which we combine to produce the final System Prompt.

\begin{showcase}[title=Prompt: Sex]


|\classbg{Sex = Male}|
You are a male author.

|\classbg{Sex = Female}|
You are a female author.

\end{showcase}

\begin{showcase}[title=Prompt: Acad. Genre]


|\classbg{Acad. Genre = Humanities}|
Your academic background is in the Humanities.

|\classbg{Acad. Genre = Social Sciences}|
Your academic background is in the Social Sciences.

|\classbg{Acad. Genre = Science \& Technology}|
Your academic background is in Science & Technology.

|\classbg{Acad. Genre = Life Science}|
Your academic background is in Life Science.

\end{showcase}

\begin{showcase}[title=Prompt: CEFR]


|\classbg{CEFR = A2\_0}|
Your English proficiency level is CEFR A2.

|\classbg{CEFR = B1\_1}|
Your English proficiency level is CEFR B1 (lower).

|\classbg{CEFR = B1\_2}|
Your English proficiency level is CEFR B1 (upper).

|\classbg{CEFR = B2\_0}|
Your English proficiency level is CEFR B2+.

|\classbg{CEFR = XX\_0}|
You are a native English speaker.

\end{showcase}

\begin{showcase}[title=Prompt: Country]


|\classbg{Country = CHN}|
You are from China, an EFL environment (English as a Foreign Language).

|\classbg{Country = THA}|
You are from Thailand, an EFL environment (English as a Foreign Language).

|\classbg{Country = JPN}|
You are from Japan, an EFL environment (English as a Foreign Language).

|\classbg{Country = KOR}|
You are from Korea, an EFL environment (English as a Foreign Language).

|\classbg{Country = IDN}|
You are from Indonesia, an EFL environment (English as a Foreign Language).

|\classbg{Country = PHL}|
You are from the Philippines, an ESL environment (English as a Second Language).

|\classbg{Country = PAK}|
You are from Pakistan, an ESL environment (English as a Second Language).

|\classbg{Country = SIN}|
You are from Singapore, an ESL environment (English as a Second Language).

|\classbg{Region = TWN}|
You are from Taiwan, an EFL environment (English as a Foreign Language).

|\classbg{Country = ENS\_USA}|
You are from the United States, a native English-speaking (NS) environment.

|\classbg{Country = HKG}|
You are from Hong Kong, an ESL environment (English as a Second Language).

|\classbg{Country = ENS\_GBR}|
You are from the United Kingdom, a native English-speaking (NS) environment.

|\classbg{Country = ENS\_CAN}|
You are from Canada, a native English-speaking (NS) environment.

|\classbg{Country = ENS\_AUS}|
You are from Australia, a native English-speaking (NS) environment.

|\classbg{Country = ENS\_NZL}|
You are from New Zealand, a native English-speaking (NS) environment.

\end{showcase}

\subsection{Prompt Illustration}
\label{sec:appendix_prompt}

For clarity, the template for constructing the \textit{System Prompt} is:
\begin{prompt}[title={Prompt Template}]
\{\texttt{sex\_prompt}\} 
\{\texttt{acad\_genre\_prompt}\} 
\{\texttt{cefr\_prompt}\} 
\{\texttt{country\_prompt}\}
\end{prompt}

Each placeholder (e.g., \{\texttt{sex\_prompt}\}) is replaced by the corresponding string from the JSON snippets in Section~\ref{sec:system_prompt}. Below is a concrete illustration using one combination of attributes:

\begin{prompt}[title={Sample Prompt: F, Humanities, B1\_1, SIN}]
You are a female author. 
Your academic background is in the Humanities. 
Your English proficiency level is CEFR B1 (lower). 
You are from Singapore, an ESL environment (English as a Second Language). 
\end{prompt}

Thus, the System Prompt succinctly captures user-specific information to guide the subsequent text generation process.

\subsection{Task Prompt}
\label{appendix:task_prompt}
We use two topics for the writing tasks, each with the following requirements. Learners must use a word processor, keep the word count between 200 and 300 words, and run spell check before finalizing.

\begin{prompt}[title={Prompt \thetcbcounter: PTJ task prompt}]
Do you agree or disagree with the following statements? Use reasons and specific details to support your opinion. \\
(Topic) It is important for college students to have a part-time job. \\ \\
Instructions \\
1. Clarify your opinions and show the reasons and some examples. \\
2. The length of your single essay should be from 200 to 300 WORDS (not letters). Too short or too long essays cannot be accepted. \\
3. You must run spell check before completing your writing.\\
\end{prompt}

\begin{prompt}[title={Prompt \thetcbcounter: SMK task prompt}]
Do you agree or disagree with the following statements? Use reasons and specific details to support your opinion. \\
(Topic) Smoking should be completely banned at all the restaurants in the country.\\ \\
Instructions \\
1. Clarify your opinions and show the reasons and some examples.\\
2. The length of your single essay should be from 200 to 300 WORDS (not letters). Too short or too long essays cannot be accepted. \\
3. You must run spell check before completing your writing. \\
\end{prompt}     
\clearpage
\section{Additional Experiments}
In this section, we treat AI text detection as a binary classification problem and conduct additional experiments using classical machine learning methods to explore potential biases.

\subsection{Training Data}
We adopt the HC3 dataset \citep{guo2023closechatgpthumanexperts} as our training corpus. HC3 contains 24.3k prompts, each accompanied by both human and ChatGPT responses, spanning diverse domains (e.g., Reddit Q\&A, medical, finance, law). Widely used in existing research, HC3 provides a representative overview of differences between LLM-generated and human-authored text.

\subsection{Feature Engineering}
We extract a range of textual features to train the classical models. Table~\ref{tab:features_explanations} summarizes the categories and specific features, including basic lexical counts, syntactic information, and readability metrics.

\begin{table*}[h]
\small
\centering
\begin{tabular}{c|c|c}
\toprule
\textbf{Category} & \textbf{Feature} & \textbf{Explanation} \\
\midrule
\multirow{4}{*}{Basic}
 & Word Count         & Total words in a text \\
 & Unique Words       & Number of distinct words \\
 & Character Count    & Total characters \\
 & Sentence Count     & Number of sentences \\
\midrule
\multirow{2}{*}{Lexical}
 & Avg Word Length    & (Characters minus special characters) / total words \\
 & Type-Token Ratio   & Distinct words / total words \\
\midrule
Syntactic / POS
 & N/V/Adj Ratio      & Proportion of nouns, verbs, adjectives \\
\midrule
\multirow{2}{*}{Readability}
 & Flesch Reading Ease & Formula-based readability score \\
 & Gunning Fog Index   & Index using complex words and sentence length \\
\bottomrule
\end{tabular}
\caption{Features and their explanations.}
\label{tab:features_explanations}
\end{table*}

\subsection{Machine Learning Detectors}
To complement neural-based approaches, we train a suite of classical machine learning models (Table~\ref{tab:classical_ml_models}) on the extracted features. By evaluating their performance, we aim to assess whether biases might arise from specific model families or feature sets.

\begin{table*}[htbp]
\centering
\small
\resizebox{1.7\columnwidth}{!}{%
\begin{tabular}{lp{7cm}}
\toprule
\textbf{Category} & \textbf{Models} \\
\midrule
\textbf{Linear Models} 
& Logistic Regression, Ridge Classifier, Perceptron, Passive Aggressive Classifier \\
\textbf{Support Vector Machines (SVM)}
& SVC (Support Vector Classifier), NuSVC \\
\textbf{Naive Bayes} 
& Multinomial NB, Bernoulli NB, Gaussian NB, Complement NB \\
\textbf{Neighbor-Based Methods}
& KNN (K-Nearest Neighbors), Nearest Centroid \\
\textbf{Discriminant Analysis}
& LDA (Linear Discriminant Analysis), QDA (Quadratic Discriminant Analysis) \\
\textbf{Tree-Based Models}
& Decision Tree, Random Forest, Extra Tree \\
\textbf{Boosting}
& Gradient Boosting, XGBoost, AdaBoost \\
\textbf{Ensemble Learning}
& Bagging, Voting Classifier, Stacking Classifier \\
\textbf{Neural Networks}
& MLP (Multi-Layer Perceptron) \\
\textbf{Stochastic Methods}
& Stochastic Gradient Descent (SGD) Classifier \\
\bottomrule
\end{tabular}%
}
\caption{Summary of machine learning models used.}
\label{tab:classical_ml_models}
\end{table*}

\subsection{Evaluation Metrics}

We treat AI text detection as a binary classification task and use \textbf{Accuracy} as the primary metric:

\begin{equation}
\mathrm{Accuracy}=\frac{\mathrm{TP}+\mathrm{TN}}{\mathrm{TP}+\mathrm{TN}+\mathrm{FP}+\mathrm{FN}}
\end{equation}

\noindent where TP is true positive (\emph{AI text correctly identified}), TN is true negative (\emph{human text correctly identified}), FP is false positive (\emph{human text misidentified as AI}), and FN is false negative (\emph{AI text misidentified as human}).

We compare results under two distinct scenarios:
\paragraph{In-domain Testing (HC3-based).} The testing data share the same distribution as the training set. 
\paragraph{Out-of-domain Testing (ICNALE + LLM).} The testing data come from a different distribution (ICNALE plus newly generated LLM texts).

This setup clarifies how well the detectors \emph{generalize} to previously unseen text distributions.

\subsection{Results}

\begin{figure}[!t]
    \centering
    \includegraphics[width=\linewidth]{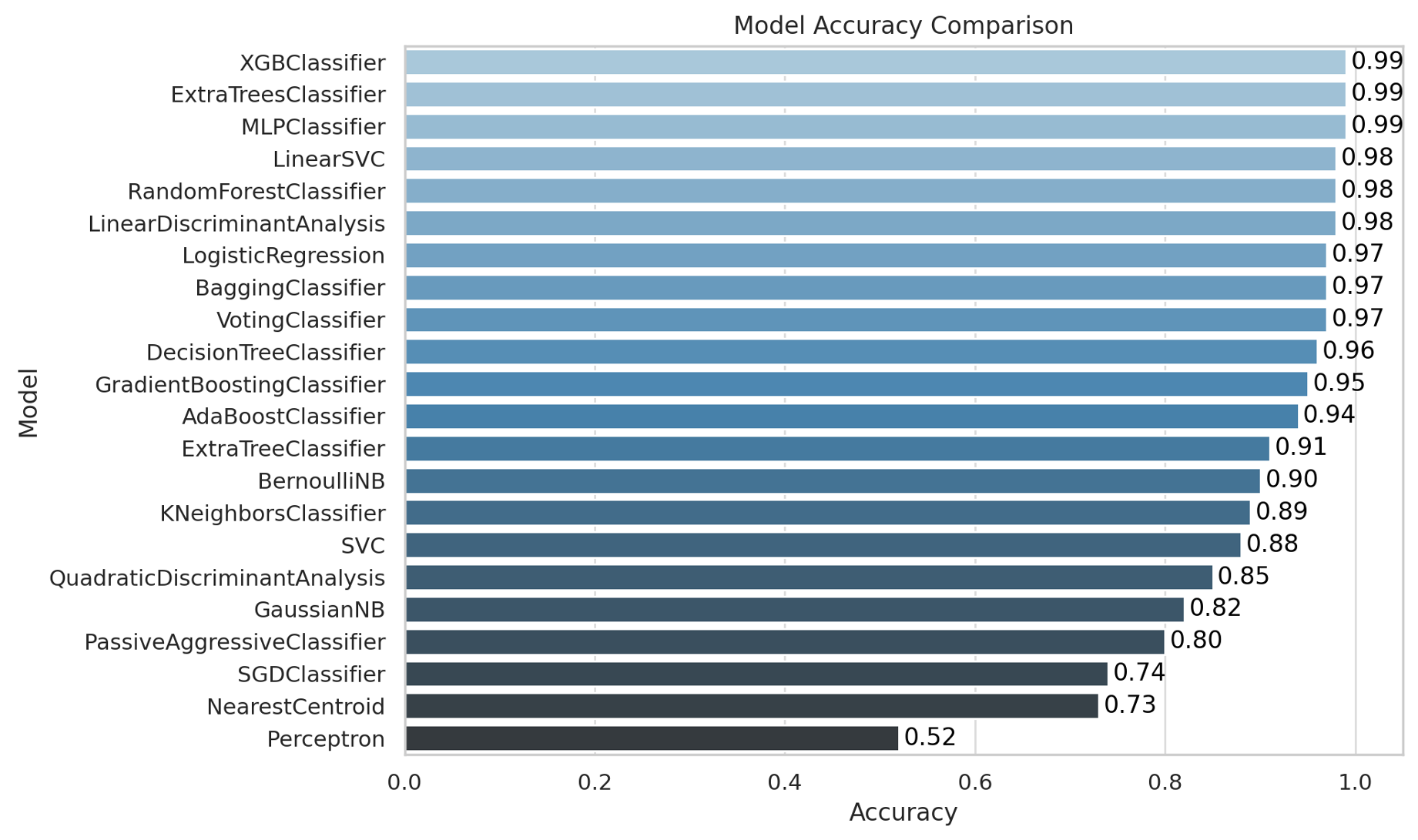}
    \caption{In-domain test results (HC3). Accuracy on the y-axis.}
    \label{fig:in_domain}
\end{figure}

\paragraph{In-domain Results.} Figure~\ref{fig:in_domain} shows the in-domain performance on HC3. Most detectors achieve accuracy scores above 0.90, with some models approaching 0.99. This indicates that, when the test data distribution is similar to training, current deep learning detectors can effectively distinguish AI from human texts. 

\begin{figure}[!t]
    \centering
    \includegraphics[width=\linewidth]{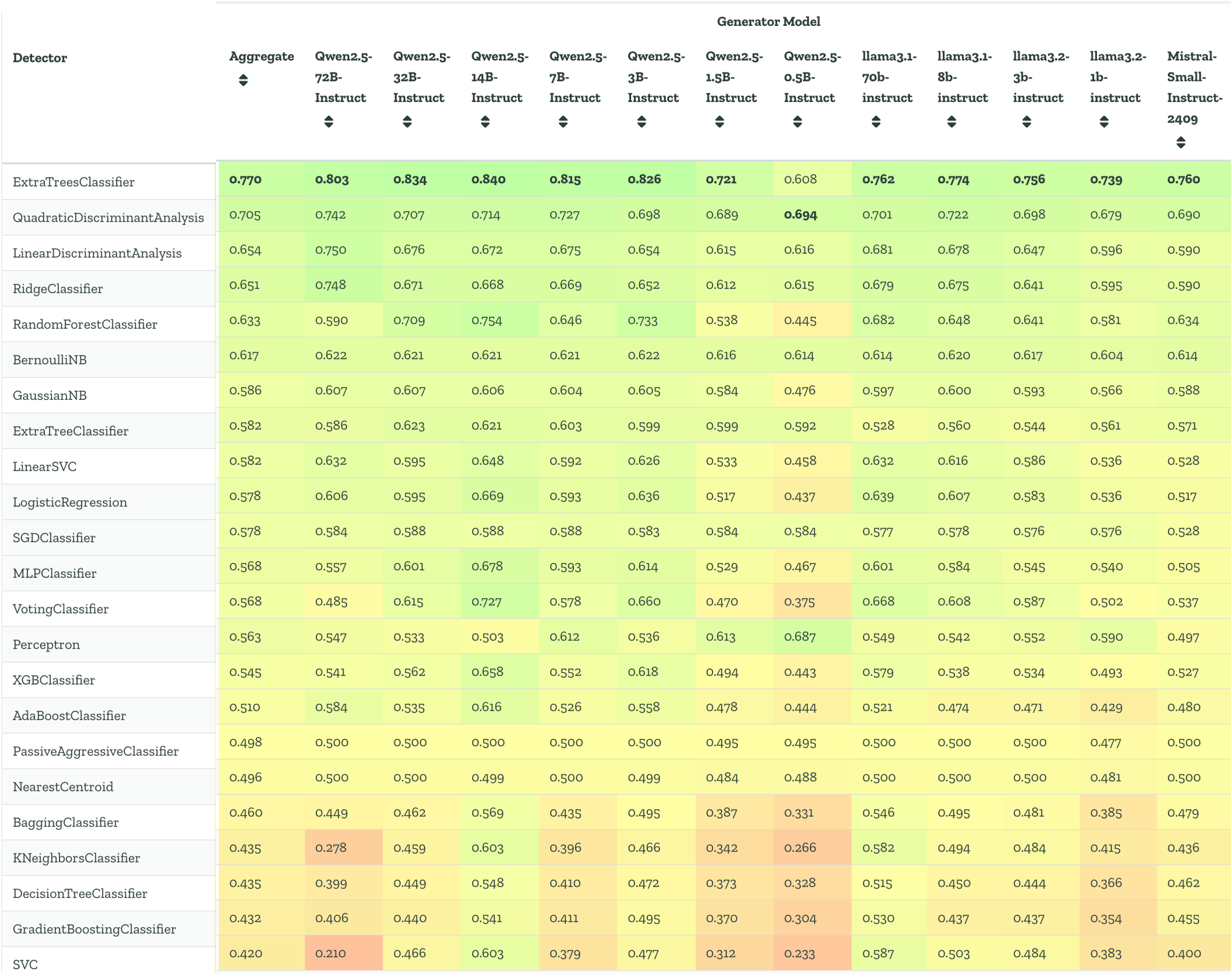}
    \caption{Out-of-domain test results (ICNALE + LLM). Accuracy on the y-axis.}
    \label{fig:out_domain}
\end{figure}

\paragraph{Out-of-domain Results.} 
Figure~\ref{fig:out_domain} presents the out-of-domain results using our ICNALE + LLM dataset. Performance drops significantly compared to in-domain, with accuracy scores generally in the 0.50--0.65 range, and some models falling as low as 0.40. This underlines the limited capacity of detectors to transfer knowledge when encountering distributions different from their training data. Such a gap highlights the need for more diverse training corpora and advanced adaptation or domain-transfer techniques.

In sum, while machine learning detectors show encouraging performance on in-distribution data, they often struggle to maintain high accuracy once faced with new domains or author attributes. Real-world AI text detection frequently involves such diverse and shifting distributions, underscoring the importance of robust generalization strategies.

\subsection{Bias Analysis}
\paragraph{Hypothesis Test Setting}
In order to investigate whether different features exhibit significant differences in the performance of various machine learning detectors, this section adopts appropriate statistical tests based on the number of possible values of each feature. Specifically, for features that have only two values (e.g., gender), we use the independent samples t-test; for features that contain more than two values (e.g., CEFR, academic writing style, language environment), we conduct ANOVA to evaluate whether there are significant differences across different feature levels.

\paragraph{Hypothesis Test Results}
Figure~\ref{fig:hypothesis_test_results} shows the t-test and ANOVA results for the four main features (gender, CEFR level, academic genre, and language environment). Each row corresponds to a different model, and each column corresponds to one of the four features. The results are indicated by ``1'' for significant (p < 0.05) and ``0'' for non-significant (p \(\ge\) 0.05). From these results, we can draw the following conclusions:

\begin{figure}[!t]
    \centering
    \includegraphics[width=\linewidth]{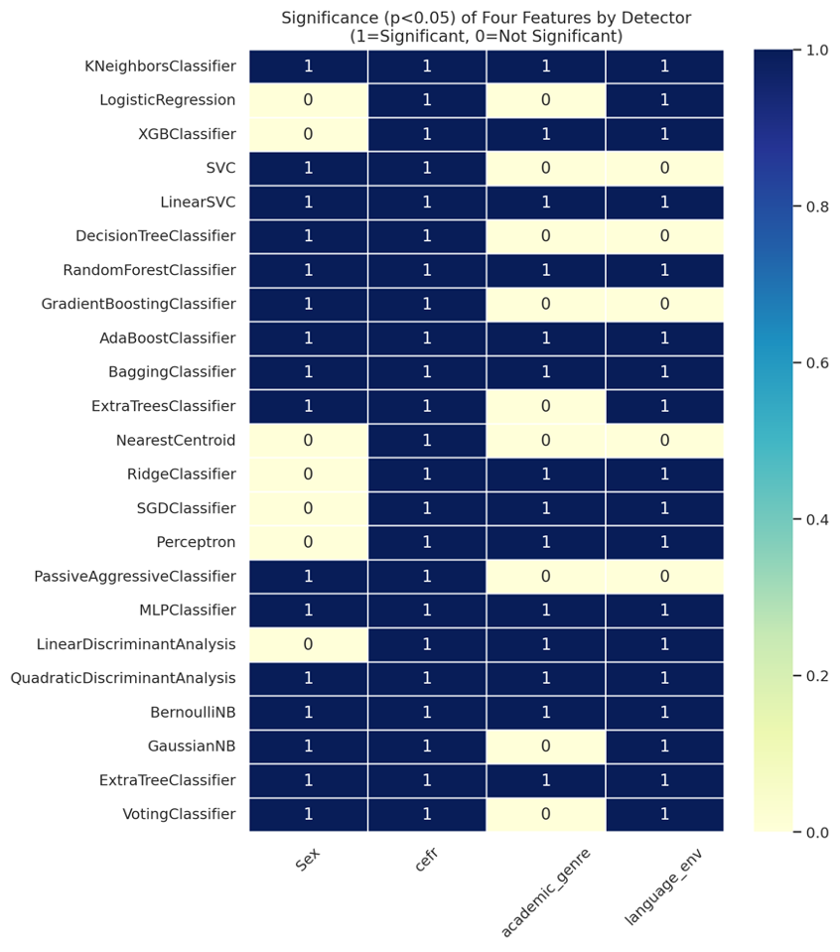}
    \caption{Hypothesis test results (t-test and ANOVA)}
    \label{fig:hypothesis_test_results}
\end{figure}

\paragraph{Gender (Sex, T-test)} 
For the binary feature of gender (Female vs.\ Male), most detectors (e.g., KNeighborsClassifier, SVC, RandomForestClassifier, etc.) exhibit significant differences (p < 0.05), indicating that gender-based distinctions have statistical significance in these detectors’ predictive performance. However, a few detectors such as LogisticRegression, XGBClassifier, and NearestCentroid do not show significant differences, suggesting that for these detectors, gender has a weaker or non-significant effect on the detection results.

\paragraph{CEFR Level (cefr, ANOVA)} 
All detectors display significant differences in the ANOVA results for the multi-class feature of CEFR level, indicating that the graded language proficiency (ranging from beginner to advanced) exerts distinct and statistically significant effects on the detectors’ predictions. This further demonstrates that CEFR levels possess strong predictive discriminative power and significantly influence the outcomes of various detectors.

\paragraph{Academic Genre (academic\_genre, ANOVA)} 
For academic genre, some detectors (e.g., XGBClassifier, RandomForestClassifier, AdaBoostClassifier, etc.) exhibit significant differences in the ANOVA, whereas others (e.g., LogisticRegression, SVC, DecisionTreeClassifier, etc.) do not. This finding suggests that different detectors vary in their sensitivity to academic writing styles; some are better at capturing and leveraging stylistic differences, while others do not reflect clear statistical differences in their predictions.

\paragraph{Language Environment (language\_env, ANOVA)} 
With respect to language environment, the ANOVA results show that the vast majority of detectors exhibit significant differences, indicating that different language environments (e.g., native vs.\ second language settings) have a substantial impact on the predictive outcomes of these detectors. However, certain detectors (e.g., SVC, DecisionTreeClassifier, GradientBoostingClassifier, etc.) do not reach the threshold of statistical significance on this feature, suggesting that their ability to differentiate language environments is insufficient to yield statistically significant results.

Taken together, the analysis reveals that detectors vary in their sensitivity and discriminative power with regard to gender, CEFR level, academic writing style, and language environment. Most detectors demonstrate high discriminative power for CEFR level and language environment, while sensitivity to gender and academic writing style depends on the specific detector. These findings provide a statistical perspective on the influence of various features in AI-based text detection and offer valuable insights for optimizing both detectors and feature selection in future research.

\begin{table*}[htbp]
\centering
\small
\begin{tabular}{lccc}
\toprule
\textbf{Detector} & \textbf{t-value} & \textbf{p-value} & \textbf{Significance (p<0.05)} \\
\midrule
KNeighborsClassifier            & -3.8818 & 1.0812e-04 & Significant \\
LogisticRegression              & -0.2408 & 8.0976e-01 & Not Significant \\
XGBClassifier                   & -1.4438 & 1.4900e-01 & Not Significant \\
SVC                             & -3.9214 & 9.1954e-05 & Significant \\
LinearSVC                       & -2.8893 & 3.9144e-03 & Significant \\
DecisionTreeClassifier          & -3.1055 & 1.9351e-03 & Significant \\
RandomForestClassifier          & -4.8084 & 1.6770e-06 & Significant \\
GradientBoostingClassifier      & -3.8380 & 1.2898e-04 & Significant \\
AdaBoostClassifier              & -4.2847 & 1.9443e-05 & Significant \\
BaggingClassifier               & -3.5719 & 3.6564e-04 & Significant \\
ExtraTreesClassifier            & -5.5451 & 3.4535e-08 & Significant \\
NearestCentroid                 & -1.3941 & 1.6347e-01 & Not Significant \\
RidgeClassifier                 & -0.7179 & 4.7293e-01 & Not Significant \\
SGDClassifier                   & -0.5139 & 6.0741e-01 & Not Significant \\
Perceptron                      & -1.3458 & 1.7856e-01 & Not Significant \\
PassiveAggressiveClassifier     & -3.2496 & 1.2048e-03 & Significant \\
MLPClassifier                   & -3.5214 & 4.4255e-04 & Significant \\
LinearDiscriminantAnalysis      & -1.0359 & 3.0043e-01 & Not Significant \\
QuadraticDiscriminantAnalysis   & -5.1645 & 2.7226e-07 & Significant \\
BernoulliNB                     & -4.2733 & 2.0588e-05 & Significant \\
GaussianNB                      & -3.3843 & 7.3276e-04 & Significant \\
ExtraTreeClassifier             & -5.4473 & 5.9765e-08 & Significant \\
VotingClassifier                & -3.4404 & 5.9649e-04 & Significant \\
\bottomrule
\end{tabular}
\caption{T-test results for Feature 1: Sex}
\end{table*}

\begin{table*}[htbp]
\centering
\small
\begin{tabular}{lccc}
\toprule
\textbf{Detector} & \textbf{F-value} & \textbf{p-value} & \textbf{Significance (p<0.05)} \\
\midrule
KNeighborsClassifier           & 9.7256   & 8.9973e-08   & Significant \\
LogisticRegression             & 10.2101  & 3.6569e-08   & Significant \\
XGBClassifier                  & 14.5891  & 1.0422e-11   & Significant \\
SVC                            & 3.3956   & 8.9432e-03   & Significant \\
LinearSVC                      & 11.2302  & 5.4780e-09   & Significant \\
DecisionTreeClassifier         & 9.4865   & 1.4025e-07   & Significant \\
RandomForestClassifier         & 47.3126  & 1.2241e-37   & Significant \\
GradientBoostingClassifier     & 9.9970   & 5.4343e-08   & Significant \\
AdaBoostClassifier             & 15.2355  & 3.1203e-12   & Significant \\
BaggingClassifier              & 9.3998   & 1.6475e-07   & Significant \\
ExtraTreesClassifier           & 51.9283  & 3.5969e-41   & Significant \\
NearestCentroid                & 3.0469   & 1.6277e-02   & Significant \\
RidgeClassifier                & 37.2302  & 8.2609e-30   & Significant \\
SGDClassifier                  & 66.5123  & 4.0857e-52   & Significant \\
Perceptron                     & 45.3916  & 3.6934e-36   & Significant \\
PassiveAggressiveClassifier    & 3.0908   & 1.5361e-02   & Significant \\
MLPClassifier                  & 20.9005  & 8.1949e-17   & Significant \\
LinearDiscriminantAnalysis     & 36.4892  & 3.1519e-29   & Significant \\
QuadraticDiscriminantAnalysis  & 39.3666  & 1.7583e-31   & Significant \\
BernoulliNB                    & 98.4916  & 5.1139e-75   & Significant \\
GaussianNB                     & 45.5083  & 3.0021e-36   & Significant \\
ExtraTreeClassifier            & 8.2371   & 1.4190e-06   & Significant \\
VotingClassifier               & 26.9124  & 1.2214e-21   & Significant \\
\bottomrule
\end{tabular}
\caption{ANOVA results for Feature 2: CEFR}
\end{table*}

\begin{table*}[htbp]
\centering
\small
\begin{tabular}{lccc}
\toprule
\textbf{Detector} & \textbf{F-value} & \textbf{p-value} & \textbf{Significance (p<0.05)} \\
\midrule
LogisticRegression             & 0.9484  & 4.1638e-01   & Not Significant \\
XGBClassifier                  & 3.8680  & 9.0291e-03   & Significant \\
SVC                            & 2.3311  & 7.2526e-02   & Not Significant \\
LinearSVC                      & 8.5848  & 1.1803e-05   & Significant \\
DecisionTreeClassifier         & 2.1278  & 9.4821e-02   & Not Significant \\
RandomForestClassifier         & 5.2634  & 1.2955e-03   & Significant \\
GradientBoostingClassifier     & 2.3275  & 7.2877e-02   & Not Significant \\
AdaBoostClassifier             & 3.7601  & 1.0475e-02   & Significant \\
BaggingClassifier              & 7.3502  & 6.8191e-05   & Significant \\
ExtraTreesClassifier           & 1.6179  & 1.8329e-01   & Not Significant \\
KNeighborsClassifier           & 8.5072  & 1.3181e-05   & Significant \\
NearestCentroid                & 0.2623  & 8.5258e-01   & Not Significant \\
RidgeClassifier                & 9.0998  & 5.6698e-06   & Significant \\
SGDClassifier                  & 5.1594  & 1.4985e-03   & Significant \\
Perceptron                     & 7.5843  & 4.8926e-05   & Significant \\
PassiveAggressiveClassifier    & 1.1490  & 3.2840e-01   & Not Significant \\
MLPClassifier                  & 6.3252  & 2.9084e-04   & Significant \\
LinearDiscriminantAnalysis     & 7.2302  & 8.0842e-05   & Significant \\
QuadraticDiscriminantAnalysis  & 44.5179 & 1.2424e-27   & Significant \\
BernoulliNB                    & 22.5459 & 2.6820e-14   & Significant \\
GaussianNB                     & 1.3856  & 2.4545e-01   & Not Significant \\
ExtraTreeClassifier            & 7.4089  & 6.2747e-05   & Significant \\
VotingClassifier               & 2.4313  & 6.3496e-02   & Not Significant \\
\bottomrule
\end{tabular}
\caption{ANOVA results for Feature 3: Academic Genre}
\end{table*}

\begin{table*}[htbp]
\centering
\small
\begin{tabular}{lccc}
\toprule
\textbf{Detector} & \textbf{F-value} & \textbf{p-value} & \textbf{Significance (p<0.05)} \\
\midrule
LogisticRegression             & 15.4867  & 2.1791e-07  & Significant \\
XGBClassifier                  & 6.2824   & 1.9152e-03  & Significant \\
SVC                            & 1.7646   & 1.7158e-01  & Not Significant \\
LinearSVC                      & 12.8361  & 2.9476e-06  & Significant \\
DecisionTreeClassifier         & 0.7064   & 4.9355e-01  & Not Significant \\
RandomForestClassifier         & 98.8694  & 3.2184e-41  & Significant \\
GradientBoostingClassifier     & 0.5297   & 5.8886e-01  & Not Significant \\
AdaBoostClassifier             & 12.1365  & 5.8702e-06  & Significant \\
BaggingClassifier              & 14.2985  & 6.9971e-07  & Significant \\
ExtraTreesClassifier           & 124.3115 & 6.4051e-51  & Significant \\
KNeighborsClassifier           & 3.0683   & 4.6771e-02  & Significant \\
NearestCentroid                & 1.5409   & 2.1451e-01  & Not Significant \\
RidgeClassifier                & 44.5998  & 1.4114e-19  & Significant \\
SGDClassifier                  & 159.0237 & 9.8119e-64  & Significant \\
Perceptron                     & 47.5704  & 8.5035e-21  & Significant \\
PassiveAggressiveClassifier    & 0.2646   & 7.6755e-01  & Not Significant \\
MLPClassifier                  & 26.3431  & 5.5349e-12  & Significant \\
LinearDiscriminantAnalysis     & 43.3016  & 4.8325e-19  & Significant \\
QuadraticDiscriminantAnalysis  & 59.9251  & 7.9565e-26  & Significant \\
BernoulliNB                    & 334.0978 & 4.9281e-122 & Significant \\
GaussianNB                     & 42.1002  & 1.5120e-18  & Significant \\
ExtraTreeClassifier            & 15.4280  & 2.3082e-07  & Significant \\
VotingClassifier               & 64.4107  & 1.2380e-27  & Significant \\
\bottomrule
\end{tabular}
\caption{ANOVA results for Feature 4: Language Environment}
\end{table*} 
\section{Sensitivity Analyses of WLS}

While the multi-factor weighted least squares (WLS) model provides a systematic approach to control for multiple confounders, datasets often exhibit \emph{imbalanced subgroup distributions} or heterogeneity that can affect statistical inferences.  To ensure the robustness of our parameter estimates, we perform a bootstrap-based sensitivity analysis.

\subsection{Bootstrap-Based Parameter Estimation}

Parameter estimates can be sensitive to random fluctuations in the data.  To assess this sensitivity, we use bootstrapping. We create many "new" datasets by resampling with replacement from the original dataset (keeping the same overall size).  We fit the WLS model on each bootstrap sample and aggregate the resulting estimates. This approach provides a distribution for each parameter. We report the mean and standard deviation of these bootstrap estimates, along with the 95\% confidence interval (CI) based on the 2.5th and 97.5th percentiles of the bootstrap distribution. We then check whether the original parameter estimate falls within this bootstrap CI ("Coverage"). If the original estimate lies within the CI, it provides evidence that the estimate is stable to sampling variability, and thus robust to the specific composition of the sample.

\subsection{Summary of Sensitivity Findings}

The bootstrap results, presented in Tables \ref{tab:bootstrap_comparison_binoculars} through \ref{tab:radar_bootstrap}, show that for all detectors and all parameters, the original coefficient estimates lie within the 95\% confidence intervals derived from the bootstrap resampling.  This indicates strong stability of the parameter estimates. The relatively narrow confidence intervals and consistent "WITHIN CI" coverage across all parameters and detectors provide substantial evidence that our main WLS findings are robust to sampling variability. This strengthens our confidence in the reported effects of CEFR level, sex, academic genre, and language environment on detector accuracy.

\begin{table*}[htbp]
  \centering
  \resizebox{\textwidth}{!}{%
    \begin{tabular}{lcccccc}
      \hline
      \textbf{Parameter} & \textbf{Original Value} & \textbf{Bootstrap Mean} & \textbf{Bootstrap Std} & \textbf{CI Lower} & \textbf{CI Upper} & \textbf{Coverage} \\
      \hline
      Intercept & 0.9482 & 0.9480 & 0.0079 & 0.9318 & 0.9625 & WITHIN CI \\
      C(cefr)[T.B1\_1] & -0.0039 & -0.0040 & 0.0085 & -0.0204 & 0.0132 & WITHIN CI \\
      C(cefr)[T.B1\_2] & -0.0007 & -0.0006 & 0.0075 & -0.0149 & 0.0150 & WITHIN CI \\
      C(cefr)[T.B2\_0] & 0.0025 & 0.0025 & 0.0078 & -0.0125 & 0.0172 & WITHIN CI \\
      C(cefr)[T.XX\_0] & -0.0501 & -0.0499 & 0.0049 & -0.0592 & -0.0400 & WITHIN CI \\
      C(Sex)[T.M] & 0.0010 & 0.0011 & 0.0054 & -0.0096 & 0.0114 & WITHIN CI \\
      C(academic\_genre)[T.Life Sciences] & -0.0075 & -0.0073 & 0.0082 & -0.0224 & 0.0086 & WITHIN CI \\
      C(academic\_genre)[T.Sciences \& Technology] & 0.0016 & 0.0018 & 0.0072 & -0.0126 & 0.0156 & WITHIN CI \\
      C(academic\_genre)[T.Social Sciences] & 0.0016 & 0.0018 & 0.0068 & -0.0110 & 0.0149 & WITHIN CI \\
      C(language\_env)[T.ESL] & -0.0144 & -0.0143 & 0.0056 & -0.0247 & -0.0038 & WITHIN CI \\
      C(language\_env)[T.NS] & -0.0501 & -0.0499 & 0.0049 & -0.0592 & -0.0400 & WITHIN CI \\
      \hline
    \end{tabular}}
   \caption{Bootstrap Sensitivity Analysis for detector: \texttt{binoculars}}
  \label{tab:bootstrap_comparison_binoculars}
\end{table*}

\begin{table*}[ht]
\centering
\resizebox{\textwidth}{!}{%
\begin{tabular}{lcccccc}
\toprule
\textbf{Parameter} & \textbf{Original Value} & \textbf{Bootstrap Mean} & \textbf{Bootstrap Std} & \textbf{CI Lower} & \textbf{CI Upper} & \textbf{Coverage} \\
\midrule
Intercept                                   & 0.7480 & 0.7480 & 0.0157 & 0.7179 & 0.7790 & WITHIN CI \\
C(Sex)[T.M]                                & 0.0035 & 0.0032 & 0.0103 & --0.0176 & 0.0226 & WITHIN CI \\
C(cefr)[T.B1\_1]                           & --0.0073 & --0.0074 & 0.0180 & --0.0403 & 0.0294 & WITHIN CI \\
C(cefr)[T.B1\_2]                           & --0.0103 & --0.0102 & 0.0158 & --0.0413 & 0.0214 & WITHIN CI \\
C(cefr)[T.B2\_0]                           & --0.0435 & --0.0432 & 0.0155 & --0.0728 & --0.0114 & WITHIN CI \\
C(cefr)[T.XX\_0]                           & --0.0071 & --0.0071 & 0.0084 & --0.0245 & 0.0100 & WITHIN CI \\
C(academic\_genre)[T.Life Sciences]         & 0.0286 & 0.0287 & 0.0143 & 0.0007 & 0.0557 & WITHIN CI \\
C(academic\_genre)[T.Sciences \& Technology]  & 0.0084 & 0.0088 & 0.0137 & --0.0179 & 0.0350 & WITHIN CI \\
C(academic\_genre)[T.Social Sciences]        & 0.0083 & 0.0085 & 0.0120 & --0.0167 & 0.0312 & WITHIN CI \\
C(language\_env)[T.ESL]                      & --0.0014 & --0.0012 & 0.0094 & --0.0201 & 0.0185 & WITHIN CI \\
C(language\_env)[T.NS]                       & --0.0071 & --0.0071 & 0.0084 & --0.0245 & 0.0100 & WITHIN CI \\
\bottomrule
\end{tabular}
}
\caption{Bootstrap Sensitivity Analysis for detector: \texttt{chatgpt-roberta}}
\label{tab:bootstrap_chatgpt-roberta}
\end{table*}

\begin{table*}[ht]
\centering
\resizebox{\textwidth}{!}{%
\begin{tabular}{lcccccc}
\toprule
\textbf{Parameter} & \textbf{Original Value} & \textbf{Bootstrap Mean} & \textbf{Bootstrap Std} & \textbf{CI Lower} & \textbf{CI Upper} & \textbf{Coverage} \\
\midrule
Intercept                                   & 0.7944    & 0.7948   & 0.0098  & 0.7754  & 0.8133 & WITHIN CI \\
C(academic\_genre)[T.Life Sciences]         & --0.0397  & --0.0398 & 0.0108  & --0.0611 & --0.0187 & WITHIN CI \\
C(academic\_genre)[T.Sciences \& Technology]  & --0.0185  & --0.0189 & 0.0093  & --0.0370 & --0.0010 & WITHIN CI \\
C(academic\_genre)[T.Social Sciences]        & --0.0060  & --0.0063 & 0.0087  & --0.0233 & 0.0101 & WITHIN CI \\
C(cefr)[T.B1\_1]                             & 0.0127    & 0.0129   & 0.0104  & --0.0074 & 0.0347 & WITHIN CI \\
C(cefr)[T.B1\_2]                             & 0.0284    & 0.0282   & 0.0093  & 0.0103  & 0.0468 & WITHIN CI \\
C(cefr)[T.B2\_0]                             & 0.0345    & 0.0339   & 0.0106  & 0.0132  & 0.0554 & WITHIN CI \\
C(cefr)[T.XX\_0]                             & --0.0223  & --0.0222 & 0.0059  & --0.0338 & --0.0104 & WITHIN CI \\
C(Sex)[T.M]                                 & --0.0068  & --0.0069 & 0.0075  & --0.0210 & 0.0079 & WITHIN CI \\
C(language\_env)[T.ESL]                       & --0.0077  & --0.0075 & 0.0071  & --0.0208 & 0.0065 & WITHIN CI \\
C(language\_env)[T.NS]                        & --0.0223  & --0.0222 & 0.0059  & --0.0338 & --0.0104 & WITHIN CI \\
\bottomrule
\end{tabular}
}
\caption{Bootstrap Sensitivity Analysis for detector: \texttt{detectgpt}}
\label{tab:bootstrap_detectgpt}
\end{table*}

\begin{table*}[ht]
\centering
\resizebox{\textwidth}{!}{%
\begin{tabular}{lcccccc}
\toprule
\textbf{Parameter} & \textbf{Original Value} & \textbf{Bootstrap Mean} & \textbf{Bootstrap Std} & \textbf{CI Lower} & \textbf{CI Upper} & \textbf{Coverage} \\
\midrule
Intercept                                   & 0.8963    & 0.8963   & 0.0084  & 0.8796  & 0.9124 & WITHIN CI \\
C(cefr)[T.B1\_1]                            & -0.0076   & -0.0071  & 0.0093  & -0.0241 & 0.0105 & WITHIN CI \\
C(cefr)[T.B1\_2]                            & 0.0060    & 0.0059   & 0.0082  & -0.0107 & 0.0222 & WITHIN CI \\
C(cefr)[T.B2\_0]                            & 0.0180    & 0.0181   & 0.0083  & 0.0023  & 0.0346 & WITHIN CI \\
C(cefr)[T.XX\_0]                            & -0.0214   & -0.0214  & 0.0049  & -0.0309 & -0.0119 & WITHIN CI \\
C(Sex)[T.M]                                & -0.0003   & -0.0005  & 0.0062  & -0.0130 & 0.0113 & WITHIN CI \\
C(academic\_genre)[T.Life Sciences]          & 0.0127    & 0.0123   & 0.0077  & -0.0025 & 0.0281 & WITHIN CI \\
C(academic\_genre)[T.Sciences \& Technology] & -0.0012   & -0.0013  & 0.0080  & -0.0174 & 0.0144 & WITHIN CI \\
C(academic\_genre)[T.Social Sciences]        & 0.0070    & 0.0073   & 0.0071  & -0.0068 & 0.0216 & WITHIN CI \\
C(language\_env)[T.ESL]                      & -0.0021   & -0.0021  & 0.0060  & -0.0141 & 0.0102 & WITHIN CI \\
C(language\_env)[T.NS]                       & -0.0214   & -0.0214  & 0.0049  & -0.0309 & -0.0119 & WITHIN CI \\
\bottomrule
\end{tabular}
}
\caption{Bootstrap Sensitivity Analysis for detector: \texttt{fastdetectgpt}}
\label{tab:bootstrap_fastdetectgpt}
\end{table*}

\begin{table*}[ht]
\centering
\resizebox{\textwidth}{!}{%
\begin{tabular}{lcccccc}
\toprule
\textbf{Parameter} & \textbf{Original Value} & \textbf{Bootstrap Mean} & \textbf{Bootstrap Std} & \textbf{CI Lower} & \textbf{CI Upper} & \textbf{Coverage} \\
\midrule
Intercept                                   & 0.4873    & 0.4873   & 0.0026  & 0.4824  & 0.4926 & WITHIN CI \\
C(cefr)[T.B1\_1]                            & -0.0031   & -0.0031  & 0.0025  & -0.0081 & 0.0017 & WITHIN CI \\
C(cefr)[T.B1\_2]                            & -0.0087   & -0.0087  & 0.0025  & -0.0137 & -0.0039 & WITHIN CI \\
C(cefr)[T.B2\_0]                            & -0.0045   & -0.0045  & 0.0028  & -0.0102 & 0.0008 & WITHIN CI \\
C(cefr)[T.XX\_0]                            & 0.0102    & 0.0102   & 0.0013  & 0.0077  & 0.0127 & WITHIN CI \\
C(Sex)[T.M]                                & 0.0018    & 0.0019   & 0.0016  & -0.0012 & 0.0049 & WITHIN CI \\
C(academic\_genre)[T.Life Sciences]          & -0.0141   & -0.0143  & 0.0040  & -0.0220 & -0.0066 & WITHIN CI \\
C(academic\_genre)[T.Sciences \& Technology] & -0.0060   & -0.0060  & 0.0020  & -0.0100 & -0.0021 & WITHIN CI \\
C(academic\_genre)[T.Social Sciences]        & -0.0017   & -0.0017  & 0.0020  & -0.0057 & 0.0022 & WITHIN CI \\
C(language\_env)[T.ESL]                      & -0.0035   & -0.0035  & 0.0017  & -0.0069 & -0.0001 & WITHIN CI \\
C(language\_env)[T.NS]                       & 0.0102    & 0.0102   & 0.0013  & 0.0077  & 0.0127 & WITHIN CI \\
\bottomrule
\end{tabular}
}
\caption{Bootstrap Sensitivity Analysis for detector: \texttt{fastdetectllm}}
\label{tab:bootstrap_fastdetectllm}
\end{table*}

\begin{table*}[ht]
\centering
\resizebox{\textwidth}{!}{%
\begin{tabular}{lcccccc}
\toprule
\textbf{Parameter} & \textbf{Original Value} & \textbf{Bootstrap Mean} & \textbf{Bootstrap Std} & \textbf{CI Lower} & \textbf{CI Upper} & \textbf{Coverage} \\
\midrule
Intercept                                    & 0.8366  & 0.8369  & 0.0175  & 0.8039  & 0.8712 & WITHIN CI \\
C(cefr)[T.B1\_1]                             & -0.0061 & -0.0064 & 0.0189  & -0.0445 & 0.0289 & WITHIN CI \\
C(cefr)[T.B1\_2]                             & -0.0160 & -0.0158 & 0.0169  & -0.0481 & 0.0179 & WITHIN CI \\
C(cefr)[T.B2\_0]                             & -0.0399 & -0.0405 & 0.0175  & -0.0755 & -0.0064 & WITHIN CI \\
C(cefr)[T.XX\_0]                             & -0.1186 & -0.1183 & 0.0107  & -0.1383 & -0.0979 & WITHIN CI \\
C(Sex)[T.M]                                  & -0.0078 & -0.0078 & 0.0130  & -0.0324 & 0.0178 & WITHIN CI \\
C(academic\_genre)[T.Life Sciences]           & 0.0065  & 0.0069  & 0.0185  & -0.0291 & 0.0427 & WITHIN CI \\
C(academic\_genre)[T.Sciences \& Technology]  & 0.0068  & 0.0068  & 0.0172  & -0.0253 & 0.0405 & WITHIN CI \\
C(academic\_genre)[T.Social Sciences]         & -0.0031 & -0.0033 & 0.0154  & -0.0333 & 0.0257 & WITHIN CI \\
C(language\_env)[T.ESL]                       & -0.0020 & -0.0021 & 0.0133  & -0.0273 & 0.0241 & WITHIN CI \\
C(language\_env)[T.NS]                        & -0.1186 & -0.1183 & 0.0107  & -0.1383 & -0.0979 & WITHIN CI \\
\bottomrule
\end{tabular}
}
\caption{Bootstrap Sensitivity Analysis for detector \texttt{gltr}}
\label{tab:gltr_bootstrap}
\end{table*}

\begin{table*}[ht]
\centering
\resizebox{\textwidth}{!}{%
\begin{tabular}{lcccccc}
\toprule
\textbf{Parameter} & \textbf{Original Value} & \textbf{Bootstrap Mean} & \textbf{Bootstrap Std} & \textbf{CI Lower} & \textbf{CI Upper} & \textbf{Coverage} \\
\midrule
Intercept                                    & 0.6493  & 0.6499  & 0.0129  & 0.6247  & 0.6755 & WITHIN CI \\
C(cefr)[T.B1\_1]                             & -0.0094 & -0.0096 & 0.0138  & -0.0373 & 0.0179 & WITHIN CI \\
C(cefr)[T.B1\_2]                             & -0.0612 & -0.0611 & 0.0127  & -0.0849 & -0.0366 & WITHIN CI \\
C(cefr)[T.B2\_0]                             & -0.0870 & -0.0872 & 0.0129  & -0.1118 & -0.0619 & WITHIN CI \\
C(cefr)[T.XX\_0]                             & -0.1011 & -0.1013 & 0.0070  & -0.1146 & -0.0873 & WITHIN CI \\
C(Sex)[T.M]                                  & -0.0043 & -0.0044 & 0.0088  & -0.0213 & 0.0126 & WITHIN CI \\
C(academic\_genre)[T.Life Sciences]          & 0.0248  & 0.0245  & 0.0126  & -0.0016 & 0.0490 & WITHIN CI \\
C(academic\_genre)[T.Sciences \& Technology] & 0.0373  & 0.0369  & 0.0111  & 0.0153  & 0.0597 & WITHIN CI \\
C(academic\_genre)[T.Social Sciences]        & -0.0084 & -0.0090 & 0.0103  & -0.0288 & 0.0115 & WITHIN CI \\
C(language\_env)[T.ESL]                       & 0.0011  & 0.0009  & 0.0084  & -0.0150 & 0.0176 & WITHIN CI \\
C(language\_env)[T.NS]                        & -0.1011 & -0.1013 & 0.0070  & -0.1146 & -0.0873 & WITHIN CI \\
\bottomrule
\end{tabular}
}
\caption{Bootstrap Sensitivity Analysis for detector \texttt{gpt2-base}}
\label{tab:gpt2base_bootstrap}
\end{table*}

\begin{table*}[ht]
\centering
\small
\resizebox{\textwidth}{!}{%
\begin{tabular}{lcccccc}
\toprule
\textbf{Parameter} & \textbf{Original Value} & \textbf{Bootstrap Mean} & \textbf{Bootstrap Std} & \textbf{CI Lower} & \textbf{CI Upper} & \textbf{Coverage} \\
\midrule
Intercept                                   & 0.5402   & 0.5412   & 0.0149   & 0.5127   & 0.5703 & WITHIN CI \\
C(cefr)[T.B1\_1]                            & 0.0274   & 0.0270   & 0.0159   & -0.0050  & 0.0583 & WITHIN CI \\
C(cefr)[T.B1\_2]                            & 0.0008   & 0.0003   & 0.0149   & -0.0266  & 0.0294 & WITHIN CI \\
C(cefr)[T.B2\_0]                            & 0.0060   & 0.0053   & 0.0159   & -0.0255  & 0.0359 & WITHIN CI \\
C(cefr)[T.XX\_0]                            & -0.0240  & -0.0244  & 0.0080   & -0.0393  & -0.0093 & WITHIN CI \\
C(Sex)[T.M]                                & -0.0026  & -0.0030  & 0.0097   & -0.0214  & 0.0150 & WITHIN CI \\
C(academic\_genre)[T.Life Sciences]         & 0.0082   & 0.0078   & 0.0145   & -0.0199  & 0.0352 & WITHIN CI \\
C(academic\_genre)[T.Sciences \& Technology] & 0.0303   & 0.0297   & 0.0133   & 0.0037   & 0.0555 & WITHIN CI \\
C(academic\_genre)[T.Social Sciences]       & -0.0165  & -0.0170  & 0.0125   & -0.0420  & 0.0066 & WITHIN CI \\
C(language\_env)[T.ESL]                      & 0.0119   & 0.0118   & 0.0102   & -0.0083  & 0.0321 & WITHIN CI \\
C(language\_env)[T.NS]                       & -0.0240  & -0.0244  & 0.0080   & -0.0393  & -0.0093 & WITHIN CI \\
\bottomrule
\end{tabular}
}
\caption{Bootstrap Sensitivity Analysis for detector \texttt{gpt2-large}}
\label{tab:gpt2large_bootstrap}
\end{table*}

\begin{table*}[ht]
\centering
\small
\resizebox{\textwidth}{!}{%
\begin{tabular}{lcccccc}
\toprule
\textbf{Parameter} & \textbf{Original Value} & \textbf{Bootstrap Mean} & \textbf{Bootstrap Std} & \textbf{CI Lower} & \textbf{CI Upper} & \textbf{Coverage} \\
\midrule
Intercept                                   & 0.5402  & 0.5412  & 0.0039  & 0.4986  & 0.5139 & WITHIN CI \\
C(cefr)[T.B1\_1]                            & $-0.0102$ & $-0.0103$ & 0.0046  & $-0.0197$ & $-0.0019$ & WITHIN CI \\
C(cefr)[T.B1\_2]                            & $-0.0251$ & $-0.0250$ & 0.0039  & $-0.0331$ & $-0.0178$ & WITHIN CI \\
C(cefr)[T.B2\_0]                            & $-0.0482$ & $-0.0482$ & 0.0044  & $-0.0570$ & $-0.0398$ & WITHIN CI \\
C(cefr)[T.XX\_0]                            & $-0.0026$ & $-0.0026$ & 0.0019  & $-0.0064$ & 0.0009 & WITHIN CI \\
C(Sex)[T.M]                                 & 0.0033  & 0.0033  & 0.0026  & $-0.0018$ & 0.0083 & WITHIN CI \\
C(academic\_genre)[T.Life Sciences]          & 0.0202  & 0.0201  & 0.0052  & 0.0101  & 0.0306 & WITHIN CI \\
C(academic\_genre)[T.Sciences \& Technology] & 0.0046  & 0.0047  & 0.0031  & $-0.0015$ & 0.0106 & WITHIN CI \\
C(academic\_genre)[T.Social Sciences]        & 0.0034  & 0.0034  & 0.0029  & $-0.0022$ & 0.0090 & WITHIN CI \\
C(language\_env)[T.ESL]                       & 0.0187  & 0.0186  & 0.0021  & 0.0144  & 0.0226 & WITHIN CI \\
C(language\_env)[T.NS]                        & $-0.0026$ & $-0.0026$ & 0.0019  & $-0.0064$ & 0.0009 & WITHIN CI \\
\bottomrule
\end{tabular}
}
\caption{Bootstrap Sensitivity Analysis for detector \texttt{llmdet}}
\label{tab:llmdet_bootstrap}
\end{table*}

\begin{table*}[ht]
\centering
\small
\resizebox{\textwidth}{!}{%
\begin{tabular}{lcccccc}
\toprule
\textbf{Parameter} & \textbf{Original Value} & \textbf{Bootstrap Mean} & \textbf{Bootstrap Std} & \textbf{CI Lower} & \textbf{CI Upper} & \textbf{Coverage} \\
\midrule
Intercept & 0.6886 & 0.6890   & 0.0148   & 0.6611   & 0.7175 & WITHIN CI \\
C(cefr)[T.B1\_1] & 0.0381 & 0.0380   & 0.0159   & 0.0065   & 0.0681 & WITHIN CI \\
C(cefr)[T.B1\_2] & 0.0324 & 0.0324   & 0.0143   & 0.0027   & 0.0604 & WITHIN CI \\
C(cefr)[T.B2\_0] & 0.0134 & 0.0129   & 0.0144   & $-0.0165$ & 0.0403 & WITHIN CI \\
C(cefr)[T.XX\_0] & -0.0200 & $-0.0201$& 0.0077   & $-0.0343$& $-0.0051$ & WITHIN CI \\
C(Sex)[T.M] & -0.0020 & $-0.0018$& 0.0105   & $-0.0222$& 0.0183 & WITHIN CI \\
C(academic\_genre)[T.Life Sciences] & -0.0409 & $-0.0409$& 0.0153   & $-0.0699$& $-0.0112$ & WITHIN CI \\
C(academic\_genre)[T.Sciences \& Technology] & 0.0021 & 0.0021 & 0.0136 & $-0.0253$& 0.0298 & WITHIN CI \\
C(academic\_genre)[T.Social Sciences] & -0.0040 & $-0.0043$& 0.0126   & $-0.0290$& 0.0194 & WITHIN CI \\
C(language\_env)[T.ESL] & -0.0215 & $-0.0215$& 0.0105   & $-0.0430$& $-0.0014$ & WITHIN CI \\
C(language\_env)[T.NS] & -0.0200 & $-0.0201$& 0.0077   & $-0.0343$& $-0.0051$ & WITHIN CI \\
\bottomrule
\end{tabular}
}
\caption{Bootstrap Sensitivity Analysis for detector \texttt{radar}}
\label{tab:radar_bootstrap}
\end{table*}

\end{document}